\pdfoutput=1
\documentclass[11pt]{article}

\usepackage[final]{acl}

\usepackage{times}
\usepackage{latexsym}

\usepackage[T1]{fontenc}
\usepackage[utf8]{inputenc}

\usepackage{microtype}

\usepackage{inconsolata}

\usepackage{graphicx}
\usepackage{algorithm}
\usepackage{algorithmic}
\usepackage{multirow}
\usepackage{multicol}
\usepackage{array}
\usepackage{tabularx}
\usepackage{booktabs}  % 导言区加上
\usepackage{bm}
\title{Probing Semantic Alignment, Lexical Invariance, and Syntactic Influence in LLM Metaphor Processing}

\author{
       Fengying Ye$^1$~~~~
       Shanshan Wang$^1$~~~~
       \textbf{Lidia S. Chao$^1$}~~~~
        Derek F. Wong$^1\thanks{Corresponding Author}$~~~~
       \\
    $^1$NLP$^2$CT Lab, Department of Computer and Information Science, University of Macau \\
    {nlp2ct.\{fengying,shanshan\}@gmail.com, \{lidiasc,derekfw\}@um.edu.mo} \\
    }

\begin{document}
\maketitle
\begin{abstract}
Large language models (LLMs) achieve strong performance on metaphor detection and interpretation tasks, yet it remains unclear what such behavioral success reveals about metaphor processing. We present a diagnostic analysis that examines the limits of behavioral evidence by probing three complementary dimensions: semantic attribute alignment, lexical invariance, and syntactic sensitivity. Using geometric probing, we assess whether model-generated interpretations align with reference semantic attributes; through context-varying substitution, we analyze the stability of lexical associations between metaphorical and literal expressions; and via controlled syntactic perturbations, we examine sensitivity in metaphor detection. Our analysis reveals that LLM-generated interpretations can exhibit semantic drift relative to reference attributes; stable lexical anchors persist across contextual conditions, potentially supporting conventional metaphors while biasing novel metaphors requiring contextual integration; and detection performance is sensitive to syntactic irregularities. These findings suggest that strong behavioral performance may reflect heterogeneous underlying signals, highlighting the need for caution when interpreting metaphor benchmarks as evidence of robust, integrated semantic understanding.
\end{abstract}

\section{Introduction}
Metaphor is a pervasive and sophisticated aspect of human language \cite{gibbs2008cambridge}. Processing metaphors requires more than recognizing unusual word usage; it involves identifying implicit relationships between attributes across semantic domains \cite{croft1993role}. With strong text comprehension capabilities and large-scale pretraining \cite{yang2024harnessing}, LLMs have been widely applied to metaphor detection and interpretation. However, it remains unclear whether such performance is accompanied by behavioral evidence consistent with deep metaphor processing.

Linguistic theories of metaphor such as Selection Preference Violation (SPV) and the Metaphor Identification Procedure (MIP) characterize metaphor through violations of conventional selection preferences or literal word meanings in context \cite{wilks1975preferential,group2007MIP}. Conceptual Metaphor Theory (CMT), in contrast, views metaphors as cross-domain mappings between a source domain describing tangible objects and a target domain representing abstract ideas \cite{lakoff1980metaphors}. A central difficulty emphasized by these theories is that the core mapping in a metaphor is often implicit rather than explicitly expressed. As a result, models may generate interpretations that focus on salient characteristics while failing to capture the intended mapping attribute. In this work, a \textit{semantic attribute} refers to the salient property selectively projected from the source domain to the target domain in a metaphor. For example, \textit{“The computer is a turtle”} may evoke (low) speed, but also peripheral attributes of turtle (e.g., (long) lifespan), complicating interpretation. This motivates analyzing metaphor processing in terms of whether model-generated interpretations align with the intended semantic attribute \cite{do-dinh-etal-2018-weeding}.

Recent studies have applied LLMs to metaphor processing across cultural contexts \cite{ichien2024large}, cross-lingual settings \cite{shao-etal-2024-cmdag}, and different genres \cite{toker-etal-2024-dataset,wang-etal-2024-best}. However, prior work has also identified behavioral patterns that complicate the interpretation of performance on metaphors: \citet{wachowiak-gromann-2023-gpt} identify \textbf{trigger word} effects, where interpretations are biased toward highly associated lexical items rather than context. For example, the word \textit{arm} may bias interpretations toward war-related meanings, even when the context does not support such mappings. While prior work focused on prediction outcomes such as multiple-choice accuracy \cite{li-etal-2024-multiple, pmlr-v139-zhao21c}, we investigate behavioral patterns that shed light on how LLMs process metaphors.

We investigate LLM metaphor processing from a diagnostic perspective, asking whether observed behavioral performance reflects a unified semantic mechanism or heterogeneous underlying signals. To this end, we analyze model behavior along three complementary dimensions. First, we examine whether model-generated interpretations exhibit semantic attribute alignment with reference interpretations. Second, we test whether LLMs exhibit context-invariant lexical associations, which may indicate reliance on stable lexical anchors rather than contextual integration while processing metaphor. Third, we assess how syntactic disruption influences metaphor detection, probing sensitivity to structural cues. Together, these dimensions provide a structured view of metaphor processing, allowing us to distinguish between semantic alignment, lexical bias, and syntactic sensitivity in LLM behavior. Our contributions are as follows:
\begin{itemize}
    \item We propose a geometric probing framework as a behavioral proxy to measure semantic attribute alignment in metaphor interpretation.
    \item We probe the persistence of context-invariant lexical associations using Metaphorical Imagination tasks under different context settings.
    \item We analyze the effects of controlled syntactic perturbations on metaphor detection by selectively disrupting word order, part-of-speech, and positional placement of metaphorical words.
    \item We identify consistent behavioral patterns across LLMs under these probes, highlighting behavioral regularities and limitations in how models process metaphor-related inputs.
\end{itemize}
Rather than focusing on performance comparison, we adopt a diagnostic perspective to analyze how LLMs process metaphors under controlled probing conditions. Definitions and terminology used in this paper are provided in Appendix~\ref{appendix:definition}.

\section{Related Work}
\subsection{Metaphor Detection}
Metaphor detection aims to determine whether a given input contains metaphorical expressions. Early approaches relied on rule-based linguistic frameworks \cite{group2007MIP,dodge-etal-2015-metanet}, whereas later neural models formulated the task as supervised classification \cite{rai2020survey}. To better capture metaphorical usage, subsequent work incorporated linguistic and conceptual signals, such as word association statistics, to model semantic relatedness and deviation \cite{wan-etal-2020-using,church1990word}. Because metaphorical expressions vary across languages, genres, and socio-cultural contexts, prior studies have also explored domain-specific detection settings \cite{montefinese2014adaptation,brysbaert2009moving,cheung2009credibility,janschewitz2008taboo,wang-etal-2025-benchmarking}. More recently, LLM-based approaches have achieved strong performance on metaphor detection benchmarks \cite{mao-etal-2024-metapro,ge2022explainable,choi-etal-2021-melbert}.

%\subsection{Metaphor Interpretation}
%Metaphor interpretation focuses on uncovering metaphorical mappings and their associated conceptual domains. Linguistic analyses highlight that metaphors rely on systematic cross-domain mappings rather than isolated lexical substitutions \cite{sullivan2013frames}. In computational settings, a common approach to metaphor interpretation is metaphor component recognition, which aims to identify the source and target domains underlying metaphorical expressions \cite{sengupta-etal-2024-analyzing,ge2022explainable}. Prior work has also reported trigger word effects in this setting \cite{wachowiak-gromann-2023-gpt}. Other approaches incorporate explicit reasoning mechanisms to guide metaphor interpretation. Frameworks combining Chain-of-Thought reasoning with external knowledge resources have been proposed to guide models toward more structured interpretations \cite{tian-etal-2024-theory}. Additionally, some studies focus on explaining metaphors through literal paraphrases, often drawing on SPV as a learning signal \cite{mao-etal-2024-metapro}. While prior work primarily evaluates metaphor interpretation through task-level performance, our work adopts a diagnostic perspective to analyze what such performance reveals about LLM behavior.

\subsection{Metaphor Interpretation}
Metaphor interpretation concerns explaining the meaning of metaphorical expressions, including the conceptual mappings that relate source and target domains. Linguistic theories emphasize that such interpretations arise from systematic cross-domain mappings rather than isolated lexical substitutions \cite{sullivan2013frames}. In computational research, one line of work formulates metaphor interpretation as metaphor component recognition, aiming to identify the source and target domains underlying a metaphor \cite{sengupta-etal-2024-analyzing,ge2022explainable}. Prior studies have also shown that performance in this setting can be affected by trigger word effects \cite{wachowiak-gromann-2023-gpt}. A separate line guides interpretation with explicit reasoning mechanisms. For example, some frameworks combine Chain-of-Thought (CoT) reasoning with external knowledge resources to generate more structured interpretations \cite{tian-etal-2024-theory}. Other studies treat interpretation as literal paraphrase generation, often including SPV as a supervision signal \cite{mao-etal-2024-metapro}.

\begin{figure*}[t]
  \centering %图片要改
  \includegraphics[width=0.95\linewidth]{framework3.pdf}
  \caption {Overview of the experimental framework. Spatial Analysis probes attribute-level semantic alignment in metaphor interpretation. Metaphorical Imagination probes the persistence of stable lexical associations under two contextual settings. Syntactic Shuffle analyzes the influence of syntactic cues on metaphor detection.}
   \label{fig:framework}
\end{figure*}

%\subsection{Metaphor in LLMs}
%Recent work has investigated metaphor processing with LLMs across diverse tasks and settings. At the task level, prior studies have explored metaphor interpretation, generation, and detection. \citet{wang2024chinese} proposed a multi-stage prompt-based framework that incorporates conceptual background knowledge for Chinese metaphor interpretation, while other work examined LLMs’ associative capabilities in metaphor generation, particularly in terms of creativity and novelty \cite{distefano2024automatic, su2025metaphor}. In metaphor detection, CoT prompting has been used to improve performance in multimodal scenarios \cite{xu-etal-2024-exploring}. Beyond task performance, representation-level analyses have shown that pretrained models encode metaphor-related structure in contextual embeddings \cite{aghazadeh-etal-2022-metaphors}. LLMs have also demonstrated cross-lingual metaphor detection without explicit fine-tuning \cite{wachowiak-gromann-2023-gpt}. However, for similar non-literal expression like idioms, surface-level distractors remains the primary challenge in cross-lingual alignment \cite{Ye-etal-2026-G-idiom}. Therefore, as \citet{ge2023survey} noted, existing work primarily reports task-level accuracy leaving open what such performance reveals about the underlying mechanisms of metaphor processing in LLMs.

\subsection{Metaphor in LLMs}

Recent work has investigated metaphor processing in LLMs at both the task and representation levels. At the task level, prior studies have examined metaphor interpretation, generation, and detection. \citet{wang2024chinese} proposed a multi-stage prompt-based framework incorporating conceptual background knowledge for Chinese metaphor interpretation, while other work explored LLMs’ associative capabilities in metaphor generation, particularly with respect to creativity and novelty \cite{distefano2024automatic, su2025metaphor}. In metaphor detection, CoT prompting has been shown to improve performance in multimodal settings \cite{xu-etal-2024-exploring}. 

Beyond task performance, representation-level analyses suggest that pretrained models encode metaphor-related structure in contextual embeddings \cite{aghazadeh-etal-2022-metaphors}, and LLMs have demonstrated cross-lingual metaphor detection without explicit fine-tuning \cite{wachowiak-gromann-2023-gpt}. However, evidence from related non-literal phenomena such as idioms suggests that surface-level distractors continue to pose challenges for cross-lingual semantic alignment \cite{Ye-etal-2026-G-idiom}. Taken together, existing work has largely focused on performance outcomes, while representation-level evidence remains indirect, leaving open how LLMs actually process metaphor during inference, and whether task success reflects genuine metaphor understanding or shallow heuristic strategies \cite{ge2023survey}.

\section{Methodology\label{sec:Methodology}}
We propose a diagnostic framework to examine LLM metaphor processing under controlled probing conditions. Our design is motivated by prior observations, including the limited diagnostic value of answer-based evaluation, trigger word effects, and sensitivity to syntactic irregularities. We construct targeted experiments to characterize LLM behavior at both the interpretation and detection levels. Specifically, there are three complementary dimensions: (1) semantic attribute alignment, (2) context-invariant lexical associations, and (3) syntactic influence. An overview is shown in Figure~\ref{fig:framework}.

\subsection{Spatial Analysis}\label{sec:Spatial Analysis}
\textbf{Problem Definition.} We propose a geometric probe to characterize semantic attribute alignment in metaphor interpretation, following similarity-based analysis frameworks \cite{wegmann-nguyen-2021-capture}. In our setting, each target metaphor sentence \(m_i\) is paired with a related metaphor instance \(m_i'\) that differs in surface form but shares the same underlying semantic attribute. We denote the LLM-generated interpretation of \(m_i\) as \(M_i\).

To construct a local reference region for \(m_i\), we define a \textbf{reference plane} \(\gamma_i\) as the affine subspace spanned by \(\{R_i, R_i', S_i\}\). Here, \(R_i\) and \(R_i'\) are human-annotated interpretations that capture the shared semantic attribute, and thus serve as the primary semantic anchors. The third point, \(S_i\), is a model-generated literal paraphrase of the target sentence, providing a complementary literal anchor tied to the same input.

We represent all sentences in a shared embedding space and quantify the alignment of \(M_i\) with \(\gamma_i\) by measuring its deviation from the reference plane. Importantly, \(\{R_i, R_i', S_i\}\) is used only to construct a local diagnostic reference, rather than a globally calibrated semantic manifold. The resulting geometry is therefore interpreted as a comparative signal of alignment, not as an absolute notion of semantic correctness or a recovery of ground-truth semantic structure. All measures are interpreted comparatively across instances and models.

\begin{figure}
  \centering
  \includegraphics[width=\columnwidth]{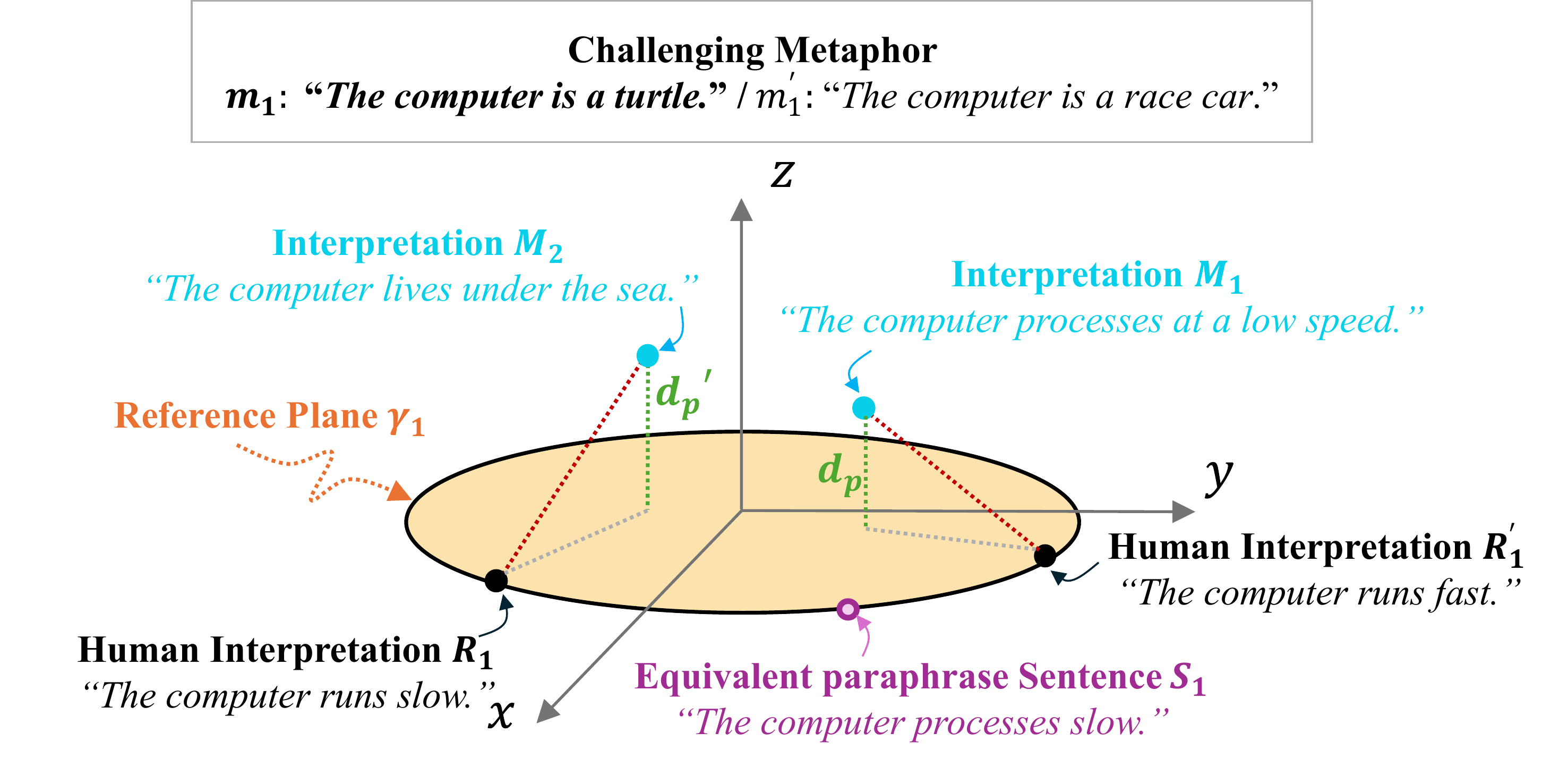}
  \caption{Example of the perpendicular distance \(d_p\) in the embedding space. \(d_p\) measures the deviation of two LLM-generated interpretations \(M_1\) and \(M_2\) for \(m_1\) from the reference plane \(\gamma_1\) (defined by \(\{R_1, R_1', S_1\}\)).}
  \label{fig:spatial distribution}
\end{figure}

\begin{figure}
  \centering
  \includegraphics[width=0.9\columnwidth]{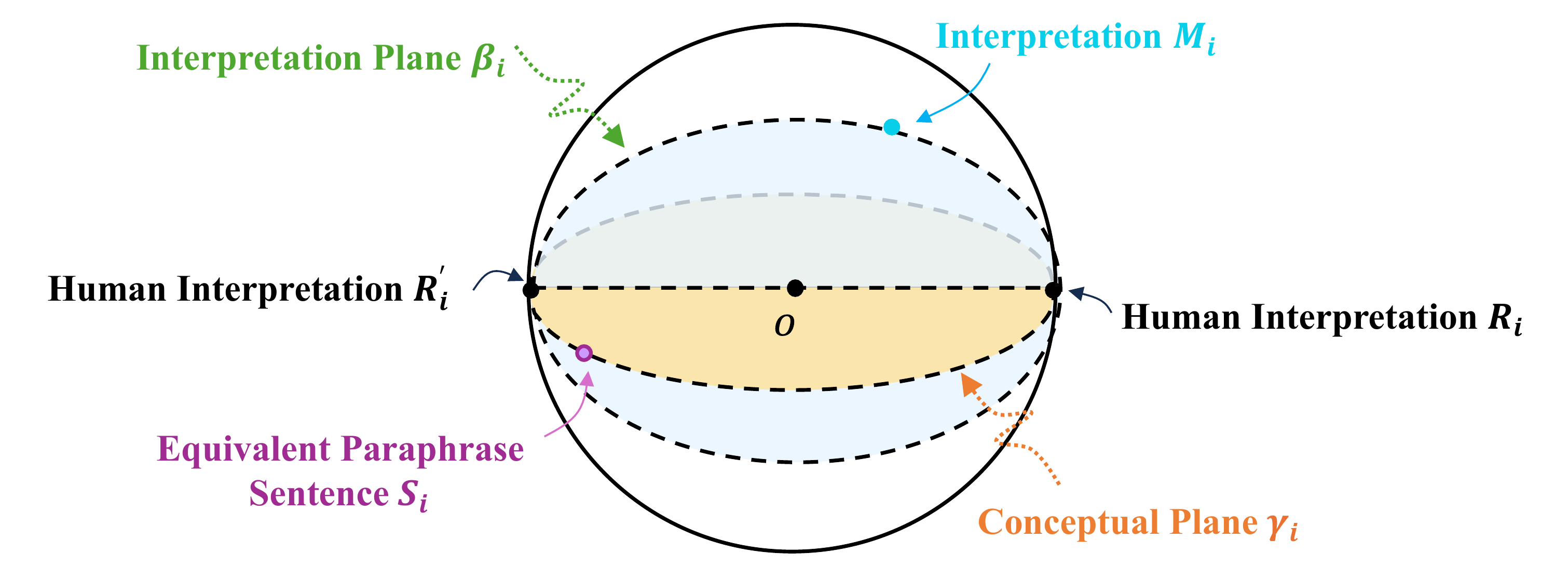}
    \includegraphics[width=0.9\columnwidth]{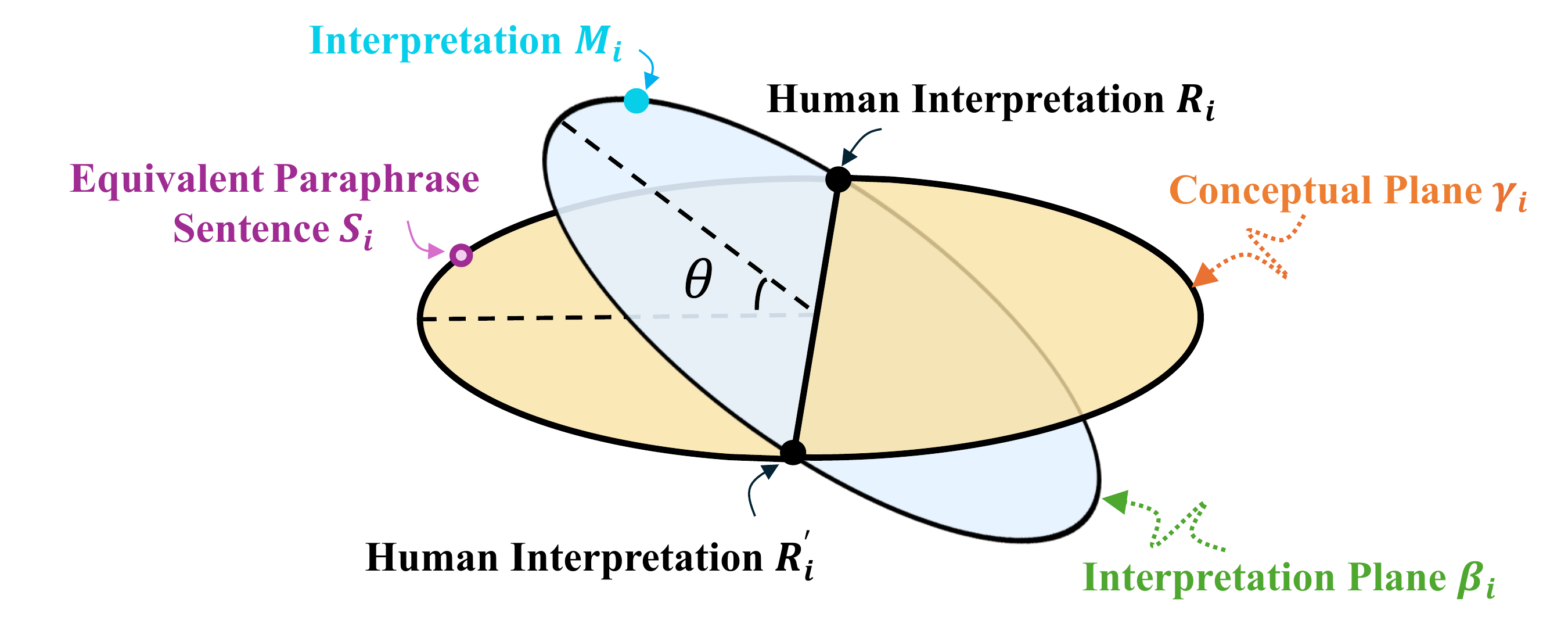}
  \caption{Illustration of the angle \(\theta\) between the reference plane \(\gamma_i\) (defined by \(\{R_i, R_i', S_i\}\)) and the interpretation plane \(\beta_i\) (defined by \(\{R_i, R_i', M_i\}\)) in the embedding space.}
  \label{fig:cosine}
\end{figure}

\textbf{Measures.} As illustrated in Figure~\ref{fig:spatial distribution} and Figure~\ref{fig:cosine}, we define two complementary measures to quantify alignment:

\(d_p\): the perpendicular distance from \(M_i\) to the reference plane \(\gamma_i\), capturing the magnitude of geometric deviation in the embedding space.

\(\cos\theta\): the cosine similarity between the reference plane \(\gamma_i\) and the \textbf{interpretation plane} \(\beta_i\), where \(\beta_i\) is spanned by \(\{R_i, R_i', M_i\}\), measuring the orientation difference between the two planes.

Together, \(d_p\) and \(\cos\theta\) capture complementary aspects of alignment, reflecting both the extent and direction of deviation of model interpretations from the reference semantic region.

%Motivated by prior observations of trigger word effects in LLM metaphor processing, we investigate whether LLMs encode stable lexical associations between words and their metaphorical or literal uses that persist across contextual conditions. We compare two settings: \textit{contextualized} generation, where the target word is produced given its sentence context, and \textit{decontextualized} (word-only) generation, where the model is prompted with the target word in isolation. We instantiate this comparison in two complementary directions: In \textit{Literal-to-Metaphor (LM)}, LLMs generate meaning-equivalent metaphorical words from literal inputs, either in isolation or embedded in a literal context. In \textit{Metaphor-to-Literal (ML)}, LLMs follow the same input settings but generate meaning-equivalent literal words from metaphorical inputs. Similarities between contextualized and decontextualized generations are interpreted as an indicator of context-invariant lexical associations, reflecting the persistence of stable lexical anchors in metaphor processing.
\subsection{Metaphorical Imagination\label{sec:Metaphorical Imagination}} 
Motivated by prior observations of trigger word effects, we examine whether LLMs rely on lexical associations that remain stable across contextual conditions in metaphor processing. We compare two generation settings: \textit{contextualized} generation, in which the target word is produced given its sentence context, and \textit{decontextualized} generation, in which the model is prompted with the target word alone. We operationalize this comparison in two directions. In \textit{Literal-to-Metaphor (LM)}, the model generates a metaphorical counterpart for a literal input, either in isolation or within a literal context. In \textit{Metaphor-to-Literal (ML)}, it generates a literal counterpart for a metaphorical input under the same two settings. Similarity between contextualized and decontextualized outputs is interpreted as a diagnostic signal of context-invariant lexical associations, indicating the persistence of stable lexical anchors in metaphor processing.

\subsection{Syntactic Shuffle\label{sec:Syntactic Shuffle}} 
Metaphors are often associated with characteristic syntactic patterns \cite{sullivan2013frames}. Because part-of-speech structure and word order encode key syntactic relations, such as argument structure and modifier attachment, perturbing them disrupts compositional structure while largely preserving lexical content. This enables us to test whether metaphor detection relies on integrated sentence structure or on shallower heuristic cues.

We consider three shuffle scenarios. 1) \textbf{Random Shuffle}: words are randomly reordered, disrupting both syntactic structure and semantic coherence. 2) \textbf{POS Shuffle}: the metaphorical word is replaced with a near-synonymous alternative of a different POS, introducing syntactic irregularity while keeping lexical meaning relatively similar. 3) \textbf{Metaphorical Word Reposition}: the metaphorical word is moved to the beginning, a random intermediate position (excluding the original, initial, and final positions), or the end of the sentence, allowing us to assess sensitivity to its positional placement. By comparing detection results across these conditions, we evaluate the extent to which metaphor detection is sensitive to syntactic regularity and positional cues.

\section{Experiment}
\subsection{Datasets}
\begin{table*}
\small
\setlength{\tabcolsep}{12pt}
\renewcommand{\arraystretch}{1.2}
\centering
\begin{tabular}{p{3cm} p{1.5cm} p{6.8cm}}
\hline
\textbf{Dataset / Setting} & \textbf{Instances} & \textbf{Example (Metaphor | Literal)} \\\hline
Fig-QA & 2.6k & The computer \textbf{is a race car}. \,|\, The computer runs fast. \\
       &      & The computer \textbf{is a turtle}. \,|\, The computer runs slow. \\\hline
MUNCH (Context) & 2.9k & The council \textbf{appealed} by case stated ... \\&      & The council \textbf{petitioned} by case stated ... \\\hline
MUNCH (Word) & 2.9k & \textbf{appealed} \,|\, \textbf{petitioned} \\
MUNCH (Original) & 2.9k & The council \textbf{appealed} by case stated. \\
MUNCH (POS) & 1.3k & The council \textbf{complainant} (n.) by case stated. \\
MUNCH (Random) & 2.9k & council case \textbf{appealed} stated by The. \\
MUNCH (Beginning) & 2.9k & \textbf{appealed} The council by case stated. \\
MUNCH (Middle) & 2.9k & The council by case \textbf{appealed} stated. \\
MUNCH (End) & 2.9k & The council by case stated \textbf{appealed}. \\
\hline
\end{tabular}
\caption{Dataset statistics and example instances for Fig-QA (used in Spatial Analysis) and MUNCH (used in Metaphorical Imagination and Syntactic Shuffle). Different MUNCH rows correspond to distinct experimental settings derived from the same set of base instances. Examples are shown as Metaphor | Literal.}
\label{tab:dataset}
\end{table*}

Table \ref{tab:dataset} summarizes the datasets used in our experiments, each corresponding to a specific probing setting introduced in Section \ref{sec:Methodology}. For the spatial analysis experiment, we use Fig-QA, a human-annotated resource designed for Winograd-style metaphorical language understanding \cite{liu-etal-2022-testing}, licensed under the MIT License. Fig-QA organizes instances into sets of four, consisting of two metaphors \(\{m_i, m_i'\}\) and their corresponding human-annotated literal interpretations (serving as \(\{R_i, R_i'\}\)), which reflect the same underlying semantic attribute while differing in surface form. To focus the analysis on metaphor interpretation rather than literal variation, model-generated interpretations are constrained, via span-level annotations, to modify only metaphor-relevant parts of each sentence, with spans identified by GPT-4o (highlighted in bold in Fig-QA examples).

For Metaphorical Imagination and Syntactic Shuffle experiments, we adopt the Metaphor Understanding Challenge Dataset (MUNCH) \cite{tong-etal-2024-metaphor}, licensed under CC BY 4.0, a linguistically annotated benchmark derived from the VU Amsterdam Metaphor Corpus \cite{steen2010method}. In MUNCH, each sentence contains a metaphor that is challenging for LLMs and is centered around a single annotated metaphorical word. We extract instances from its paraphrase generation task and construct multiple experimental settings from the same set of base sentences. For syntactic shuffle, tokenization and controlled lexical substitutions are performed using WordNet 2020 \cite{mccrae-etal-2020-english} to introduce systematic perturbations while preserving lexical meanings.

%\subsection{Models}
%We evaluate a diverse set of LLMs, including DeepSeek-V3-671B (V3-671B) \cite{liu2024deepseek}, Qwen-Turbo (Qwen-T) \cite{yang2024qwen2}, GPT-4 \cite{achiam2023gpt}, GPT-4o \cite{hurst2024gpt}, o3-mini and DeepSeek-R1-671B (R1-671B) \cite{guo2025deepseek}, LLaMA-3.1-8B (LLaMA-3.1-8B) \cite{grattafiori2024llama}. To ensure comparability in the spatial analysis, all model-generated interpretations are encoded using \texttt{text-embedding-3-small} from OpenAI. This embedding model is used only for post-hoc geometric analysis and does not affect generation or decision-making. This allows us to perform consistent geometric comparisons across models with different internal representations. Open-source models were run on a T4 GPU of Google Colab. For all generations, we set the temperature parameter to 0 to minimize stochastic variation and improve reproducibility.

\subsection{Models}
We evaluate a diverse set of LLMs: DeepSeek-V3-671B (V3-671B) \cite{liu2024deepseek}, Qwen-Turbo (Qwen-T) \cite{yang2024qwen2}, GPT-4 \cite{achiam2023gpt}, GPT-4o \cite{hurst2024gpt}, o3-mini, DeepSeek-R1-671B (R1-671B) \cite{guo2025deepseek}, and LLaMA-3.1-8B \cite{grattafiori2024llama}. For spatial analysis, all model-generated interpretations are encoded using OpenAI's \texttt{text-embedding-3-small} so that they can be compared in a shared embedding space. This embedding model is used only for post-hoc geometric analysis and does not influence model generation or task outputs. As a result, geometric comparisons remain consistent across models with different internal representations. Open-source models were run on Google Colab with a T4 GPU. For all generations, we set the temperature to 0 to reduce stochastic variation and improve reproducibility.

\begin{figure*}[t]
  \centering
  \begin{minipage}[t]{0.4\textwidth}
    \centering
    \includegraphics[width=\textwidth]{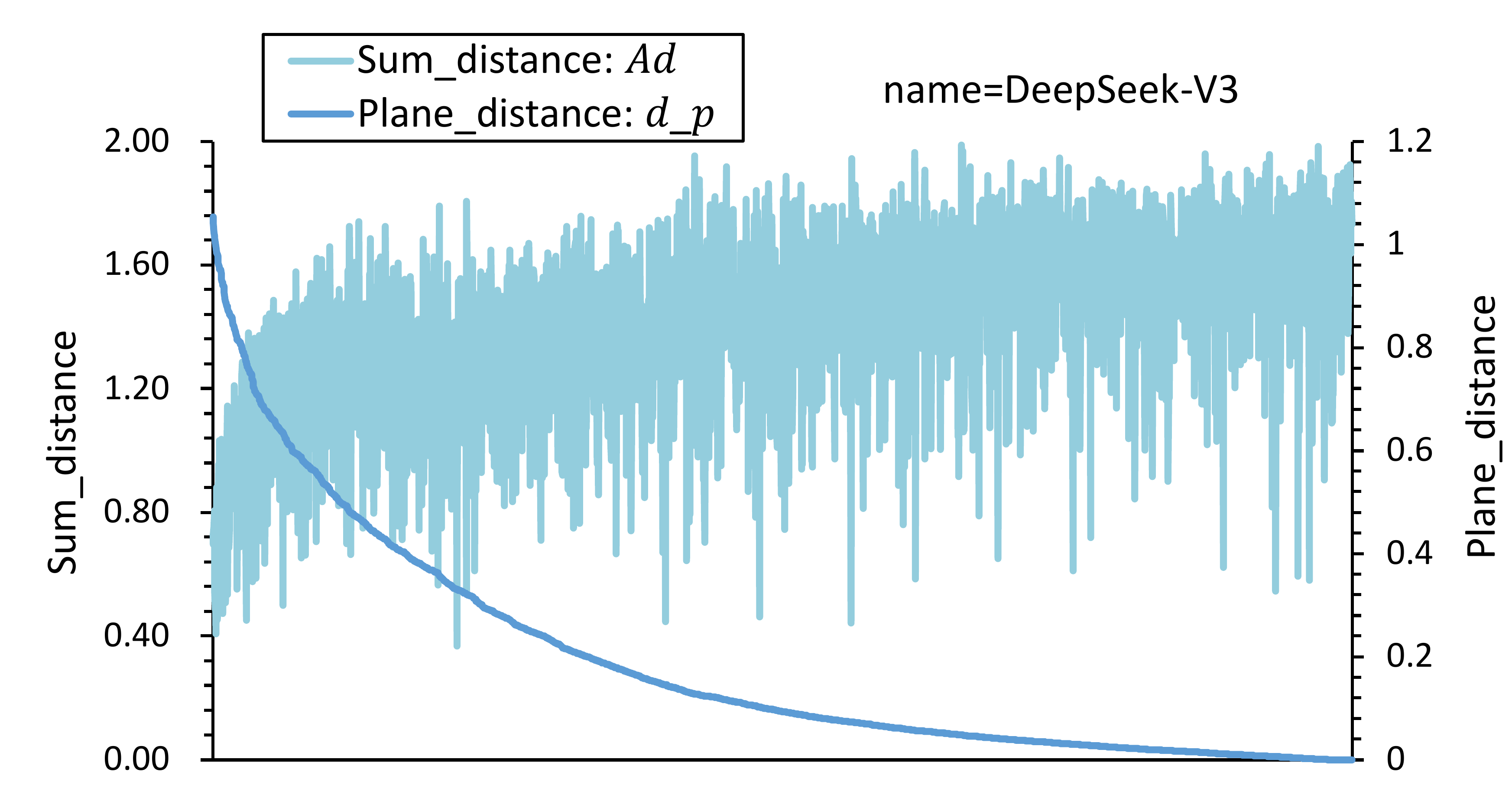}
  \end{minipage}
  \hfill
  \begin{minipage}[t]{0.4\textwidth}
    \centering
    \includegraphics[width=\textwidth]{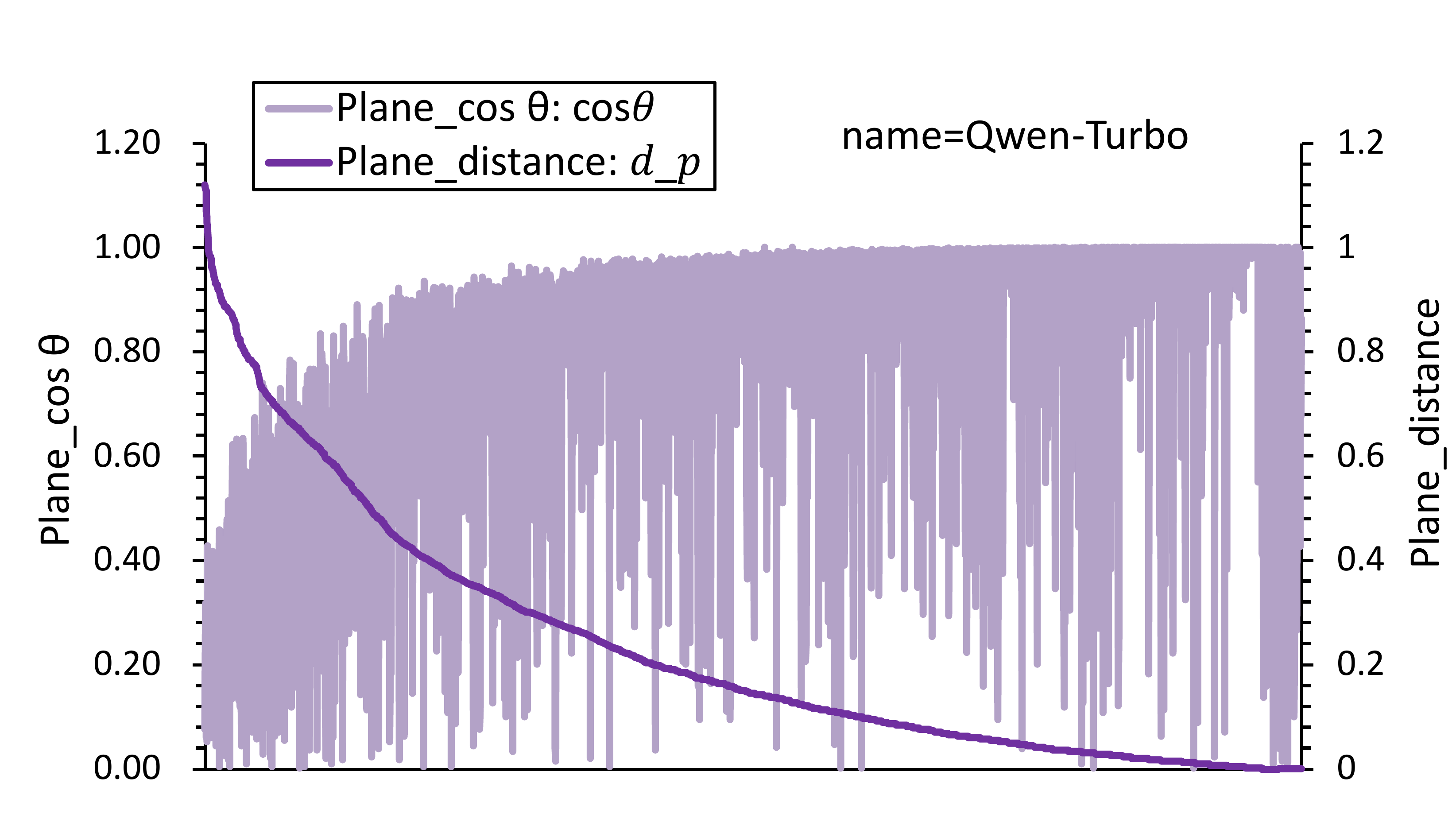}
  \end{minipage}
\caption{Representative examples of two complementary relationships in Spatial Analysis: left, $d_p$ vs.\ $A_d$ for V3-671B; right, $d_p$ vs.\ $\cos\theta$ for Qwen-T.}
  \label{fig:result_spatial}
\end{figure*}

\begin{table*}
  \small
  \setlength{\tabcolsep}{10pt}
  \renewcommand{\arraystretch}{1.2}
  \centering
    \begin{tabular}{llcccc}\cline{1-1}\cline{3-6}
       \textbf{\(m_1\): This blanket is as insulating as a wet tissue.}  &&\multicolumn{4}{c}{\textbf{Accuracy of Multiple-choice Validation}}\\\cline{1-1}\cline{3-6}
        \(L_{11}'\): The blanket keeps me really cozy.  && V3-671B& Qwen-T& GPT-4&GPT-4o
\\\cline{3-6}
        \(R_1\): The blanket does not keep me warm.  && 50.89& 51.69& 50.77&50.92
\\\cline{3-6}
        \(L_{12}'\): The blanket makes me feel quite warm.  && o3-mini& R1-671B& \multicolumn{2}{c}{LLaMA-3.1-8B}\\\cline{3-6}
        \(R_1'\): The blanket keeps me very warm.  && 50.85& 47.31& \multicolumn{2}{c}{46.04}\\\cline{1-1}\cline{3-6}
    \end{tabular}
            \caption{The example and accuracy of multiple-choice validation.}
              \label{tab:multi-choice}
\end{table*}

\begin{table*}
  \setlength{\tabcolsep}{3pt}
  \centering  \begin{tabular}{cccccccc}
    \hline
    & V3-671B&Qwen-T&GPT-4& GPT-4o
 & o3-mini&R1-671B &LLaMA-3.1-8B\\\hline
\({d_p}_M\)& \underline{0.1903}& 0.2319 &0.2267 
&\textbf{0.1772} & 0.2020&0.2063 &0.2866 
\\
${\cos\theta}_M$& \textbf{0.8207}& 0.7835 &\underline{0.7940}&0.7526  & 0.7931&0.7804 &0.7396\\
 \({d_p}_{SD}\)& \underline{0.2194}& 0.2386& 0.2342
&\textbf{0.2182} & 0.2343&0.2204 &0.2649\\
 ${\cos\theta}_{SD}$& \textbf{0.2531}& 0.2742
& \underline{0.2669}&0.2905 & 0.2703&0.2698 &0.2918
\\\hline
\end{tabular}
\caption{Average \(d_p\) and \(\cos\theta\) across models. Mean (M) and standard deviation (SD) are reported. Bold and underlined values indicate the lowest and second-lowest geometric deviation, respectively.}
  \label{tab:result_spatial}
\end{table*}

\setlength{\tabcolsep}{2pt} % 调整列间距

\begin{table*}[t]
  \small
  \setlength{\tabcolsep}{1pt}
  \centering
  \renewcommand{\arraystretch}{1.3}
  \begin{tabular}{
    >{\centering\arraybackslash}m{1.6cm}
    >{\centering\arraybackslash}m{1.6cm}|
    >{\centering\arraybackslash}m{2.9cm}
    >{\centering\arraybackslash}m{3.4cm}
    >{\centering\arraybackslash}m{3.4cm}
    >{\centering\arraybackslash}m{1.1cm}
    >{\centering\arraybackslash}m{1.1cm}
  }
    \hline
    $\bm{R_1}$ & $\bm{{R_1}'}$ & \textbf{\textit{Metaphor}} & $\bm{S}$ & $\bm{M_i}$ & $\bm{d_p}$ & $\bm{\cos\theta}$ \\
    \hline
    \multirow[c]{2}{=}{\centering The monks were very honorable.}
    & \multirow[c]{2}{=}{\centering The monks were not honorable.}
    & The monks had the honor of a knight.
    & ... were highly respected.
    & ... had a prestigious recognition.
    & 0.1153
    & 0.9034 \\
    \cline{3-7}
    &  & The monks had the honor of a lawyer.
    & ... were highly respected.
    & \textit{... had the privilege of legal representation.}
    & \textit{0.7913}
    & \textit{0.2609} \\
    \cline{3-7}
    \hline
    \multirow[c]{2}{=}{\centering I can eat a lot.}
    & \multirow[c]{2}{=}{\centering I eat little.}
    & I have the appetite of an elephant.
    & I consume a moderate amount of food.
    & I have a very large appetite.
    & 0.1367
    & 0.9784 \\
    \cline{3-7}
    &  & I have the appetite of a chipmunk.
    & I consume a moderate amount of food.
    & I have a very small appetite.
    & 0.1573
    & 0.9646 \\
    \cline{3-7}
    \hline
  \end{tabular}
  \caption{Example interpretations illustrating attribute-level semantic alignment. Interpretations with lower alignment to the reference semantic attribute region, indicated by higher $d_p$ and lower $\cos\theta$, are shown in \textit{italic}.}
  \label{tab:case-study}
\end{table*}

\subsection{Implementation Details}
\textbf{Spatial Analysis.} For each metaphor instance $m_i$ and its LLM-generated interpretation $M_i$, we construct a reference plane $\gamma_i$ from $\{R_i, R_i', S_i\}$ and an interpretation plane $\beta_i$ from $\{R_i, R_i', M_i\}$. We quantify the geometric deviation of $M_i$ in the shared embedding space using two complementary measures: the perpendicular distance $d_p$ from $M_i$ to $\gamma_i$, and an angular similarity measure $\cos\theta$ between $\gamma_i$ and $\beta_i$.

To compute these quantities, we derive subspace bases using Singular Value Decomposition (SVD). Because $\gamma_i$ and $\beta_i$ are affine planes, we first represent each plane by centered direction vectors with respect to one anchor point, and stack these vectors into a matrix $A$. We then perform SVD:
\begin{equation}
\label{eq:SVD1}
A = U \Sigma V^{\top}.
\end{equation}
The top singular vectors define an orthonormal basis for the plane, which is used to compute $d_p$ and $\cos\theta$. Larger $d_p$ or smaller $\cos\theta$ indicates greater deviation from the reference semantic attribute. All computations are performed in the shared embedding space.

\textbf{Metaphorical Imagination} MUNCH contains metaphors whose meaning is centered around a single metaphorical word, paired with literal substitutions. LLMs are prompted to generate twenty candidate substitutions for the target word under either metaphorical or literal interpretation settings, providing sufficient lexical diversity. To assess the persistence of lexical associations across contextual conditions, we compare contextualized and decontextualized generations using an \textbf{Anchor Score}. Specifically, when shared words occur between comparative sets, the Anchor Score is set to 1, indicating a shared lexical choice across contexts (a potential lexical anchor). If no word is shared between the two sets, we define Anchor scores by computing the maximum cosine similarity between words across the two sets using 300-dimensional GloVe embeddings \cite{pennington-etal-2014-glove}. We further analyze Anchor Scores across annotated discourse genres to examine whether such lexical invariance varies across discourse types.

%LLMs are required to produce twenty substitutions of the target words, reflecting literal or metaphorical semantic concept. If the metaphor-literal repository exists, generated contextual and de-contextualized predictions should demonstrate substantial overlap or similarity. Therefore, the \textbf{Anchor Scores} is defined to measure the inherent metaphorical imagination regardless of context: Anchor Scores equals to one when shared words occur between comparative sets. For non-overlapping sets, compute the Anchor Scores using maximum cosine similarity based on 300-dimensional GloVe embeddings \cite{pennington-etal-2014-glove}. Additionally, we analyze the results across different genres.

\section{Results \& Analysis\label{sec:Result_Analysis}}
The complete experimental results and model-specific analyses are reported in the Appendix, including the prompts, detailed per-model results, additional breakdowns across conditions.
%\begin{figure}[t]
%  \centering
%  \includegraphics[width=\columnwidth]{sa.pdf}
%\caption{Distributions of (\(d_p\), \(Ad\)) and (\(d_p\), \(\cos\theta\)) for V3-671B and Qwen-T.}
%  \label{fig:result_spatial}
%\end{figure}

\begin{table*}
  \setlength{\tabcolsep}{3pt}
  \centering  \begin{tabular}{cccccccc}
    \hline
    & V3-671B&Qwen-T&GPT-4& GPT-4o
 & o3-mini&R1-671B &LLaMA-3.1-8B\\\hline
\textbf{LM}& 73.30& 73.45&\textbf{78.92}&76.28 & 72.25&\underline{78.11} &65.09\\
\textbf{ML}& 76.96& 75.17&79.22&78.01& \textbf{81.55}&\underline{80.87} &72.86\\\hline
 \textbf{News (LM)}& 74.43& 74.91&\textbf{80.96}&76.90& 72.76&\underline{78.00} &66.28\\
 \textbf{News (ML)}&77.82& 75.23& \textbf{83.77}&79.32& \underline{81.27}&80.68 &72.11\\\hline
 \textbf{Fiction (LM)}& 73.95& 73.55&76.06&\textbf{79.30} & 68.20&\underline{76.22} &64.32\\
 \textbf{Fiction (ML)}&75.21& 73.19& 74.73&75.70& \textbf{78.73}&\underline{77.97} &67.79\\\hline
\textbf{Academic (LM)}& 75.02& 76.11&\textbf{82.71}&77.15& 71.09&\underline{80.37} &65.46\\
\textbf{Academic (ML)}&80.69& 79.69& \underline{84.93}&81.84& \textbf{85.25}&83.62 &75.09\\\hline
\textbf{Conversation (LM)}& 66.80& 66.17& 69.74&71.20& \underline{73.22}&\textbf{75.00} &61.27\\ 
\textbf{Conversation (ML)}&73.35& 71.13& 71.16&73.86& \textbf{81.87}&\underline{81.11} &73.94\\\hline
\end{tabular}
\caption{Anchor Scores for Metaphorical Imagination under LM and ML settings. The four genres include \textit{News}, \textit{Fiction}, \textit{Academic}, and \textit{Conversation}. Best values are shown in bold, and second-best values are underlined.}
  \label{tab:result_imagination}
\end{table*}

\subsection{Semantic Attribute Alignment}
We contrast geometric probing with two evaluation metrics: a similarity-based signal and a discrete answer-based signal.

\textbf{Geometric Signal Characterization.} As a similarity-based evaluation signal, we define \(A_d\) as the sum of cosine similarities between a model-generated interpretation \(M_i\) and two reference interpretations \(\{R_i, R_i'\}\). Interpretations that better align with the intended semantic attribute are expected to exhibit higher \(A_d\) and smaller deviation from the reference plane (\(d_p\)). The magnitude of \(d_p\) does not have a calibrated absolute interpretation; instead, we treat it as a comparative diagnostic signal. As a reference distribution, $d_p$ values for o3-mini generations have median 0.102, interquartile range [0.029, 0.298], and 95th percentile 0.729. We interpret $d_p$ relative to empirical distribution rather than assigning thresholded semantic meaning.

In practice, as shown in Figure~\ref{fig:result_spatial}, lower \(A_d\) is associated with larger \(d_p\), while smaller \(d_p\) corresponds to larger \(\cos\theta\). These trends are supported by Spearman correlations between \(d_p\) and \(A_d\) (\(\rho=-0.62\)) and between \(\cos\theta\) and \(d_p\) (\(\rho=-0.64\)), indicating that \(d_p\) and \(\cos\theta\) capture coherent geometric signals. Distributions across all models are provided in Appendix~\ref{appendix:distribution_dp}. To ensure that these correlations are not artifacts of marginal distributions, we conduct permutation tests that break instance-level pairing; under permutation, correlations collapse to near-zero values, confirming that the observed relationships depend on meaningful alignment rather than spurious structure.

\textbf{Limitations of Discrete Evaluation.} To contrast with discrete evaluation signals, we consider a multiple-choice interpretation setup in which candidate interpretations are restricted to an attribute-aligned option set. For a metaphor pair \(\{m_i, m_i'\}\) in Fig-QA, \(R_i\) denotes the correct interpretation of \(m_i\), while \(R_i'\) corresponds to \(m_i'\), which differs in surface form but shares the same underlying attribute. We generate two paraphrastic variants \(\{L_{i1}', L_{i2}'\}\) of \(R_i'\), yielding a four-way option set \(\{R_i, R_i', L_{i1}', L_{i2}'\}\), and ask models to select the correct interpretation of \(m_i\). The generated paraphrases are used to construct semantically close alternatives around the reference interpretations, so that the candidate options remain attribute-aligned and reduce reliance on superficial lexical cues. 

Despite this controlled design, models exhibit near-chance accuracy when alternatives differ only in fine-grained polarity or intensity (Table~\ref{tab:multi-choice}, e.g., ``does not keep me warm'' vs.\ ``keeps me very warm''). This suggests that even under attribute-aligned candidate sets, discrete evaluation provides limited visibility into model behavior, as it does not reveal how far a generated interpretation deviates from the intended attribute.

\textbf{Main results.} Spatial Analysis provides a complementary geometry-based view. Because the reference plane includes the model-generated literal anchor \(S_i\), we verify it remains semantically close to both human references \(\{R_i, R_i'\}\), supporting its role as a local anchor rather than an off-topic artifact (Appendix~\ref{appendix:si_sanity}). Table~\ref{tab:result_spatial} reports aggregate results: GPT-4o achieves the lowest mean \(d_p\), while V3-671B shows the highest mean \(\cos\theta\). Across models, interpretations still deviate from the intended semantic attribute, suggesting systematic drift. All experiments rely on a shared embedding model, without task-specific fine-tuning.

To further validate that $d_p$ reflects meaningful semantic differences, we conduct a small-scale human evaluation, evaluated by one senior master student with expertise in the relevant
languages. Each sample consists of a metaphor and a model-generated interpretation, which annotators rate on a 3-point scale (incorrect / partial / correct) based on whether the intended metaphorical meaning is captured. We sample instances from both low-$d_p$ and high-$d_p$ regions to contrast extreme cases. Low-$d_p$ interpretations receive substantially higher human alignment scores than high-$d_p$ ones (mean 1.96 vs.\ 0.84; $\Delta=1.12$), indicating a clear separation between the two groups and supporting the semantic relevance of the geometric signal (see Appendix~\ref{appendix:human_eval}).

We also present representative cases in Table~\ref{tab:case-study}. For instance, in the monk/lawyer example, the intended attribute is social honor or respectability; the interpretation ``the privilege of legal representation'' instead shifts toward legal entitlement and does not preserve the intended mapping. Overall, conventional metrics provide only coarse-grained views of alignment, whereas geometric measures reveal finer-grained structure in semantic deviation.

\begin{figure}[t]
  \centering
  \includegraphics[width=\columnwidth]{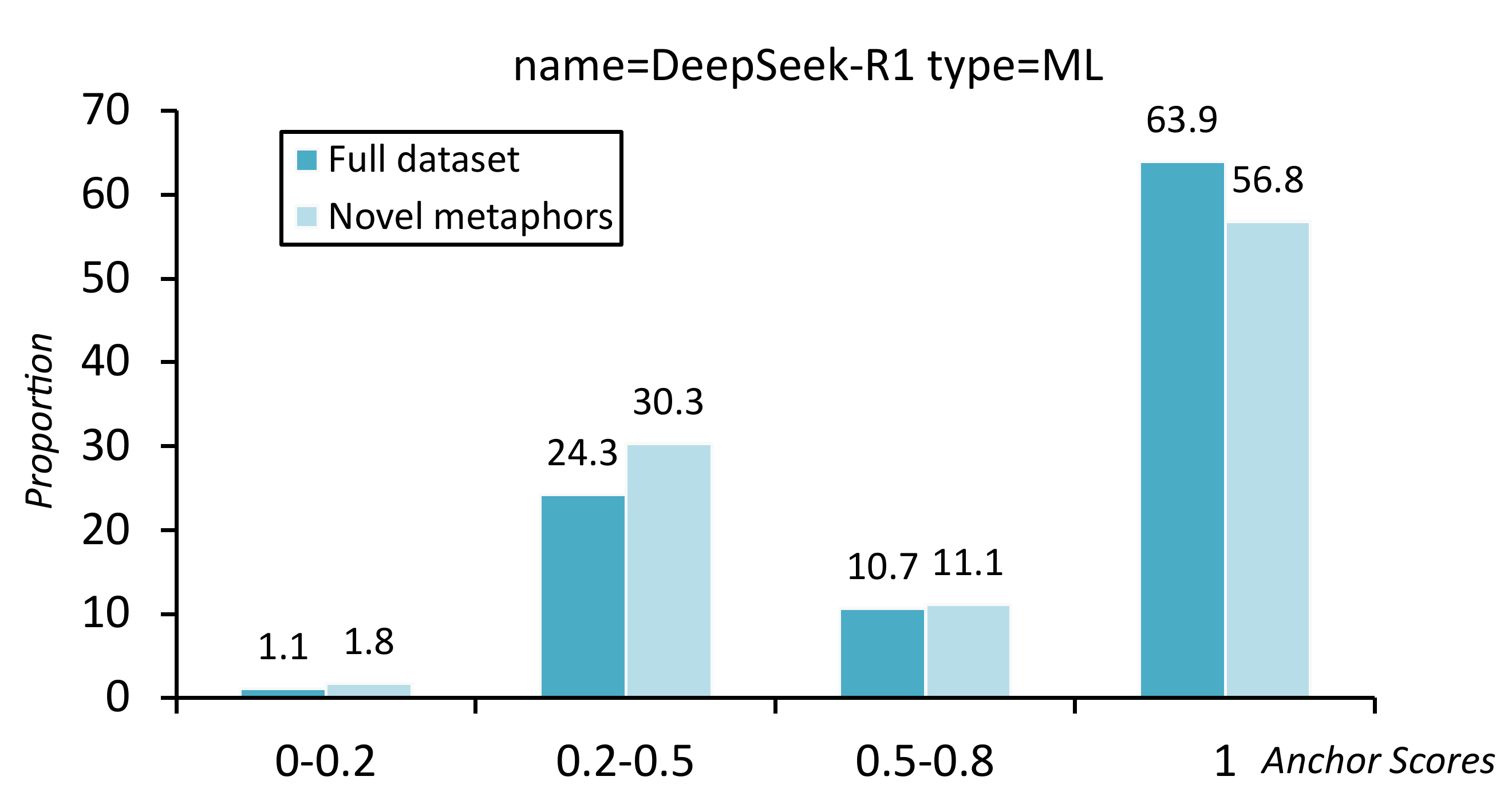}
\caption{Distribution of ML Anchor Scores under R1-671B for the full dataset and the subset of novel metaphors (novelty score \(> 0.3\)).}
  \label{fig:result_imagination}
\end{figure}

\subsection{Lexical Invariance}
Beyond semantic alignment, we next examine whether the lexical content associated with a metaphor remains stable across contextual conditions, or instead shifts more substantially with context. Table~\ref{tab:result_imagination} summarizes results for Metaphorical Imagination. Anchor Scores between contextualized and decontextualized generations are consistently high (approximately 65\%--80\%), suggesting that lexical anchors often persist across metaphor--literal substitution settings. Among the evaluated models, GPT-4, o3-mini, and R1-671B achieve relatively higher scores. Meanwhile, Metaphor-to-Literal (ML) consistently yields higher scores than Literal-to-Metaphor (LM), aligning with prior observations that mapping metaphorical expressions to literal paraphrases is generally more constrained than the reverse \cite{liu-etal-2022-testing}. Genre-level results show stable within-model patterns across LM and ML settings, with o3-mini and R1-671B exhibiting more consistent behavior on conversational metaphors, and GPT-4 showing higher Anchor Scores on news metaphors. This variation may reflect differences in conventionalization and contextual dependence across genres; for example, conversational metaphors may be more conventionalized and compatible with stable lexical priors.

To examine whether this pattern also appears for metaphors that may depend more strongly on context, we analyze MUNCH items with novelty scores \(> 0.3\) \cite{tong-etal-2024-metaphor}. The novelty score ranges from 0 to 1, with higher values indicating greater novelty. Figure~\ref{fig:result_imagination} presents the distribution of ML Anchor Scores under R1-671B for both the full dataset and the subset of more novel metaphors. Although overall scores decrease relative to the full dataset, more than 50\% of cases still reach an Anchor Score of 1, while the remaining instances concentrate around 0.2--0.5. This pattern suggests that persistent lexical anchoring remains common even for novel metaphors, but is less uniformly expressed across items. Complete Anchor Score distributions across models are shown in Appendix~\ref{appendix:overlap_ratio}. 

High Anchor Scores should therefore be interpreted cautiously: lexical priors and contextual evidence may sometimes point in the same direction, so strong overlap does not by itself imply that context is ignored. We interpret these results more narrowly as evidence that lexical associations can remain stable across contextual conditions, rather than as proof of general context-insensitivity. At the same time, this pattern does not necessarily translate into reliable metaphor detection performance, as discussed in Section~\ref{sec:Syntactic Influence}. Stable lexical anchoring may support familiar metaphors by providing readily accessible associations; however, in richer contexts or more novel metaphors, it may also bias interpretations toward highly associated lexical cues, contributing to trigger word effects.

\begin{table*}
  \setlength{\tabcolsep}{10pt}
  \centering  \begin{tabular}{ccccccc}
    \hline
    &  \textit{Original}&\textit{Random} &\textit{POS}
& \textit{Beginning}& \textit{Middle}&\textit{End}\\\hline
 V3-671B& 18.31& 22.71 &
\textbf{23.59}& 
19.14& 22.10&22.63\\
 Qwen-T& 30.37& 1.17 &\textbf{33.59}& 25.46& 27.15&27.50\\
 GPT-4& 34.73& 12.93 &\textbf{43.74}& 36.07& 37.92&37.60\\
 GPT-4o& 28.89& 7.78 &\textbf{36.87}& 30.92& 30.84& 29.98\\ 
 o3-mini& 29.87& 5.40 &\textbf{38.63}& 25.27& 25.85& 25.58\\
 R1-671B& 28.68& 12.22 & \textbf{46.41}& 39.25& 30.88&36.03\\ 
 LLaMA-3.1-8B& 53.36& 50.33 &\textbf{53.81}& 51.75& 53.08& 53.67\\ \hline
\end{tabular}
\caption{Metaphor detection accuracy under Syntactic Shuffle perturbations. Highest values are shown in bold.}
  \label{tab:result_shuffle}
\end{table*}

\subsection{Syntactic Influence}\label{sec:Syntactic Influence}
We finally examine how controlled syntactic perturbations affect metaphor detection. Across perturbation settings, detection accuracy varies substantially across models, as shown in Table~\ref{tab:result_shuffle}. LLaMA-3.1-8B remains relatively stable around chance level (approximately 50\%), showing comparatively little variation across perturbation types. Combined with its low Anchor Scores, this pattern suggests limited responsiveness to the targeted probes on MUNCH. Other models show clearer variation across perturbation settings. In particular, models achieve higher accuracy under POS shuffle than on the original sentences. Because POS shuffle preserves much of the local lexical content while introducing syntactic irregularity, it creates inputs that may amplify cues similar to the kinds of anomalous combinations highlighted in SPV-style accounts of metaphor. By contrast, random shuffle disrupts both syntactic structure and sentence-level coherence, producing more heterogeneous and less interpretable effects across models.

Detection accuracy is relatively stable across positional perturbations (beginning/middle/end), suggesting that the absolute position of a metaphorical word has a weaker effect than the type of structural disruption applied to the sentence. Models therefore appear more responsive to local irregularity, such as POS shuffle, than to changes in word position alone. For example, V3-671B underperforms most other models (except LLaMA-3.1-8B) and even exceeds its original-sentence accuracy under random shuffle, suggesting comparatively unstable behavior under extreme perturbation. We therefore treat random shuffle primarily as a stress-test condition rather than as evidence about natural metaphor processing. Setting random shuffle aside, several syntactic perturbations lead to improved detection accuracy across models, suggesting that syntactic irregularity itself can function as a useful cue for metaphor detection without necessarily implying better metaphor understanding.

Overall, metaphor detection varies across both models and perturbation settings, highlighting the diagnostic value of syntactic manipulation on MUNCH. Taken together with the preceding experiments, these results suggest that metaphor-related behavior in current LLMs is sensitive not only to semantic and lexical factors, but also to surface structural disruption, while providing limited evidence for robust sentence-level syntactic integration under perturbation.

\section{Conclusion}
This work examines LLM metaphor processing from complementary perspectives: semantic attribute alignment, context-invariant lexical associations, and syntactic influence. Spatial Analysis reveals consistent attribute-level deviation in model-generated interpretations. Metaphorical Imagination shows substantial overlap between contextualized and decontextualized generations, suggesting models may rely on stable metaphor and literal lexical associations that support familiar metaphors while biasing decisions toward highly associated cues. Syntactic Shuffle suggests that models respond to syntactic irregularities as heuristic signals for metaphor detection. Overall, although LLMs achieve strong performance on metaphor processing tasks, this performance may arise from a heterogeneous combination of lexical associations and heuristic cues rather than robust semantic understanding. These signals may support behaviors in conventional metaphors but can constrain context-sensitive integration of semantic attributes and syntactic structure. More broadly, Our results highlight the need for evaluation and modeling approaches that probe attribute-level alignment, contextual reasoning, and robust syntactic integration beyond pattern-based cues.

\section*{Limitations}
Our analysis is subject to several limitations. First, Spatial Analysis relies on a constructed reference semantic attribute region derived from human interpretations and LLM-generated sentences. While this provides a behavioral proxy for analyzing model outputs, it does not directly capture underlying cognitive representations of semantic attributes. In addition, the two geometric measures \(d_p\) and \(\cos\theta\) depend on the embedding space used for analysis, different embedding choices may affect absolute distances. Moreover, we introduce a third reference sentence \(S_i\) to enable a locally structured geometric comparison. This formulation represents a pragmatic design choice that balances interpretability and representational richness, and we do not claim that the resulting dimensionality is optimal.

Second, the Metaphorical Imagination probe focuses on metaphors instantiated by a single annotated word, and therefore primarily captures word-level metaphor-literal associations. Our findings do not directly generalize to multi-word or discourse-level metaphors. Understanding such phenomena likely requires different experimental designs and remains an important direction for future work. 

Furthermore, in Syntactic Shuffle, random shuffling is not intended to model natural language use, but rather to function as an extreme stress test. We do not interpret results under this condition as reflecting natural metaphor processing. Instead, controlled POS and positional perturbations provide interpretable evidence about sensitivity to syntactic irregularity. The observation that metaphor detection accuracy can increase under such conditions suggests that detection behavior may, rely on lexical or heuristic cues independent of sentence-level semantic and syntactic integration.

Finally, our experiments are conducted on English metaphor datasets, and the generality of our findings to other languages or culturally specific metaphors remains to be explored.

\section*{Ethics Statement}
This work uses two publicly available datasets: Fig-QA \cite{liu-etal-2022-testing} and MUNCH \cite{tong-etal-2024-metaphor}. These datasets are used solely for probing experiments on LLMs. The experiments strictly excluded any materials associated with personal identifiers or sensitive data categories. 

\section*{Acknowledgments}
This work was supported in part by the Science and Technology Development Fund of Macau SAR (Grant Nos. FDCT/0007/2024/AKP, EF2024-00185-FST), the UM and UMDF (Grant Nos. MYRG-GRG2024-00165-FST-UMDF, MYRG-GRG2025-00236-FST), the Tencent AI Lab Rhino-Bird Research Program (Grant No. EF2023-00151-FST), the Stanley Ho Medical Development Foundation (Grant No. SHMDF-AI/2026/001), and the National Natural Science Foundation of China (Grant No. 62266013).

\bibliography{main}
\appendix
\section{Definitions and Terminology}\label{appendix:definition}
To support the motivation and experimental design of this study, we introduce several key definitions and terms. Table~\ref{tab:definition_part1} provides a consolidated list of the terms used throughout the paper and their definitions.

\section{Prompts for Spatial Analysis}\label{appendix:spatial_analysis}
\begin{figure*}[h]
  \centering
  \includegraphics[width=\columnwidth]{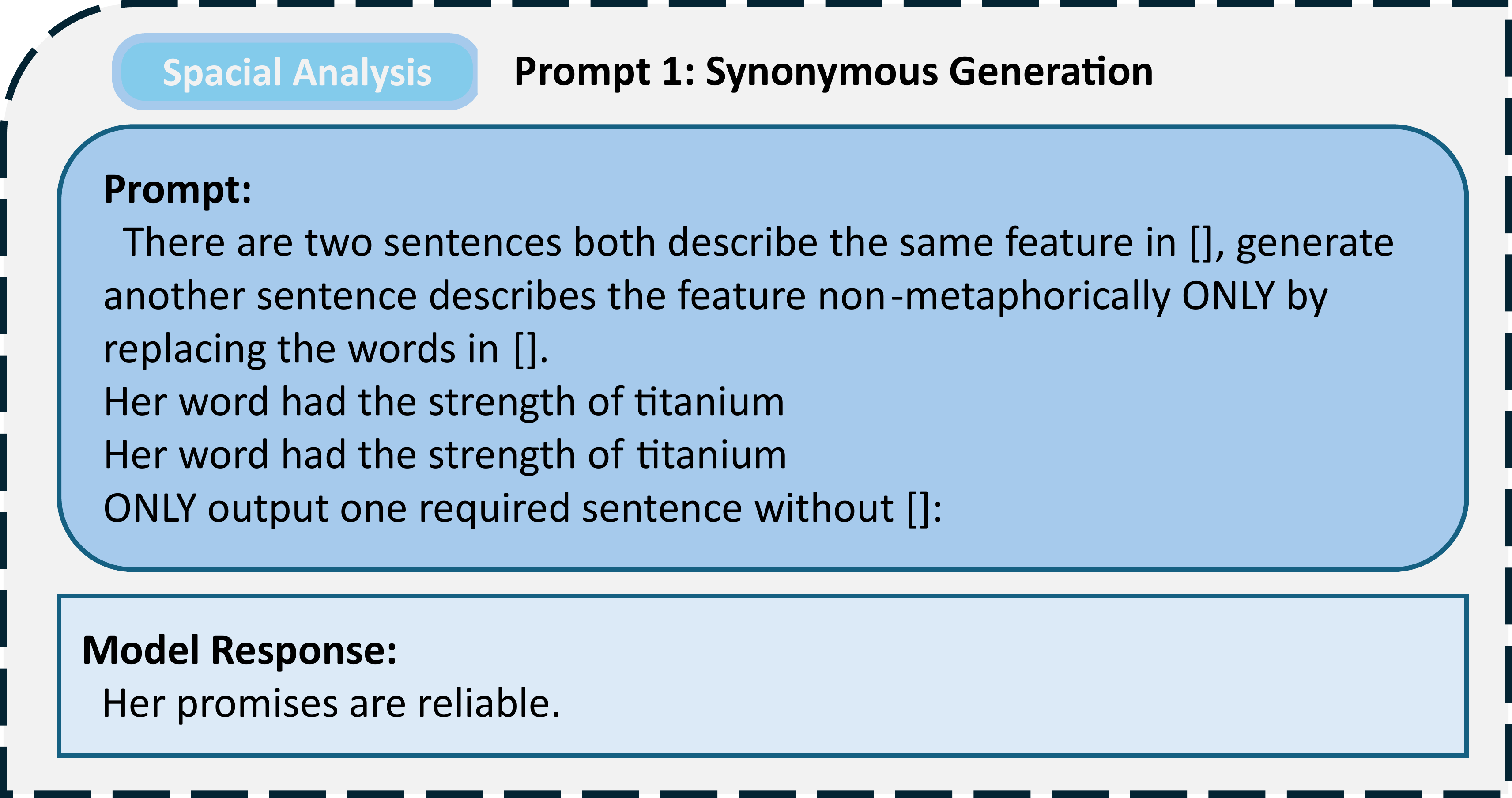}      \includegraphics[width=\columnwidth]{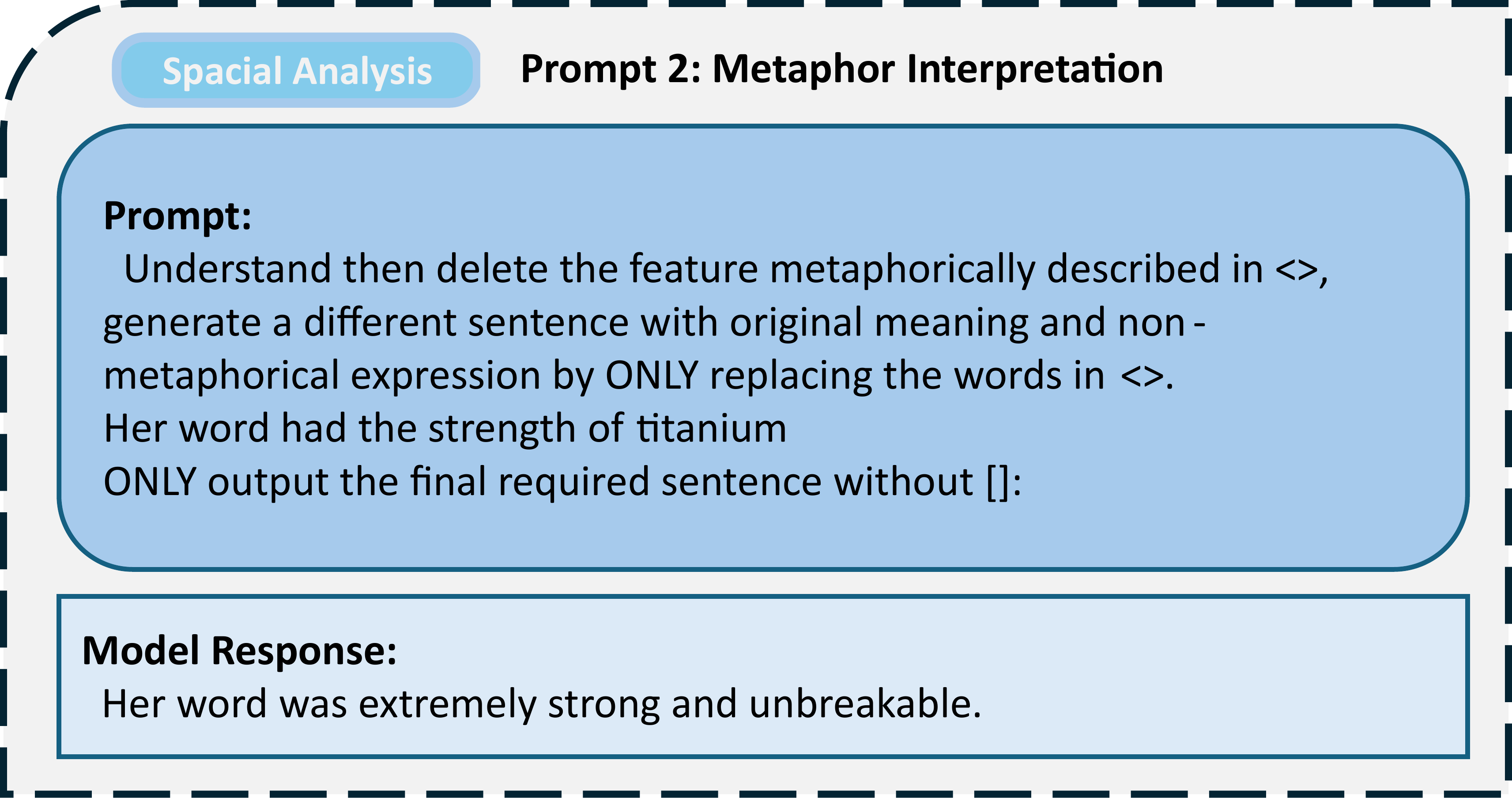}
  \caption{The prompts of Spatial Analysis.}
  \label{fig:prompts_spatial}
\end{figure*}

Figure~\ref{fig:prompts_spatial} shows the two prompts used in the Spatial Analysis experiment. Prompt~1 instructs the model to generate a literal sentence \(S\) that is a semantic paraphrase of the human-annotated interpretations. Prompt~2 instructs the model to generate a literal interpretation \(M_i\) of a given metaphor \(m_i\) by replacing the metaphorical expression with a literal alternative.

\section{Prompts for Syntactic Shuffle}\label{appendix:syntactic_shuffle}
\begin{figure*}[h]
  \centering
  \includegraphics[width=\columnwidth]{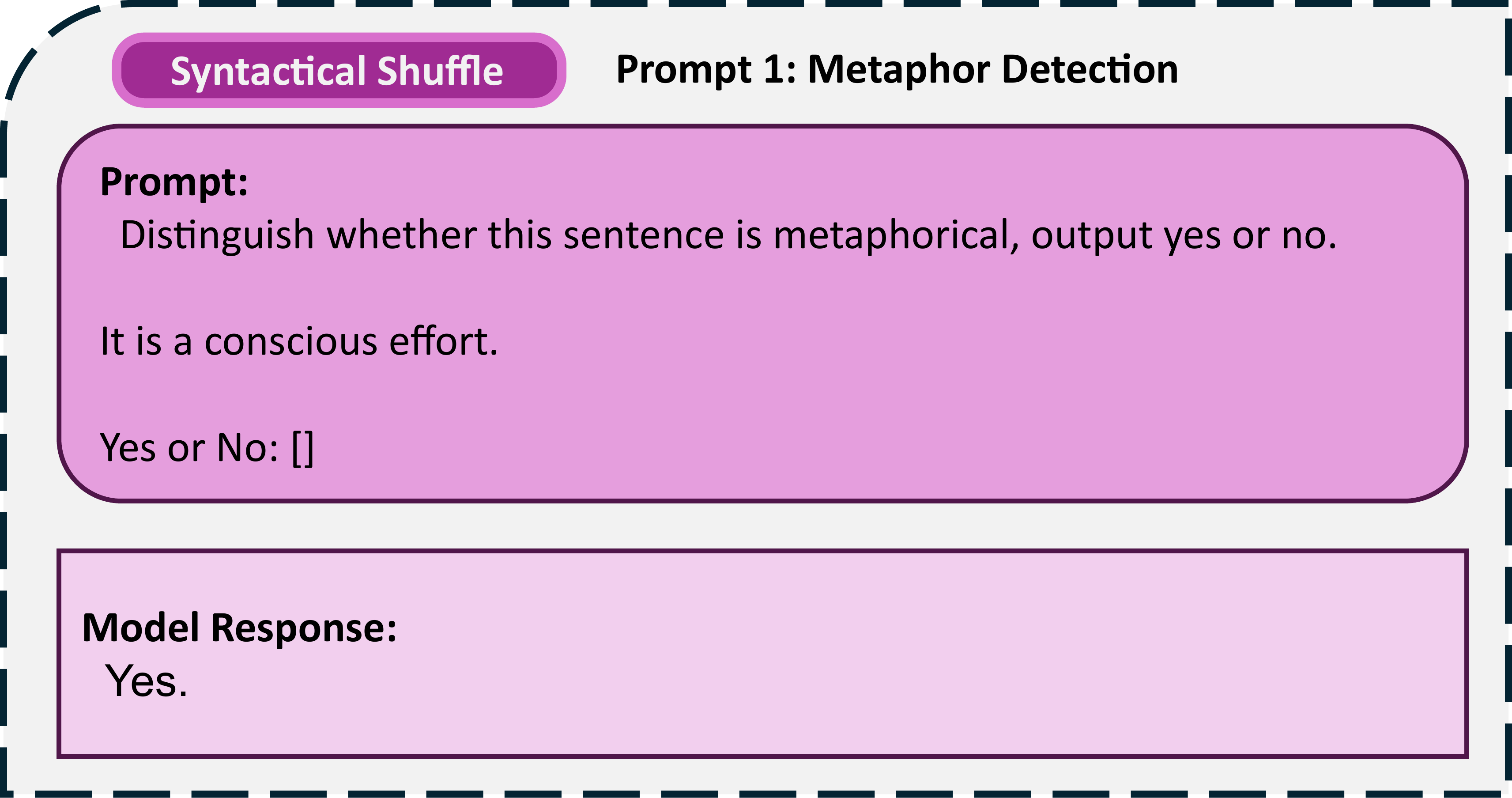}
  \caption{The prompt of Syntactic Shuffle.}
  \label{fig:prompts_shuffle}
\end{figure*}

Figure~\ref{fig:prompts_shuffle} presents the prompt used for metaphor detection under different syntactic perturbations in the Syntactic Shuffle experiment.

\section{Prompts for Metaphorical Imagination}\label{appendix:metaphorical_imagination}
\begin{figure*}[h]
  \centering
    \includegraphics[width=\columnwidth]{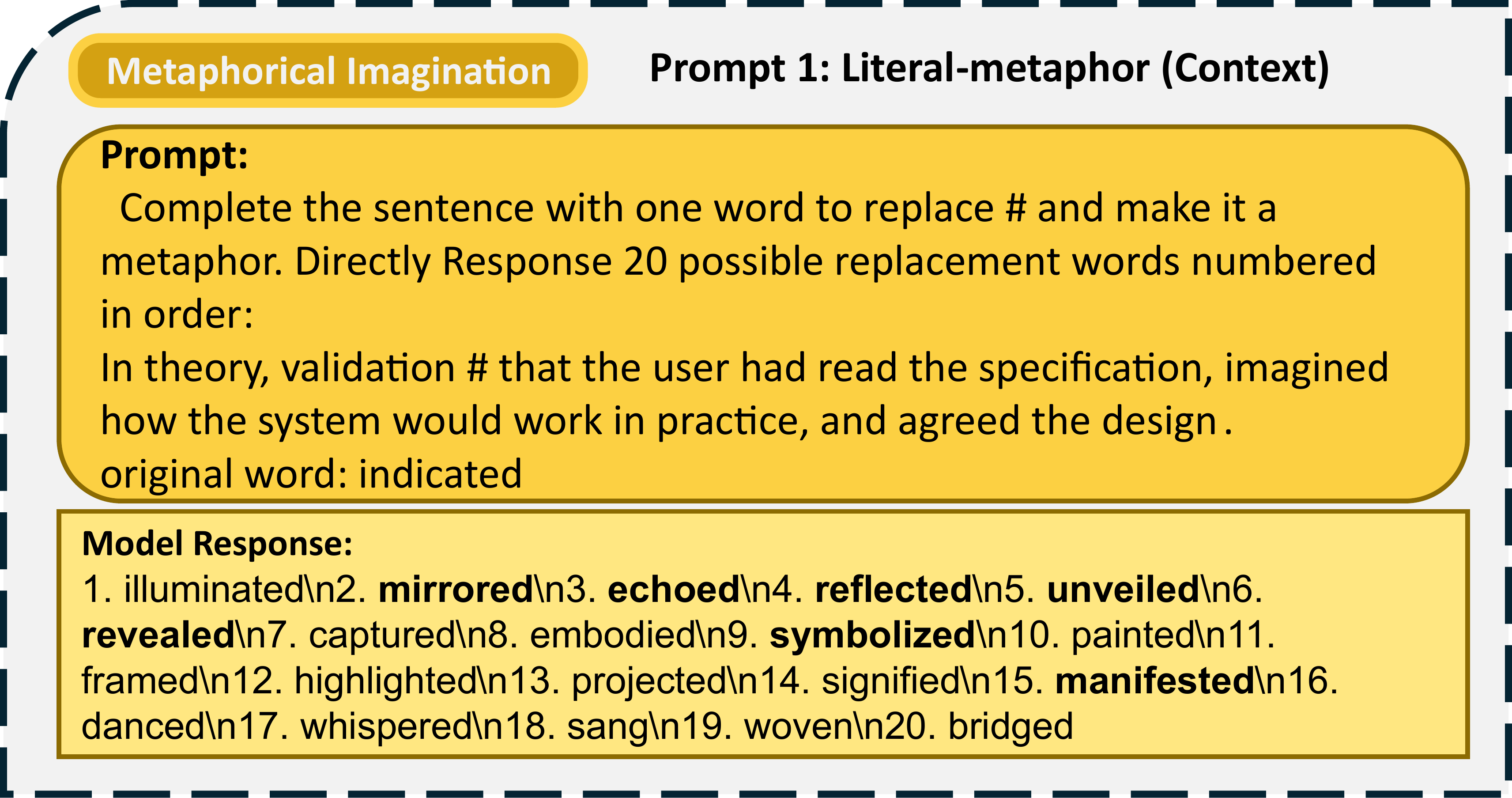}    \includegraphics[width=\columnwidth]{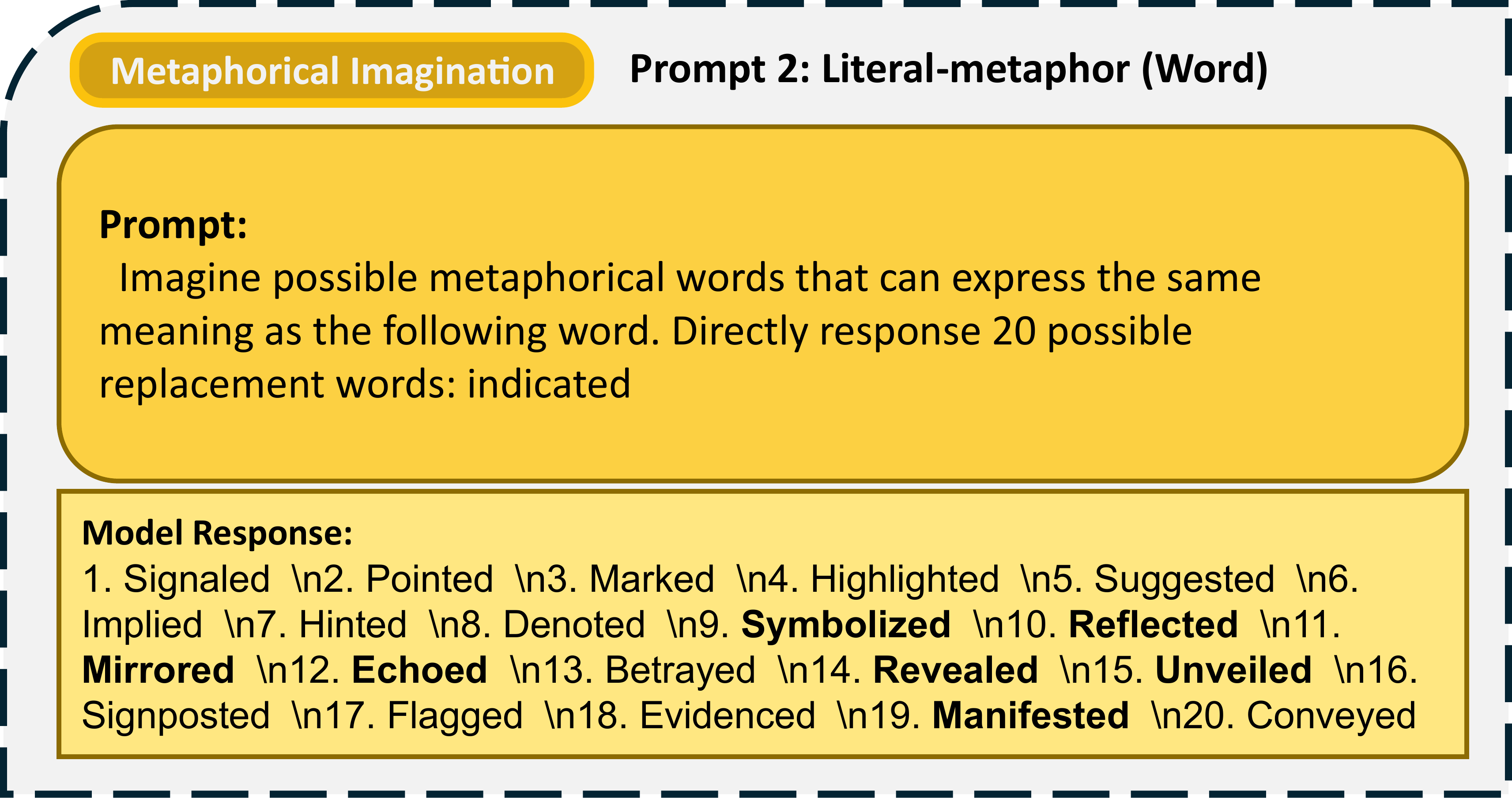}    \includegraphics[width=\columnwidth]{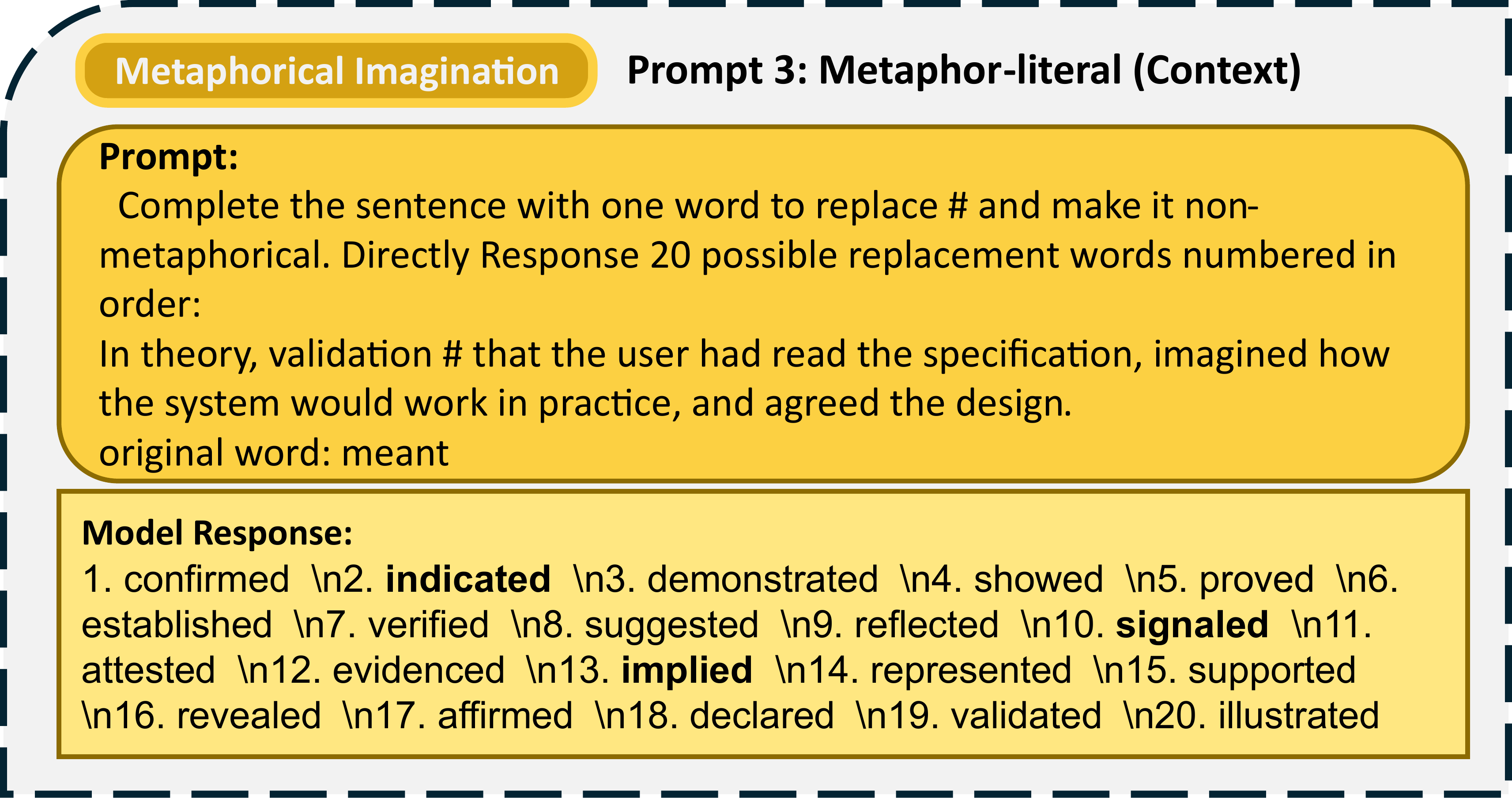}  \includegraphics[width=\columnwidth]{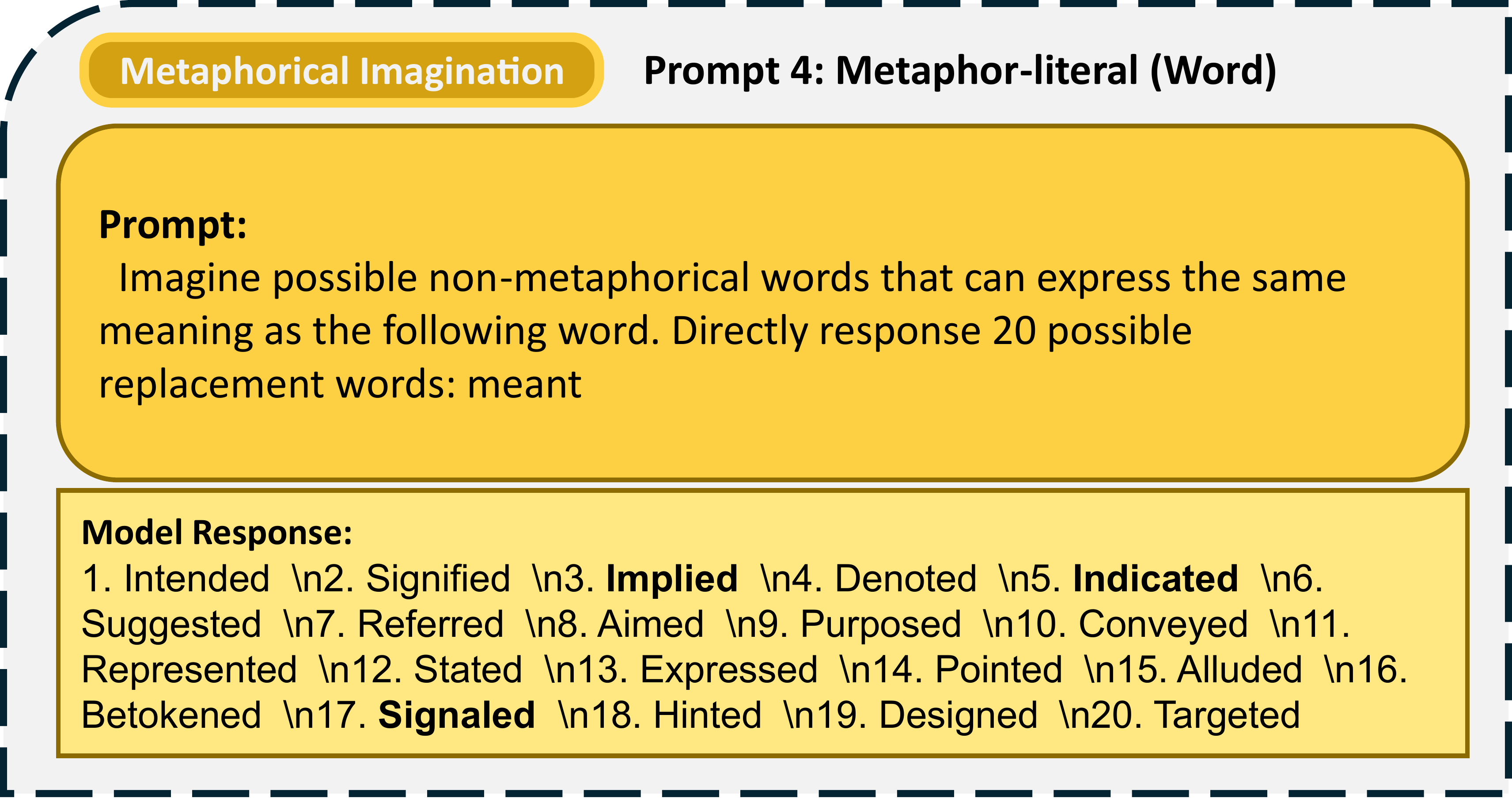}
  \caption{The prompts of Metaphorical Imagination.}
  \label{fig:prompts_imagination}
\end{figure*}

Figure~\ref{fig:prompts_imagination} presents the prompts used for the Metaphorical Imagination experiment. The prompts require LLMs to perform two complementary tasks: (1) \textit{Metaphor-to-Literal} (ML), in which literal words are generated from metaphorical inputs, and (2) \textit{Literal-to-Metaphor} (LM), in which metaphorical words are generated from literal inputs. Both tasks are conducted under contextualized and decontextualized settings. Accordingly, the prompts are divided into two types depending on whether sentence context is provided.

\section{Distributions of \((d_p, \cos\theta)\) and \((d_p, A_d)\)}\label{appendix:distribution_dp}
\begin{figure*}[h]
  \centering
  \includegraphics[width=0.5\columnwidth]{spatial_analysis/ds_ad.pdf}
      \includegraphics[width=0.5\columnwidth]{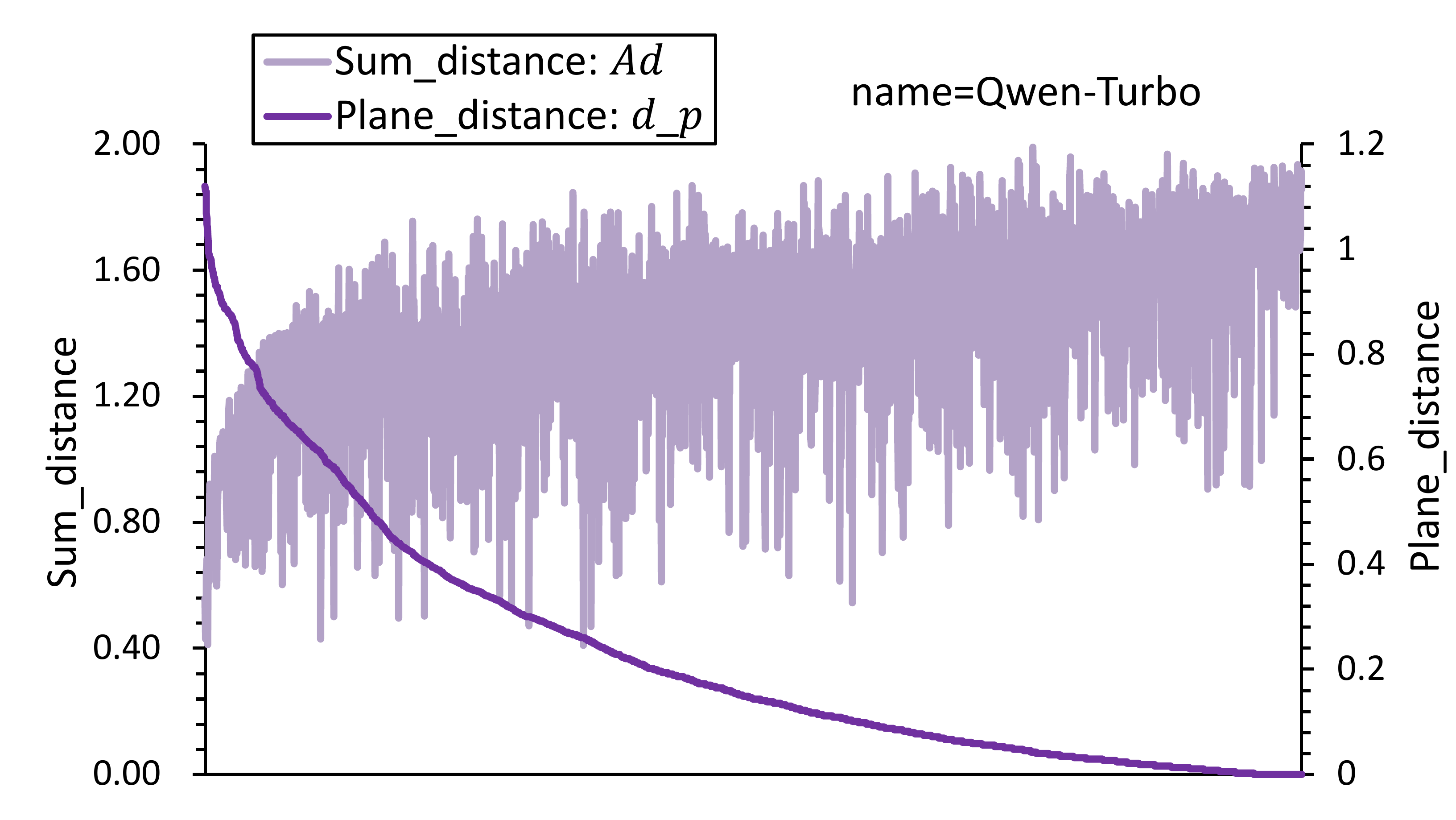}
            \includegraphics[width=0.5\columnwidth]{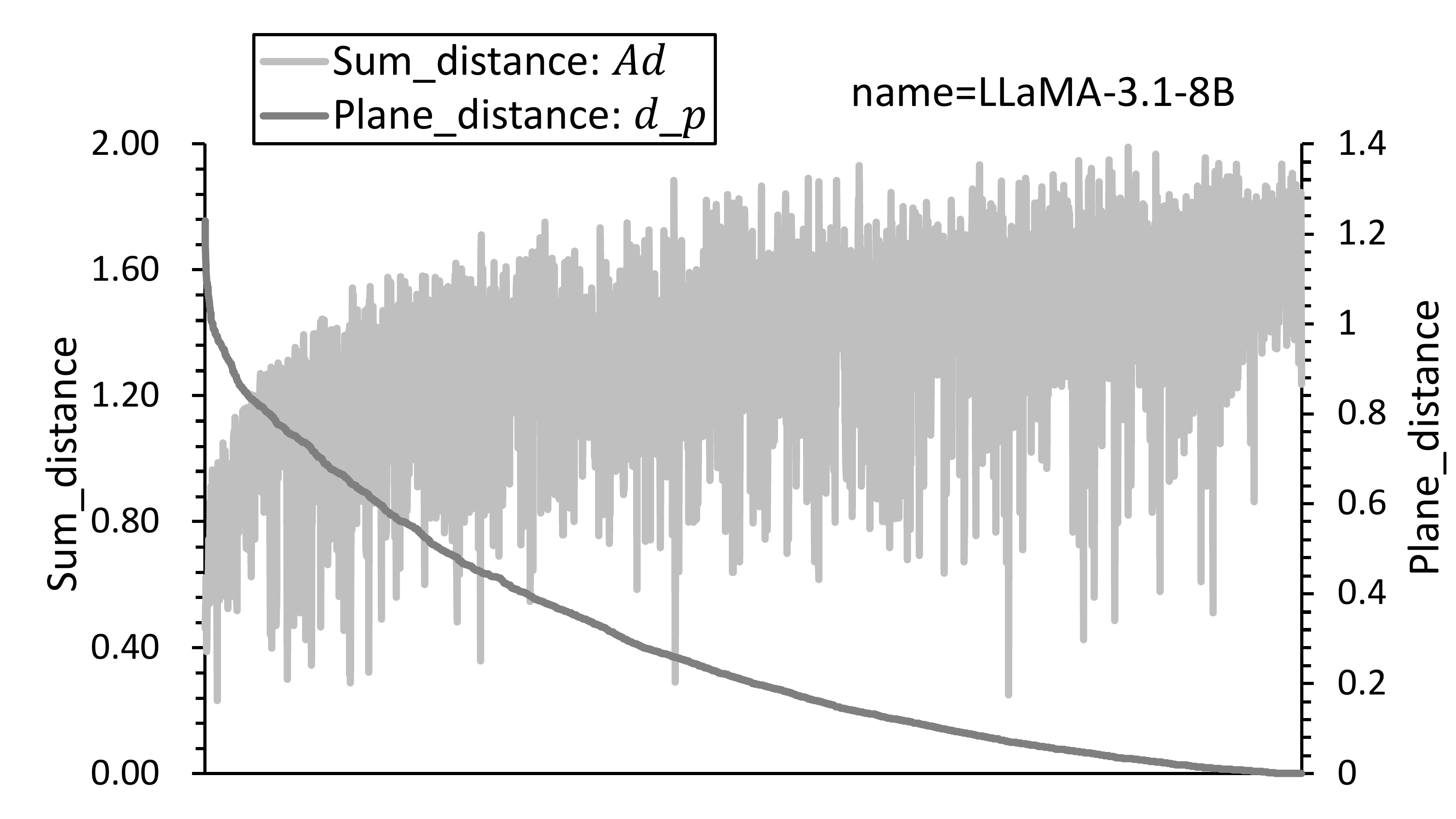}
                  \includegraphics[width=0.5\columnwidth]{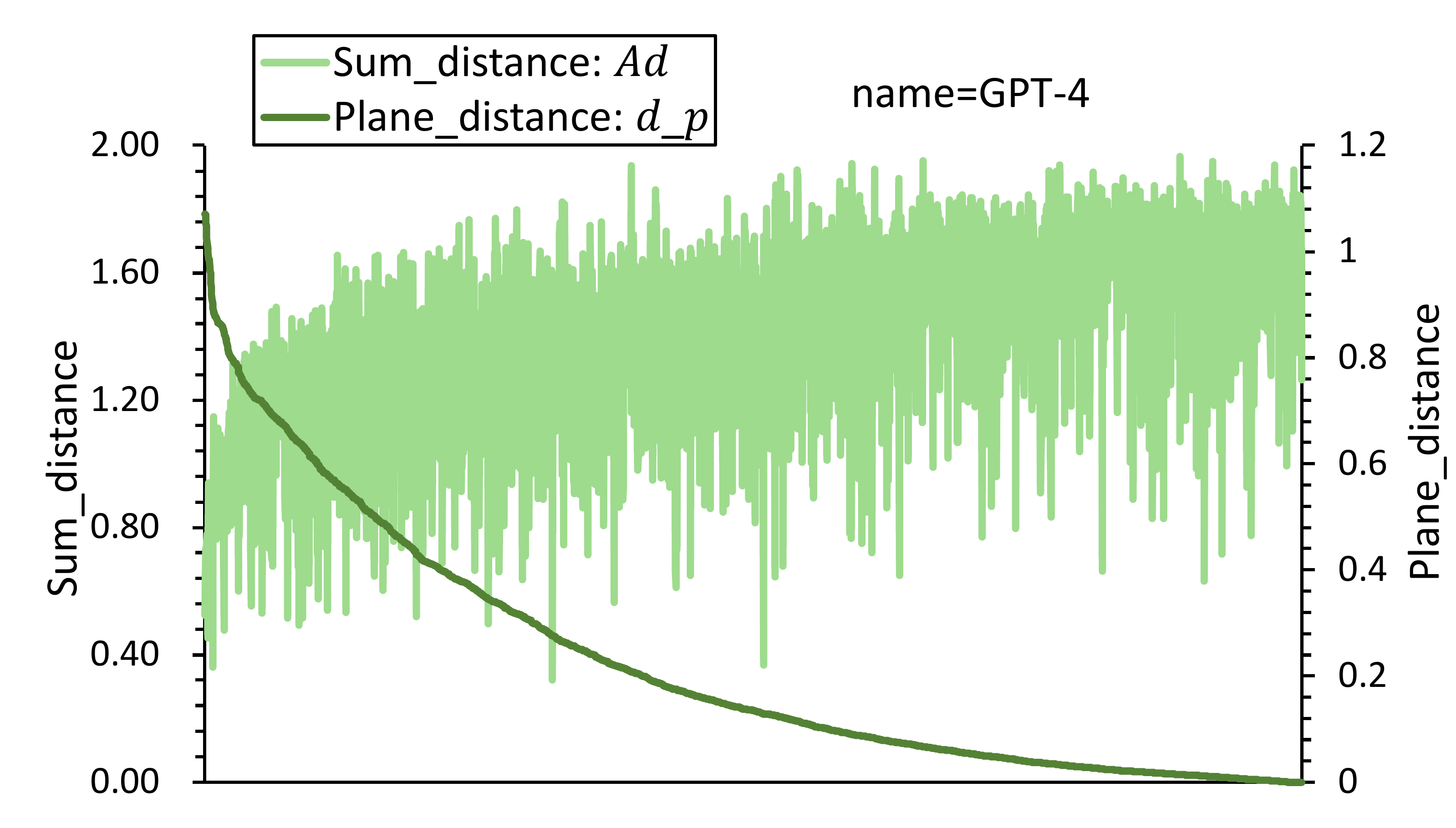}
                        \includegraphics[width=0.5\columnwidth]{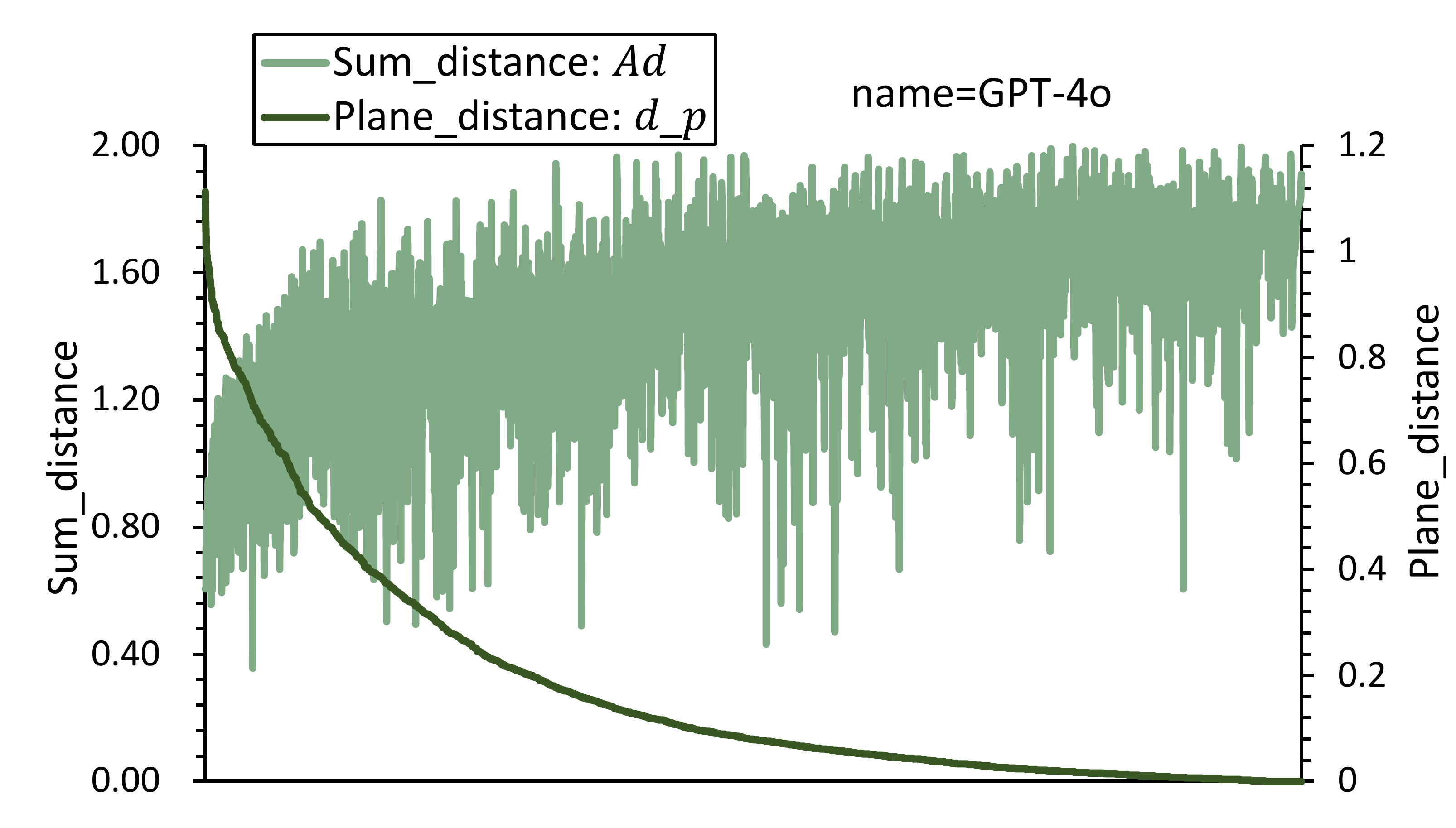}
                        \includegraphics[width=0.5\columnwidth]{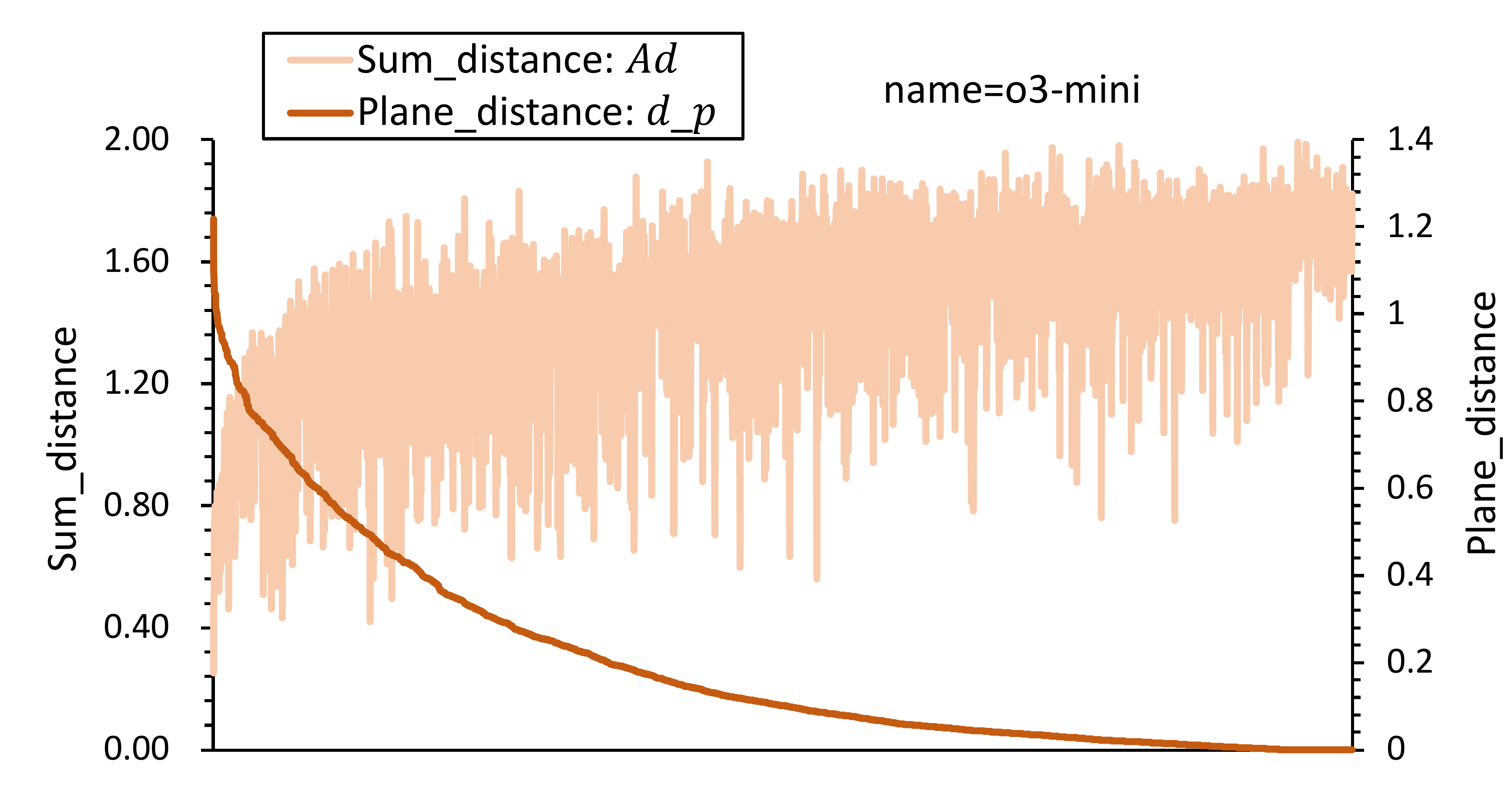}
                        \includegraphics[width=0.5\columnwidth]{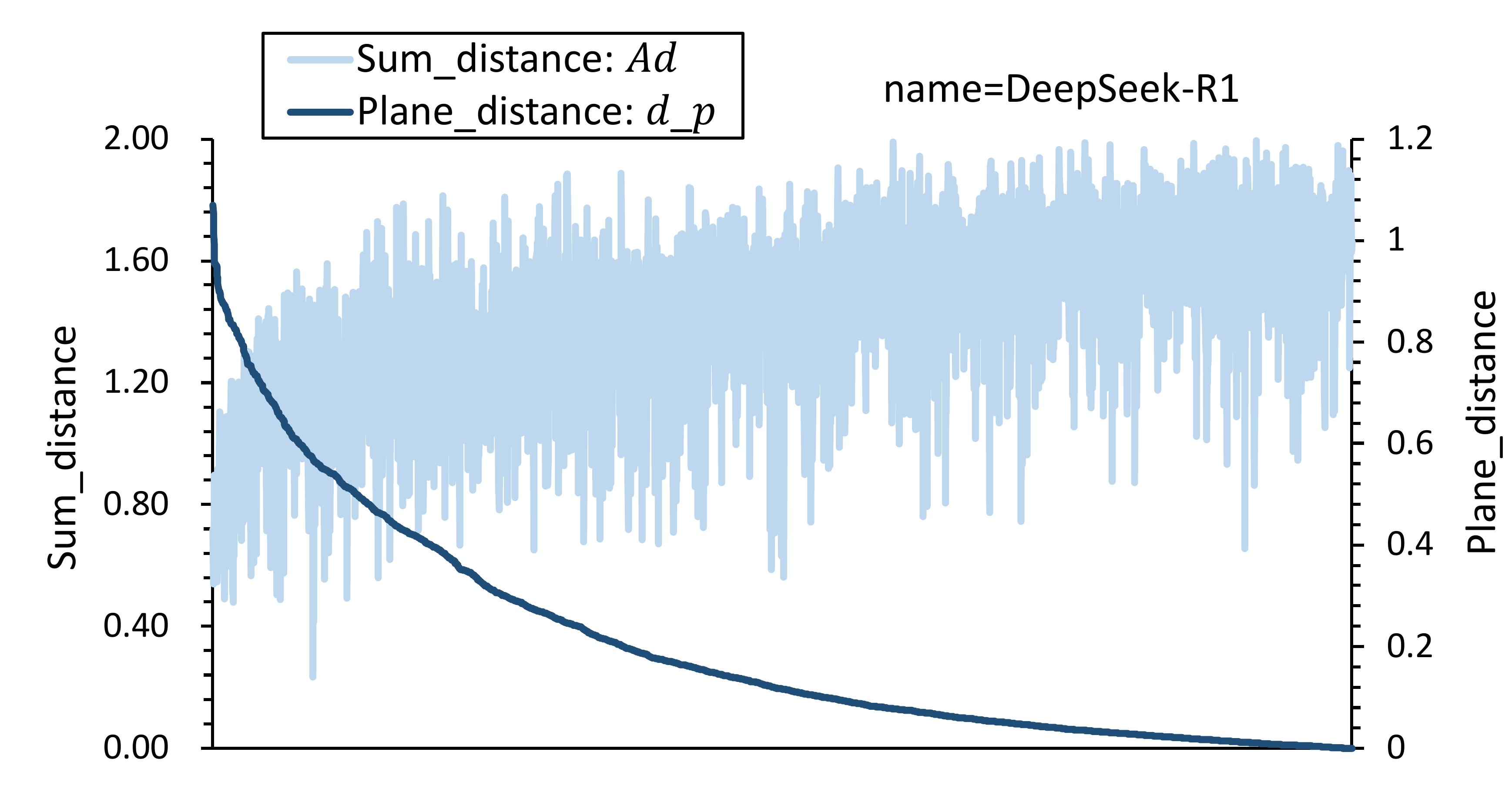}
  \caption{The (\(d_p\), $Ad$) distribution of every model. Sort \(d_p\) in decreasing order. Significant fluctuation can be observed due to the variance in the non-metaphorical part of sentences.}
  \label{fig:result_spatial2}
\end{figure*}

\begin{figure*}[h]
  \centering
  \includegraphics[width=0.5\columnwidth]{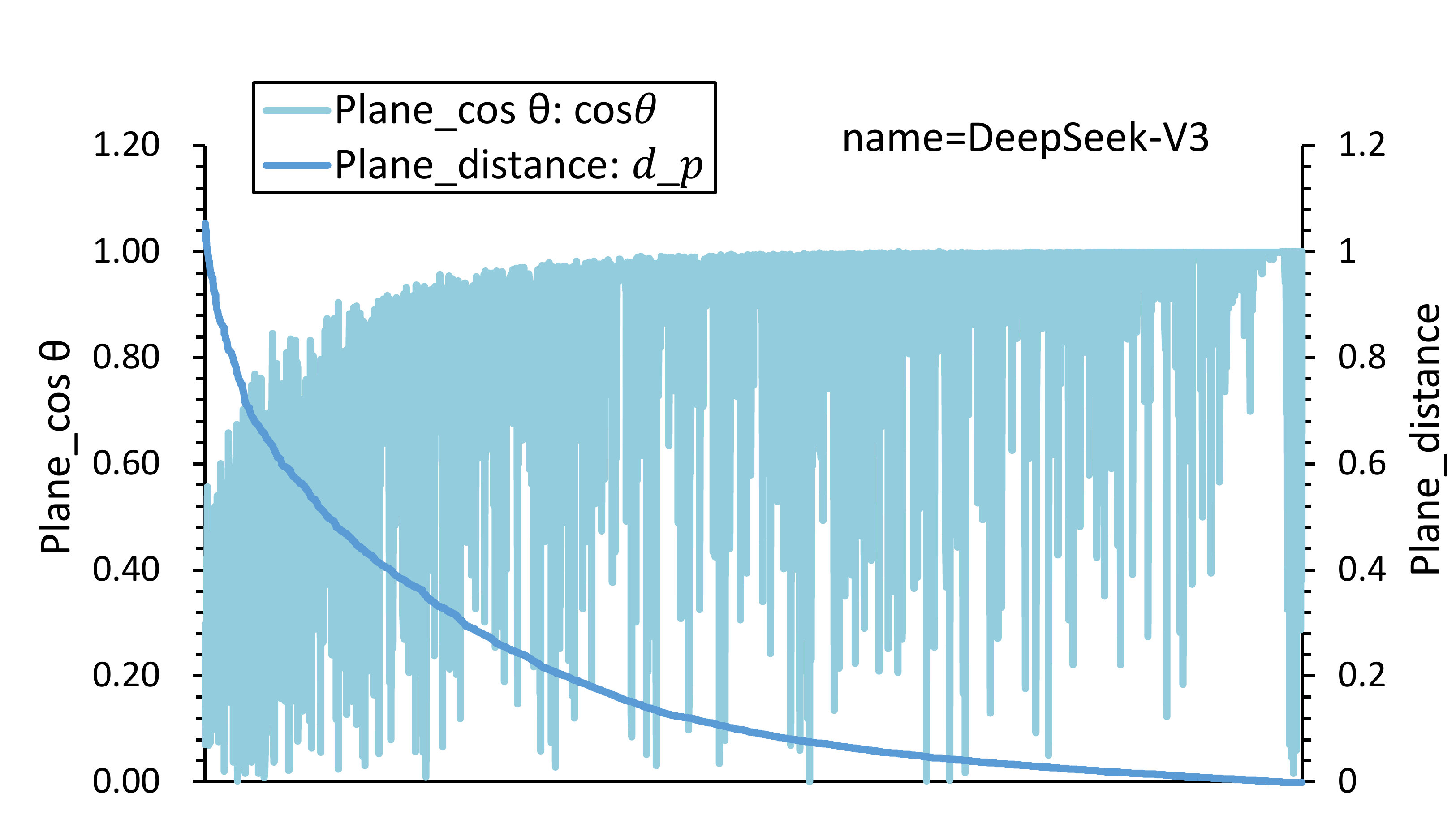}
      \includegraphics[width=0.5\columnwidth]{spatial_analysis/qwen_cos.pdf}
            \includegraphics[width=0.5\columnwidth]{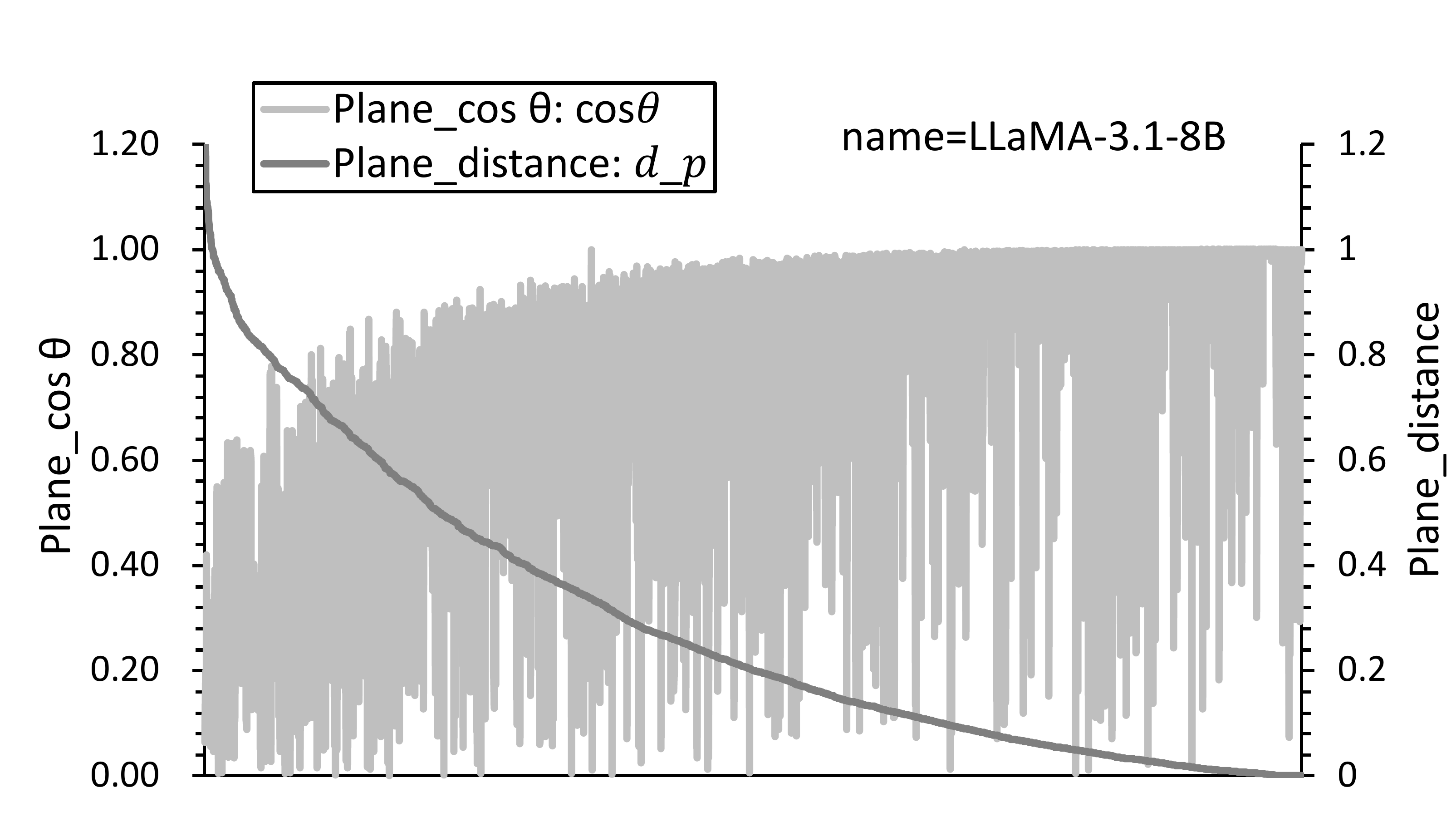}
                  \includegraphics[width=0.5\columnwidth]{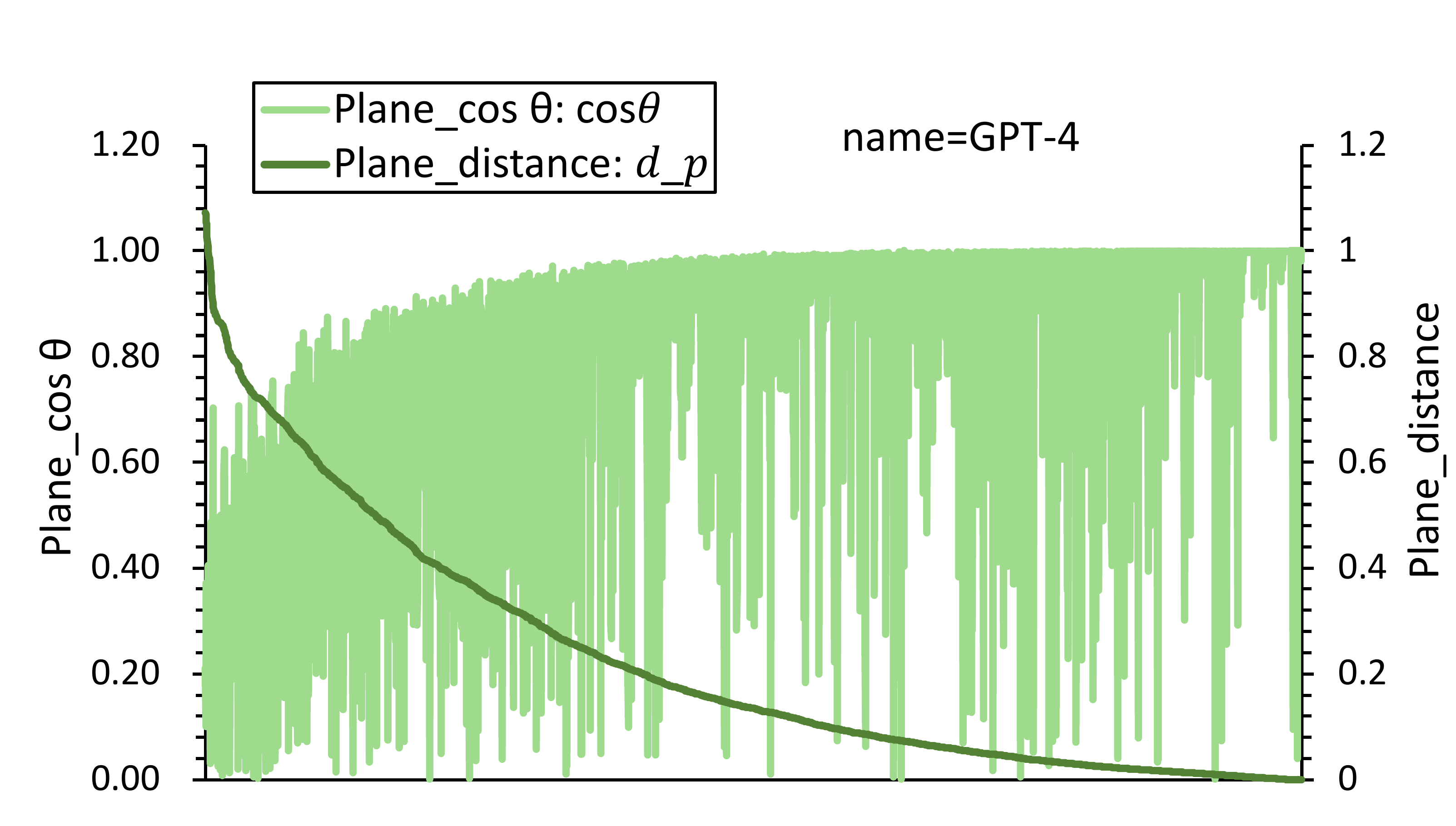}
                        \includegraphics[width=0.5\columnwidth]{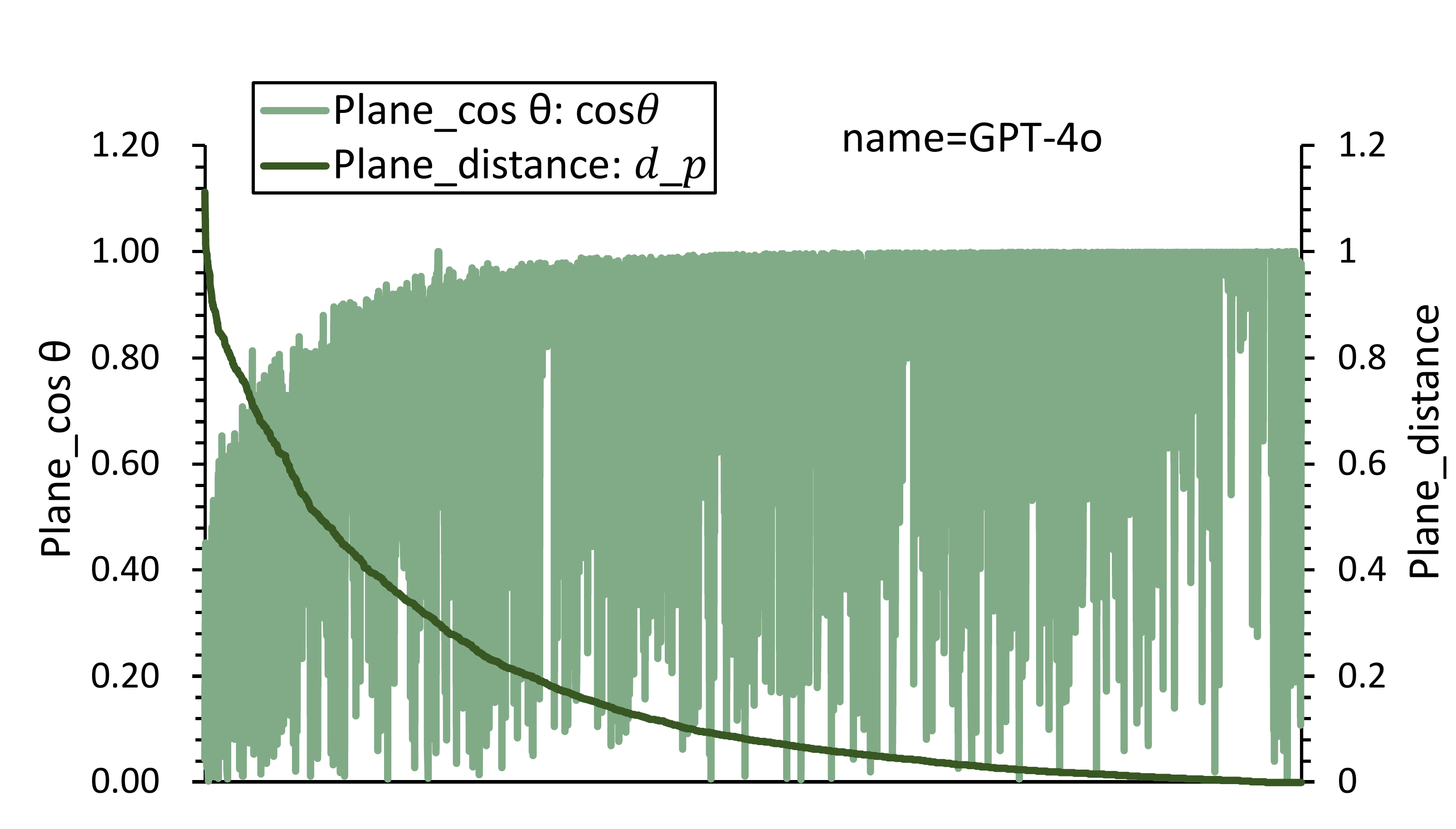}
                        \includegraphics[width=0.5\columnwidth]{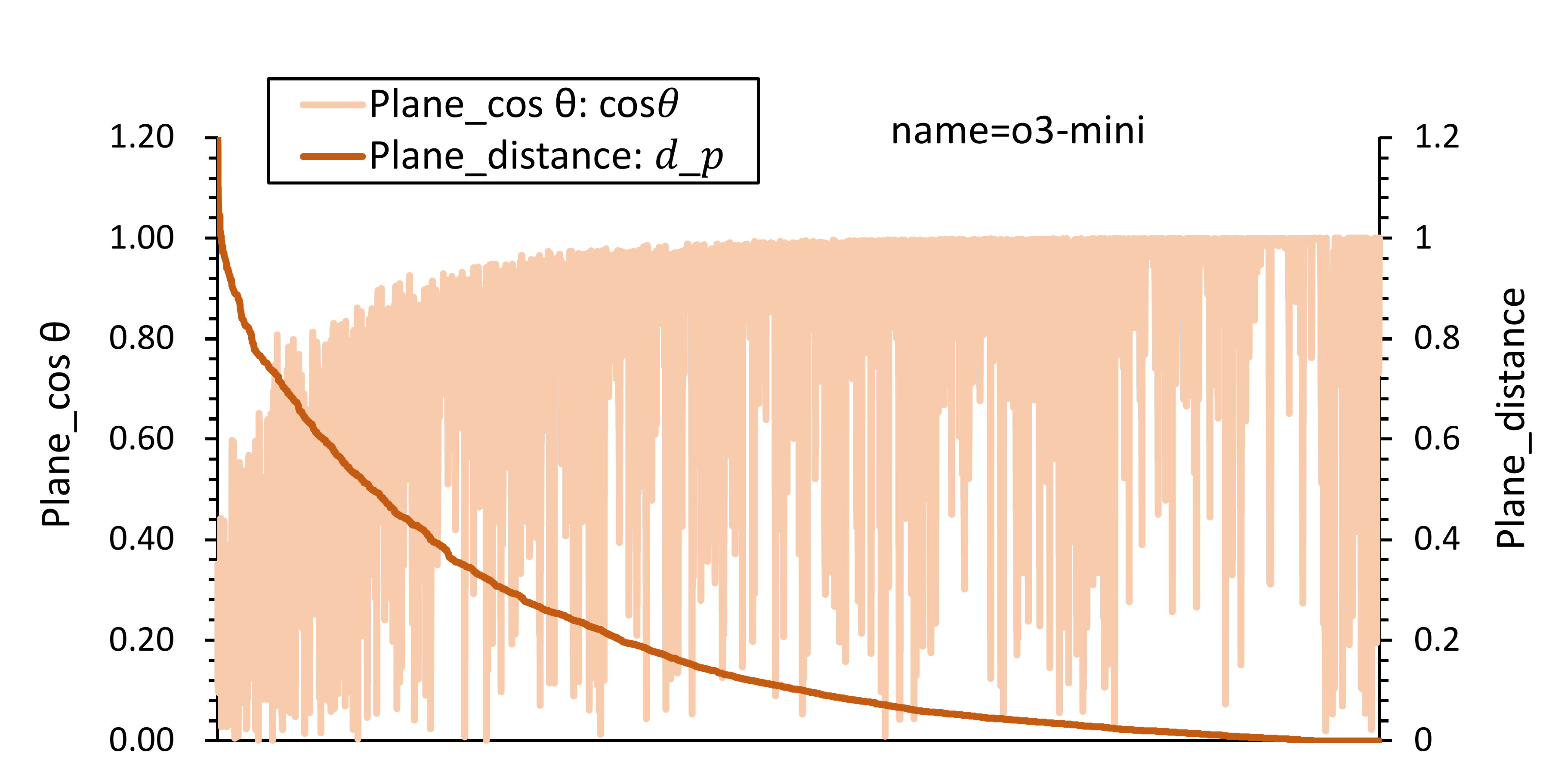}
                        \includegraphics[width=0.5\columnwidth]{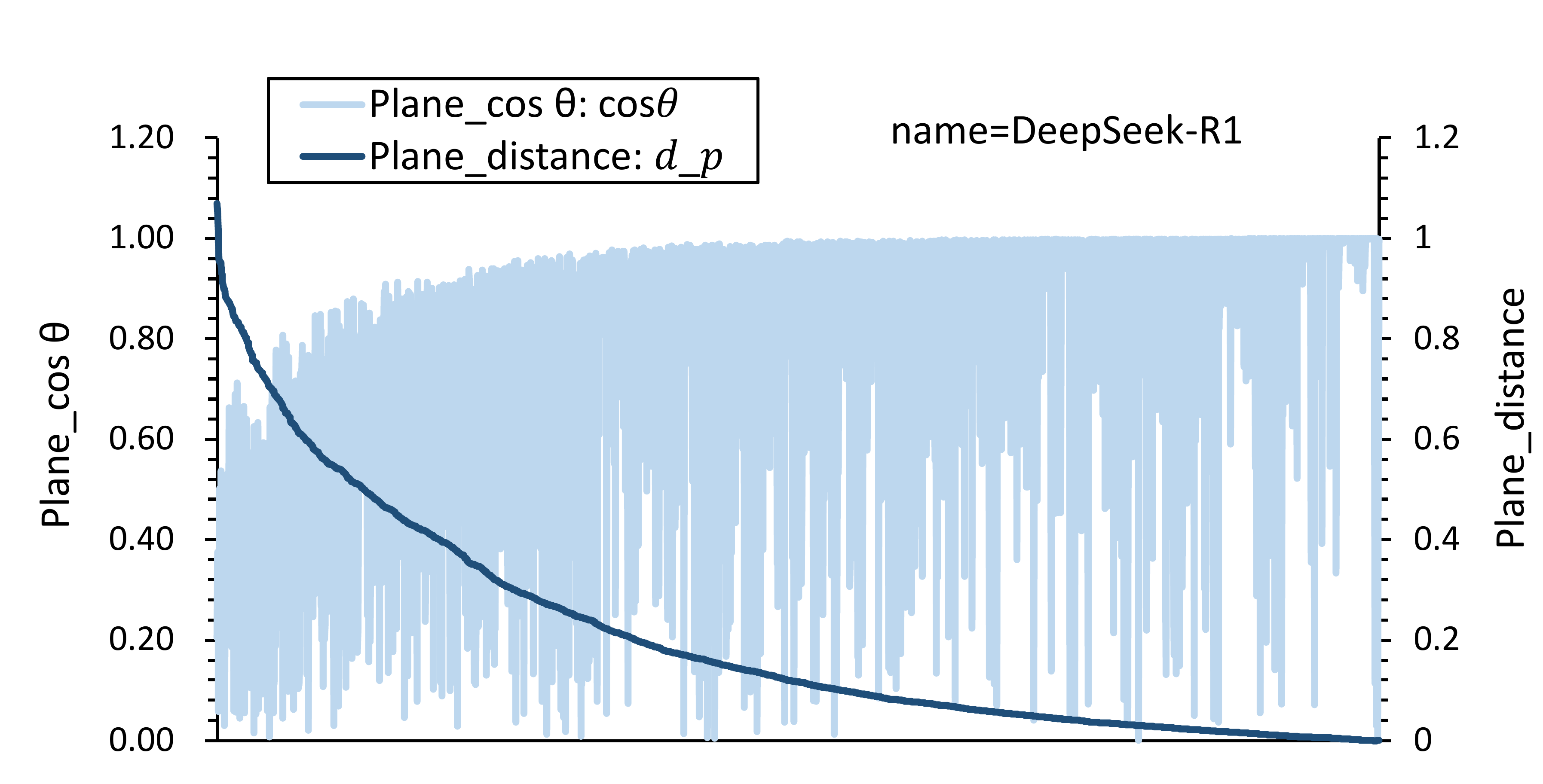}
  \caption{The (\(d_p\), $\cos\theta$)  distribution of every model. Sort \(d_p\) in decreasing order. Significant fluctuation can be observed due to the variance in the non-metaphorical part of sentences.}
  \label{fig:result_spatial3}
\end{figure*}

Figures~\ref{fig:result_spatial2}--\ref{fig:result_spatial3} present the distributions of the distance \(d_p\) (between the model-generated interpretation \(M_i\) and the reference plane \(\gamma_i\)), plotted against the cosine similarity \(\cos\theta\) (between the interpretation plane \(\beta_i\) and the reference plane \(\gamma_i\)), as well as against the auxiliary similarity measure \(A_d\).

\section{Sanity Check for the Literal Anchor \(S_i\)}
\label{appendix:si_sanity}

To assess whether the model-generated literal anchor \(S_i\) introduces unintended semantic bias, we analyze its alignment with the human reference interpretations \(\{R_i, R_i'\}\) in the shared embedding space.

\paragraph{Alignment with human anchors.}
We compute cosine similarities between \(S_i\) and each of the human references:
\[
\cos(S_i, R_i), \quad \cos(S_i, R_i').
\]
Across all instances (reported here for o3-mini), we observe that \(S_i\) remains consistently aligned with both anchors. Specifically, 
\(\cos(S_i, R_i)\) has a median of 0.753 (IQR: [0.662, 0.829]; 5–95\% range: [0.513, 0.922]), 
while \(\cos(S_i, R_i')\) has a median of 0.686 (IQR: [0.600, 0.767]; 5–95\% range: [0.467, 0.873]). 
For reference, the similarity between the two human interpretations \(\cos(R_i, R_i')\) has a median of 0.712 (IQR: [0.626, 0.779]).

These results indicate that \(S_i\) is typically located within the same semantic neighborhood as the human anchors, rather than drifting toward unrelated meanings.

\paragraph{Balance between the two anchors.}
To evaluate whether \(S_i\) is disproportionately closer to one reference than the other, we measure the absolute difference:
\[
|\Delta| = \big|\cos(S_i, R_i) - \cos(S_i, R_i')\big|.
\]
We find that this imbalance is generally moderate, with a median of 0.140 (IQR: [0.086, 0.192]) and a 95th percentile of 0.300. 
This suggests that \(S_i\) does not strongly bias the reference plane toward either anchor in most cases.

\paragraph{Interpretation.}
Overall, these diagnostics support the use of \(S_i\) as a stable literal anchor for constructing the local reference plane. 
While \(S_i\) is model-generated, it typically remains semantically aligned with the intended attribute encoded by \(\{R_i, R_i'\}\), and does not introduce systematic off-topic artifacts. 
Accordingly, the resulting geometric measures are best interpreted as capturing relative semantic deviation within a locally consistent reference region.

\section{Human Evaluation of Semantic Alignment}
\label{appendix:human_eval}

To assess whether the geometric deviation measure \(d_p\) reflects meaningful semantic differences in metaphor interpretation, we conduct a small-scale human evaluation.

\paragraph{Setup.}
Each evaluation instance consists of (i) an English metaphorical sentence and (ii) a corresponding model-generated interpretation. The annotator is asked to judge whether the interpretation captures the intended metaphorical meaning based solely on the given text, without external context.

We use a 3-point rubric:
\begin{itemize}
    \item \textbf{2 (Correct):} The interpretation captures the intended abstract meaning or mapping.
    \item \textbf{1 (Partial):} The interpretation is related but underspecified, ambiguous, or only partially aligned.
    \item \textbf{0 (Incorrect):} The interpretation is off-topic, incorrect, or reduces the metaphor to a literal or unrelated description.
\end{itemize}

\paragraph{Sampling.}
To test whether \(d_p\) meaningfully separates aligned and misaligned interpretations, we sort instances by \(d_p\) and sample from both extremes. Specifically, we select 25 instances from the lowest-\(d_p\) region (bottom-200) and 25 instances from the highest-\(d_p\) region (top-200), based on o3-mini generations.

\paragraph{Results.}
We observe a clear separation between the two groups. For high-\(d_p\) instances, the label distribution is \(11/7/7\) for scores \(0/1/2\), with a mean score of 0.84. In contrast, low-\(d_p\) instances yield \(0/1/24\), with a mean score of 1.96. The difference between the two groups is substantial (\(\Delta = 1.12\)).

To assess statistical reliability, we perform bootstrap resampling and obtain a 95\% confidence interval of \([0.76, 1.44]\) for the mean difference. A Mann--Whitney U test further indicates that the difference is statistically significant (\(p \approx 1.05 \times 10^{-6}\)).

\paragraph{Interpretation.}
These results provide independent evidence that lower geometric deviation (\(d_p\)) corresponds to better semantic alignment as judged by humans. While the evaluation is limited in scale, it supports the interpretation of \(d_p\) as a meaningful proxy for relative semantic alignment in metaphor interpretation.

\section{Anchor Score Distributions in Metaphorical Imagination}\label{appendix:overlap_ratio}
\begin{figure*}[h]
  \centering
  \includegraphics[width=0.5\columnwidth]{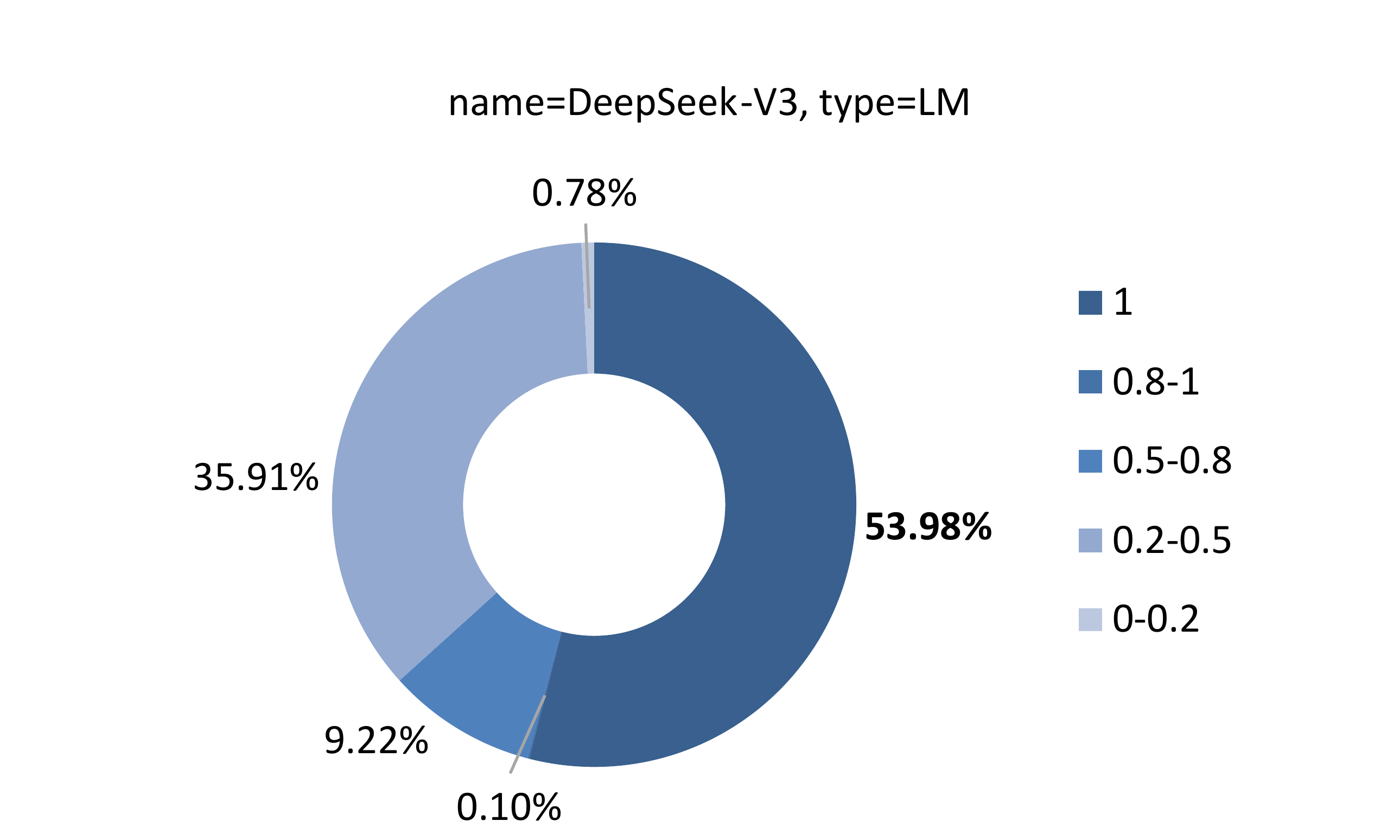}
      \includegraphics[width=0.5\columnwidth]{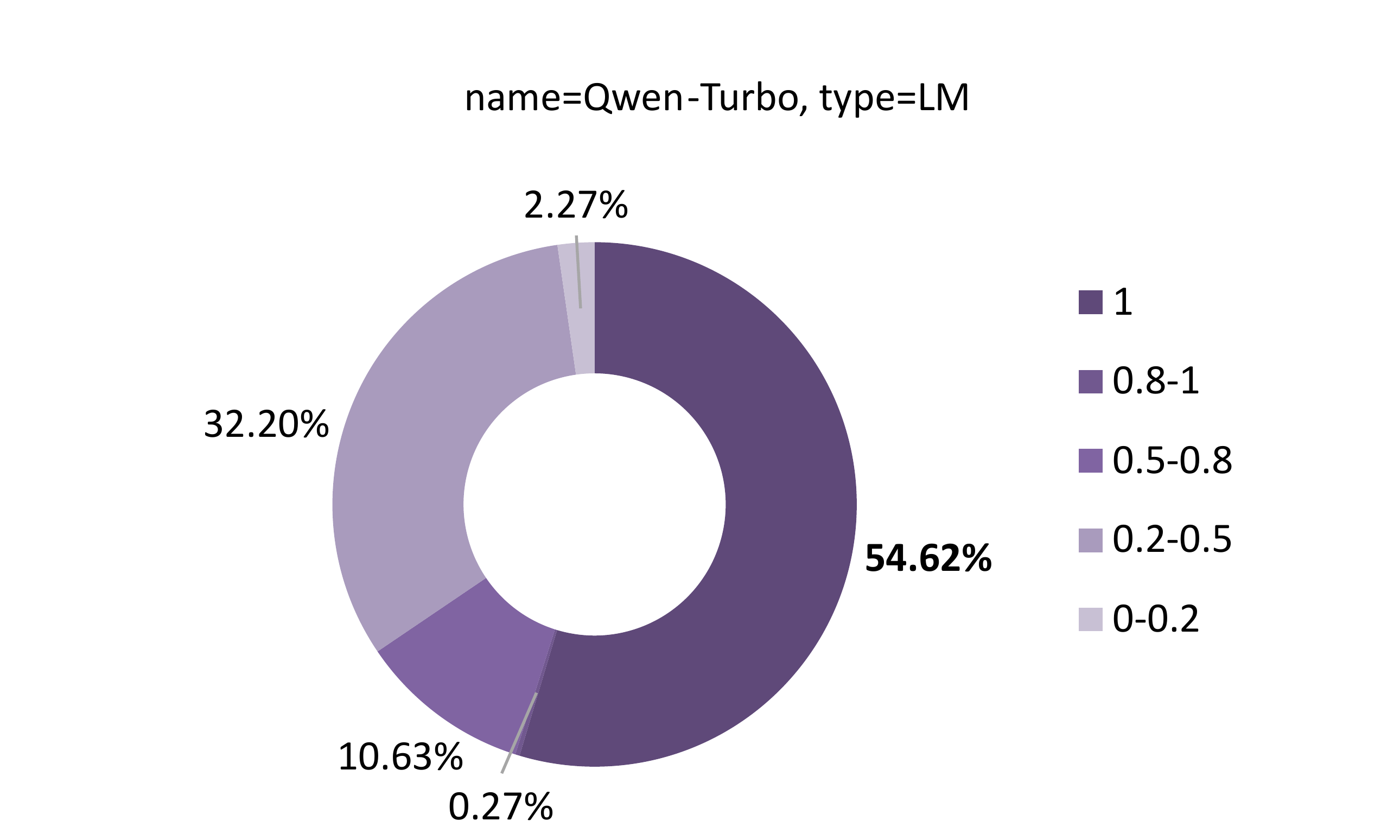}
            \includegraphics[width=0.5\columnwidth]{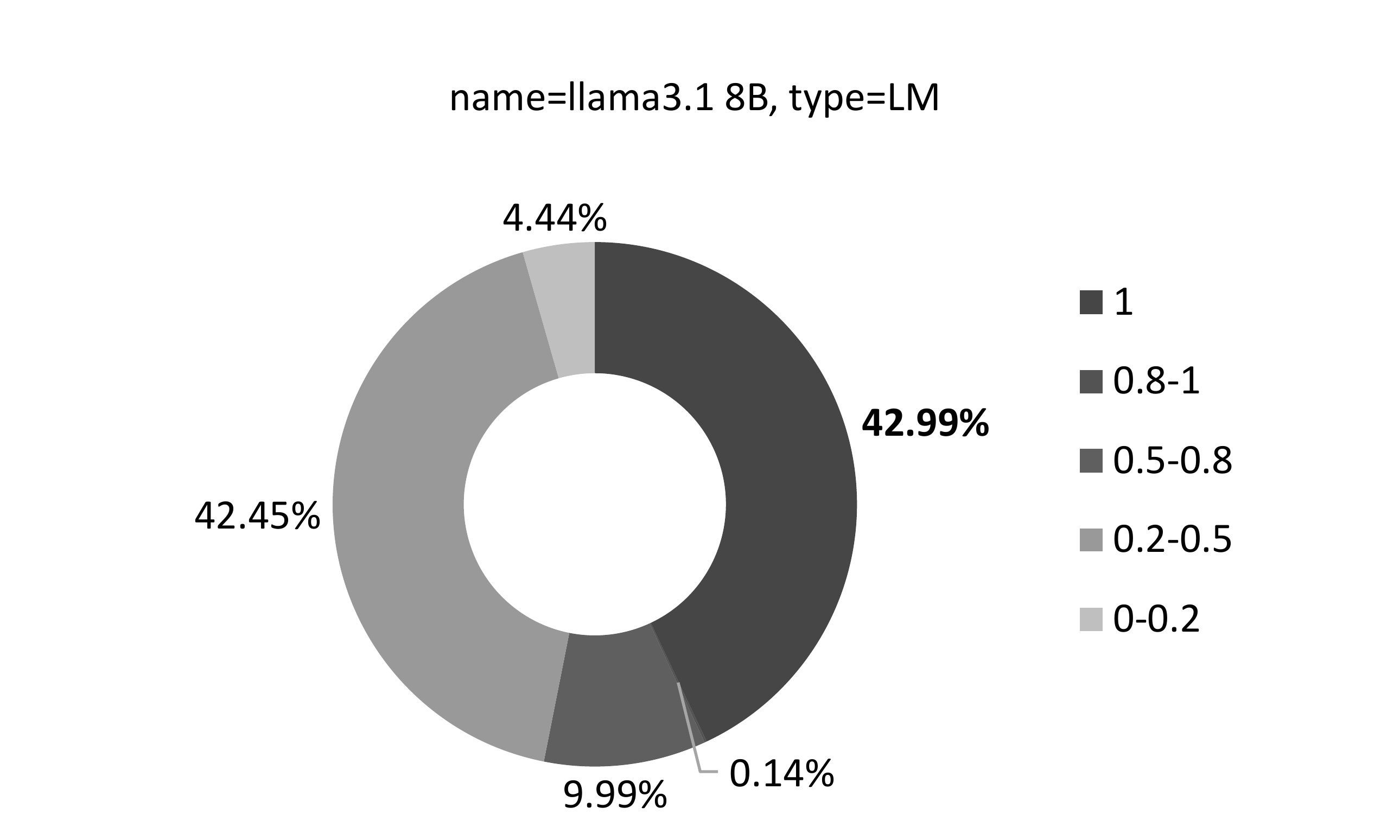}
                  \includegraphics[width=0.5\columnwidth]{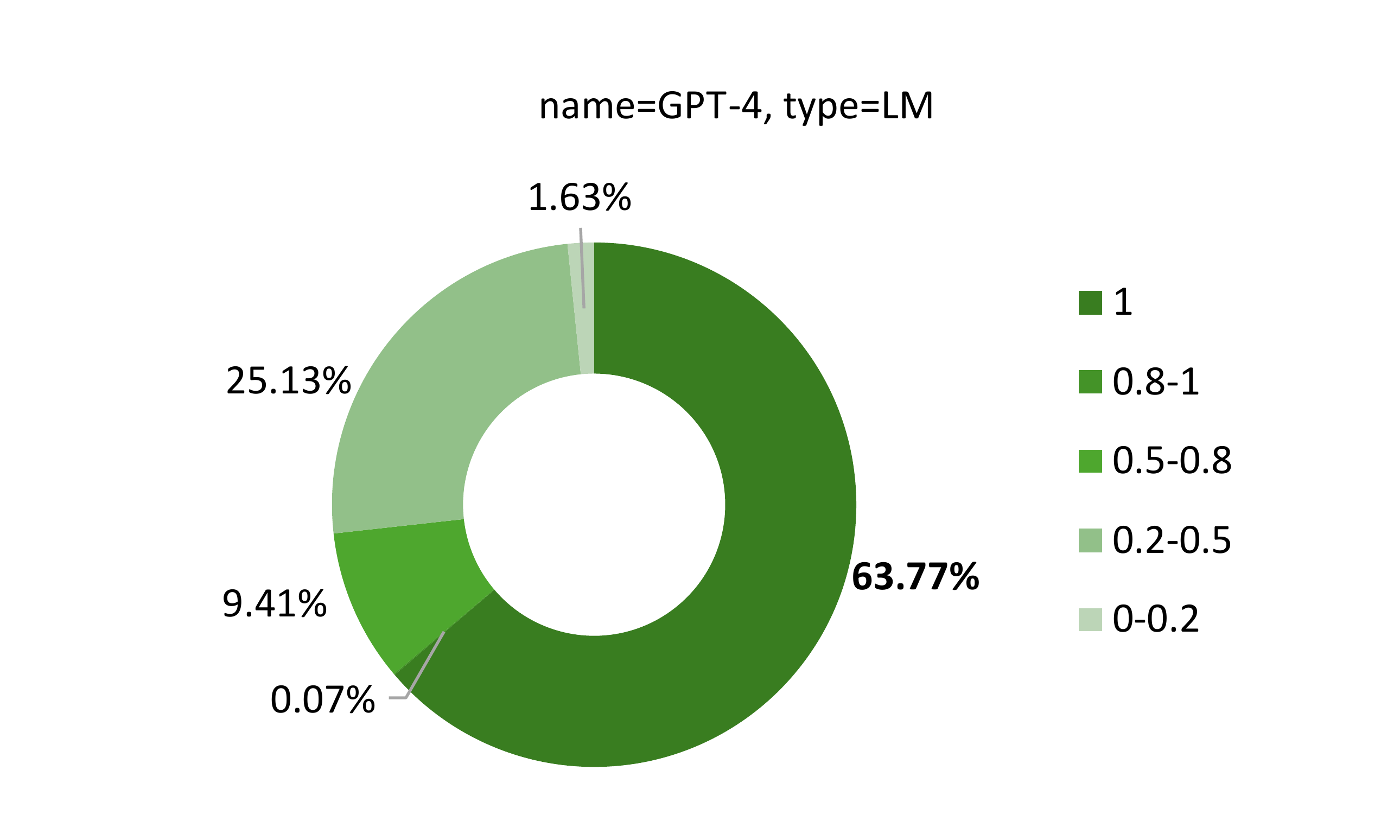}
                        \includegraphics[width=0.5\columnwidth]{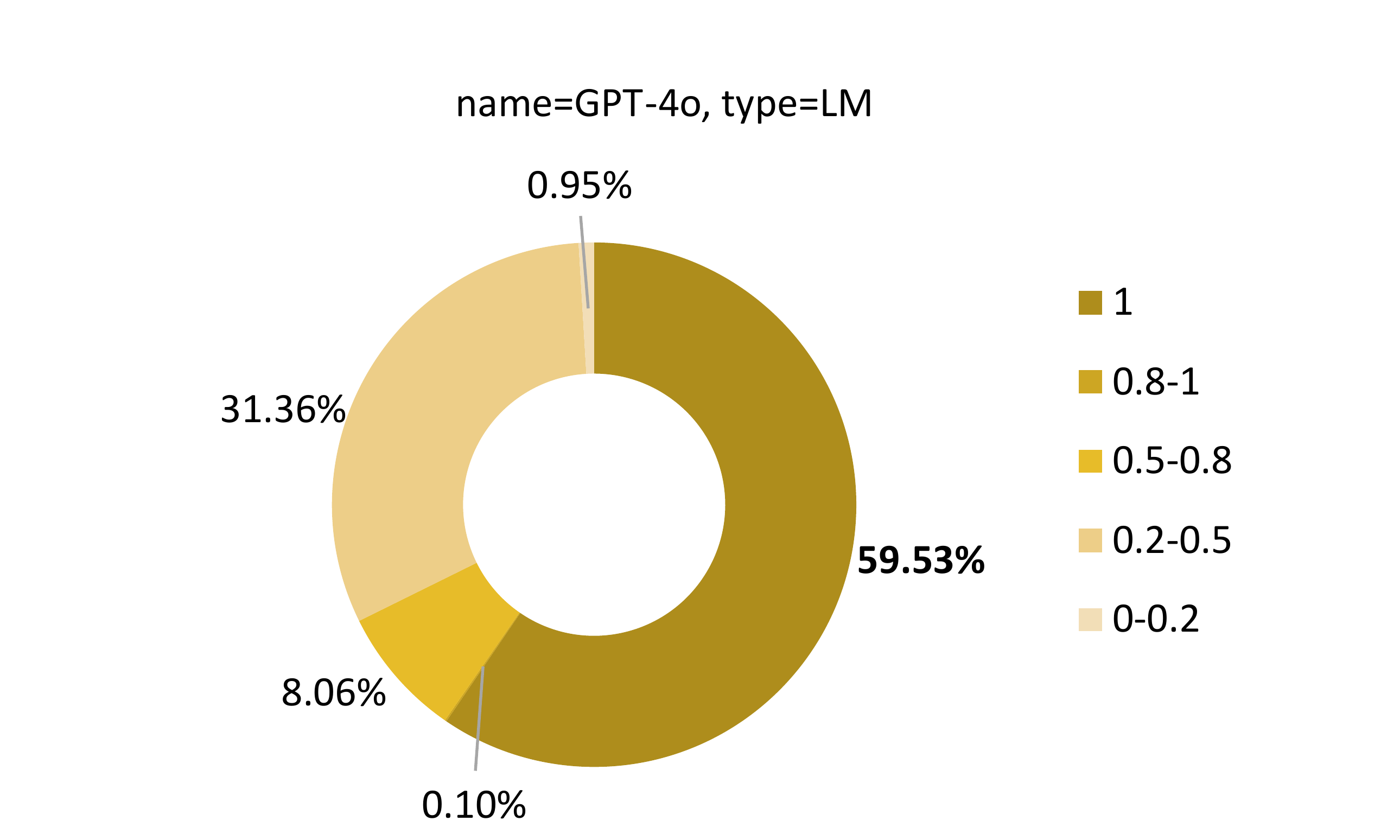}
                        \includegraphics[width=0.5\columnwidth]{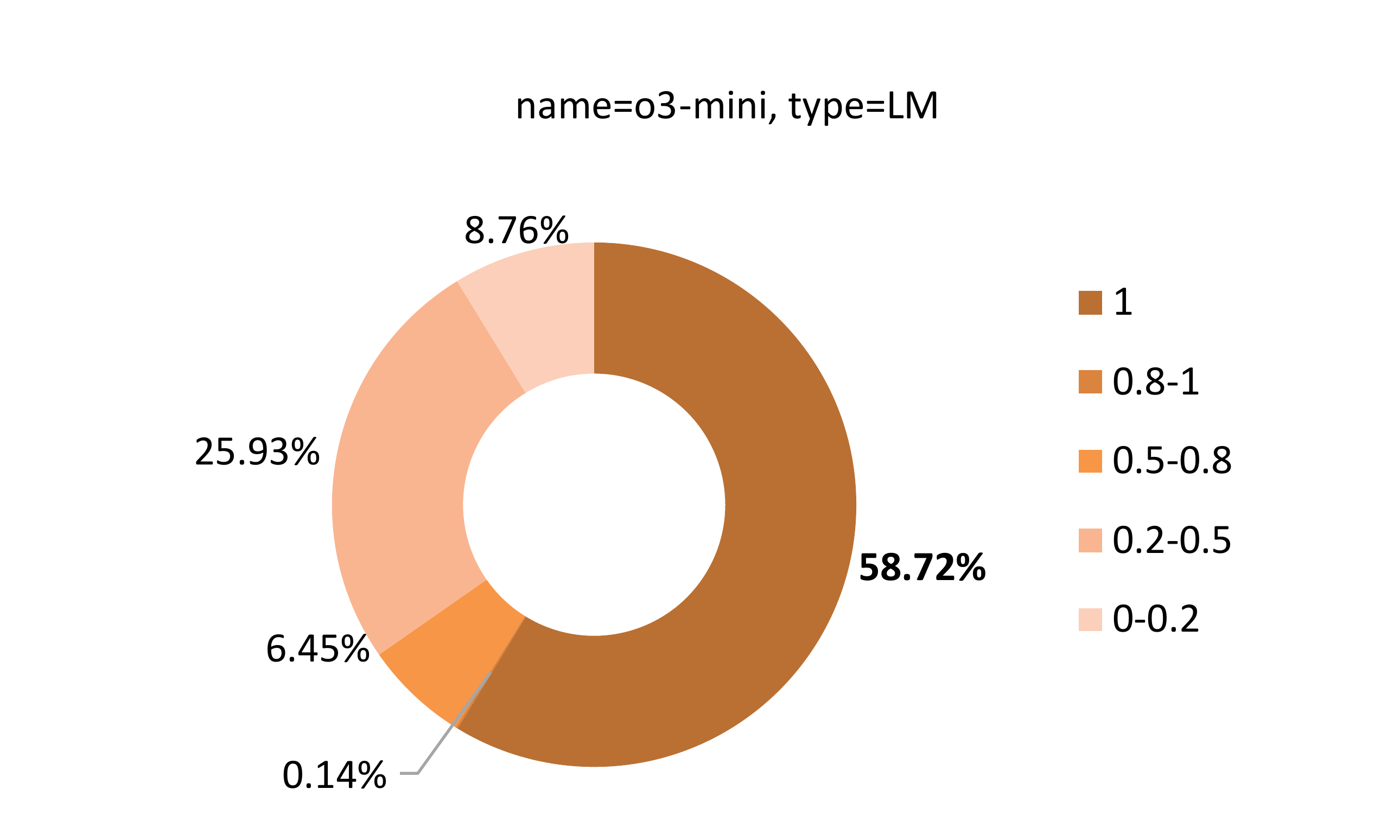}
                        \includegraphics[width=0.5\columnwidth]{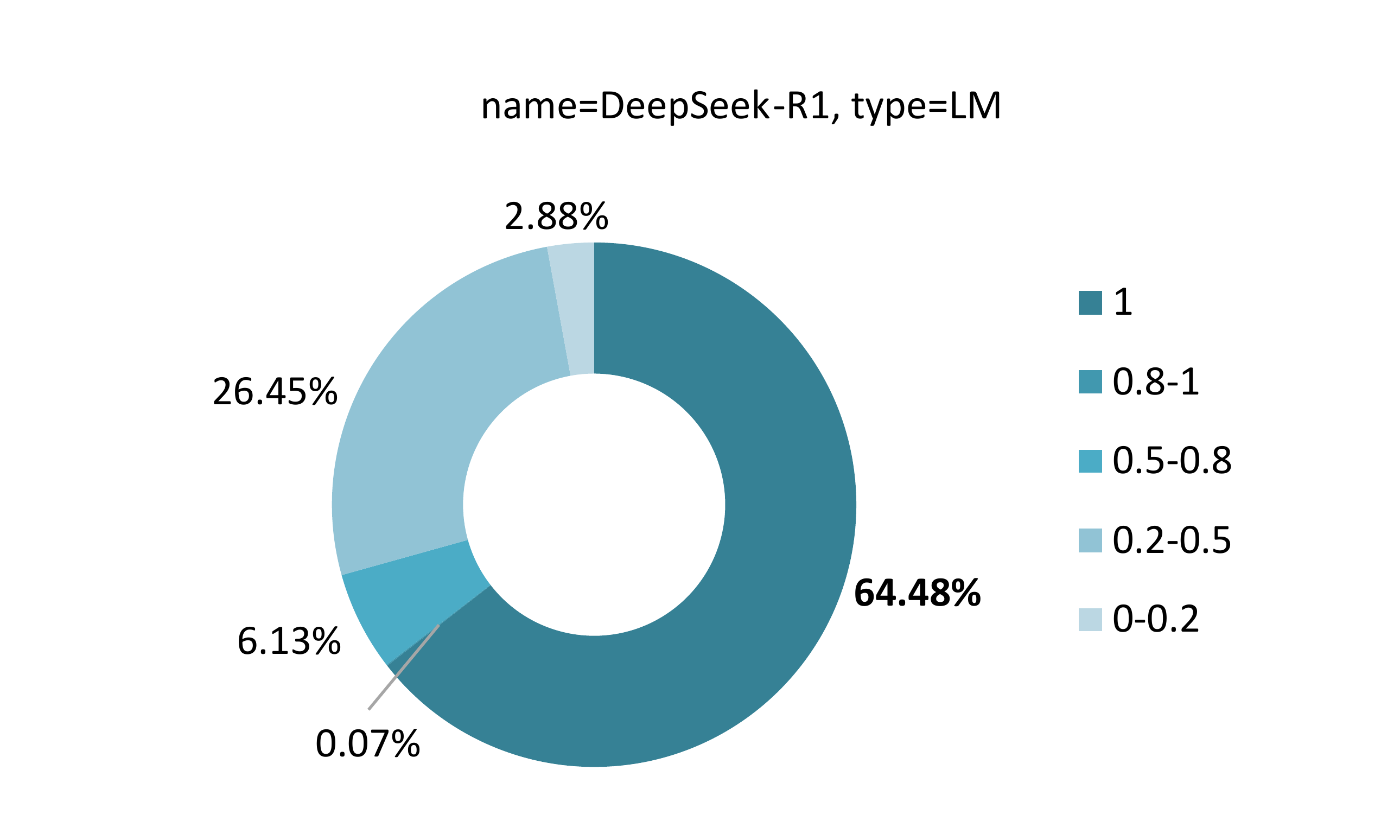}
  \caption{The Anchor Scores distributions of literal-metaphor (LM) word imagination task on every model (the largest portion is in bold and the second largest is underlined).}
  \label{fig:result_imagination2}
\end{figure*}

\begin{figure*}[h]
  \centering
  \includegraphics[width=0.5\columnwidth]{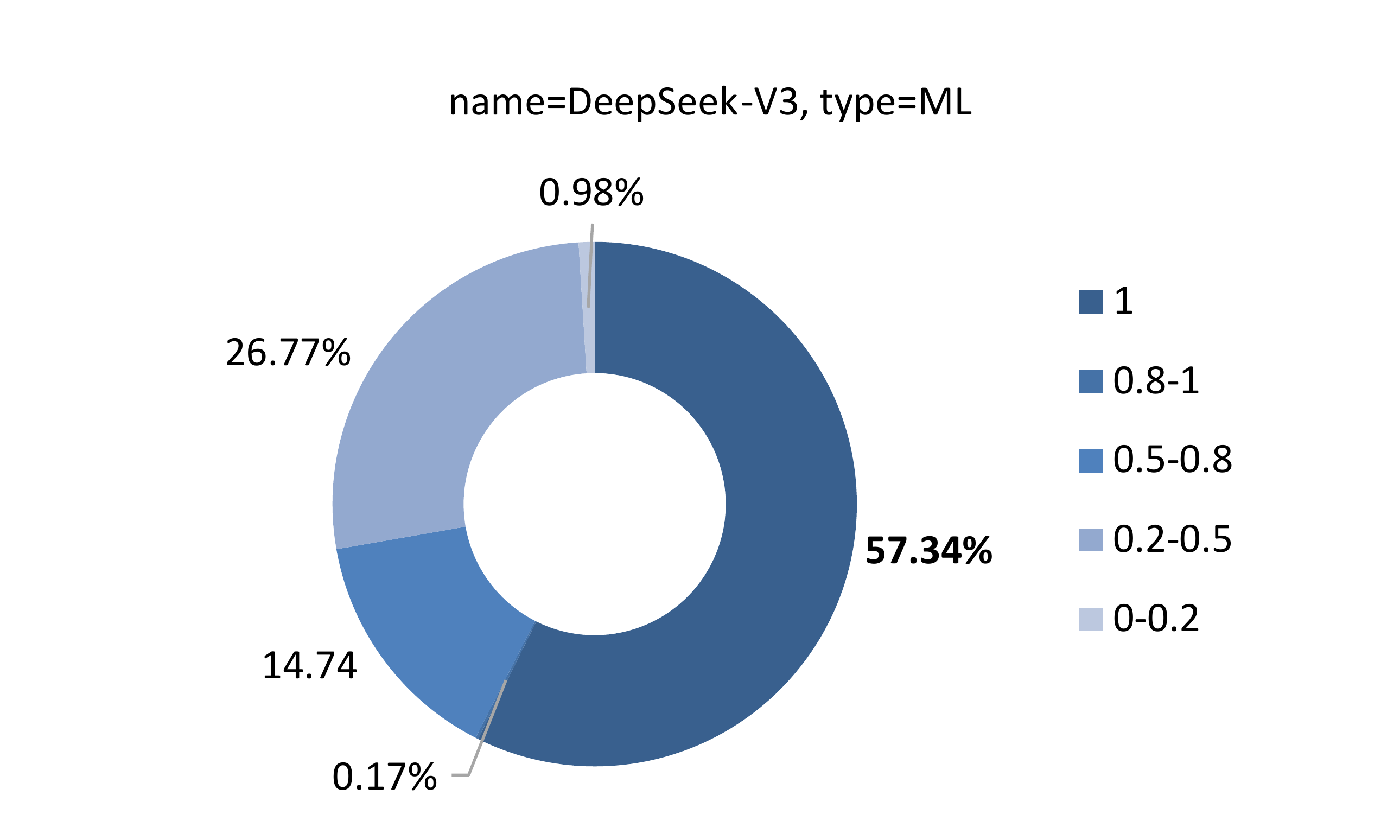}
      \includegraphics[width=0.5\columnwidth]{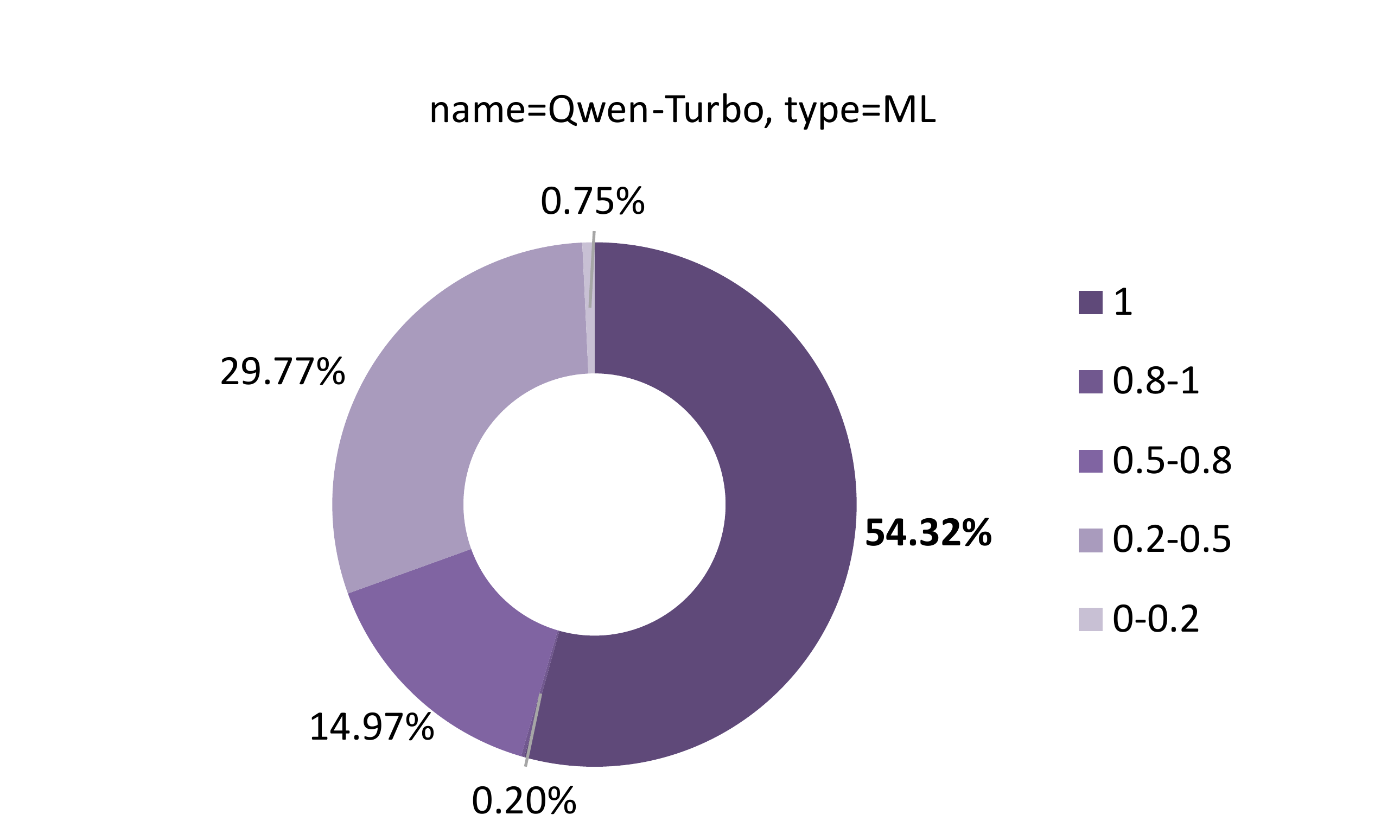}
            \includegraphics[width=0.5\columnwidth]{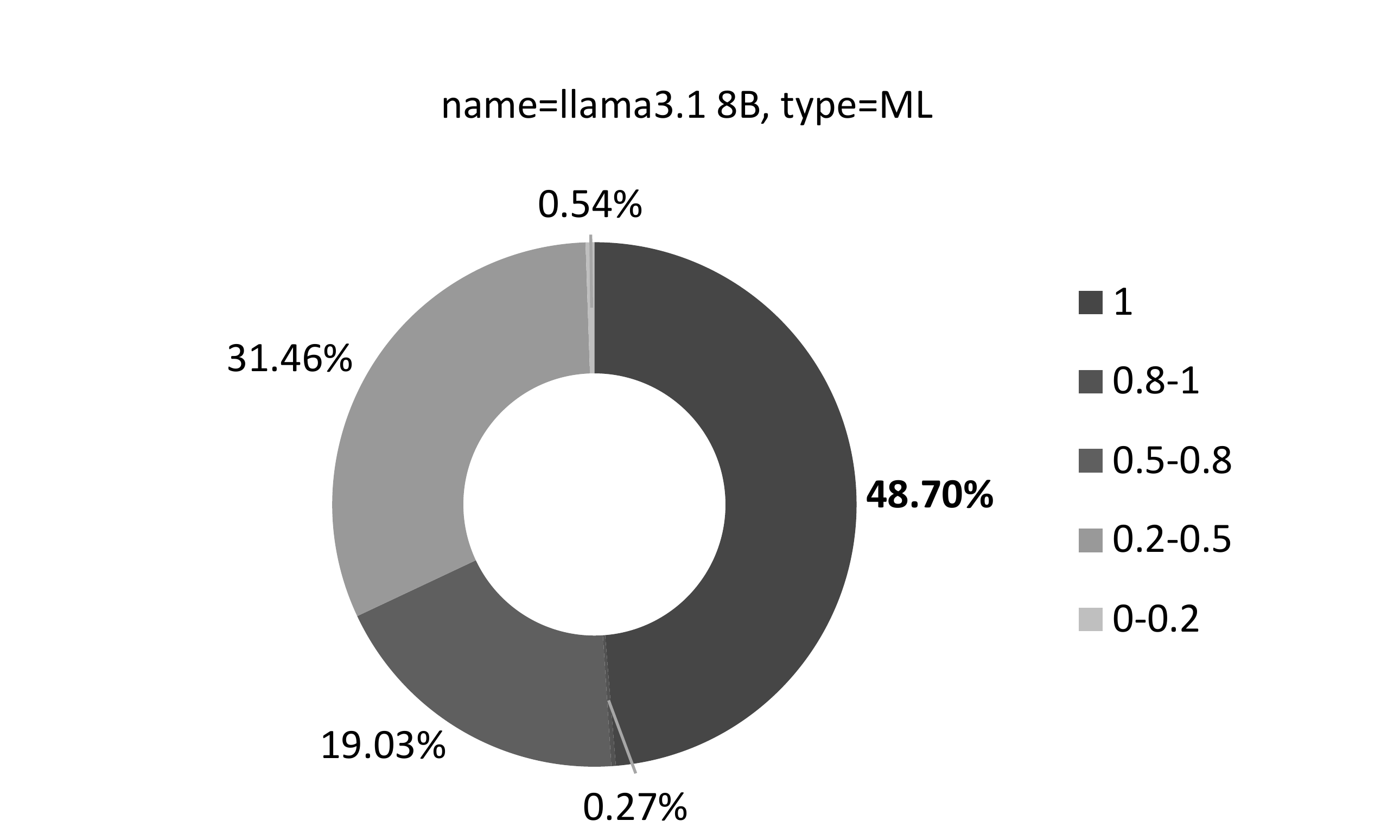}
                  \includegraphics[width=0.5\columnwidth]{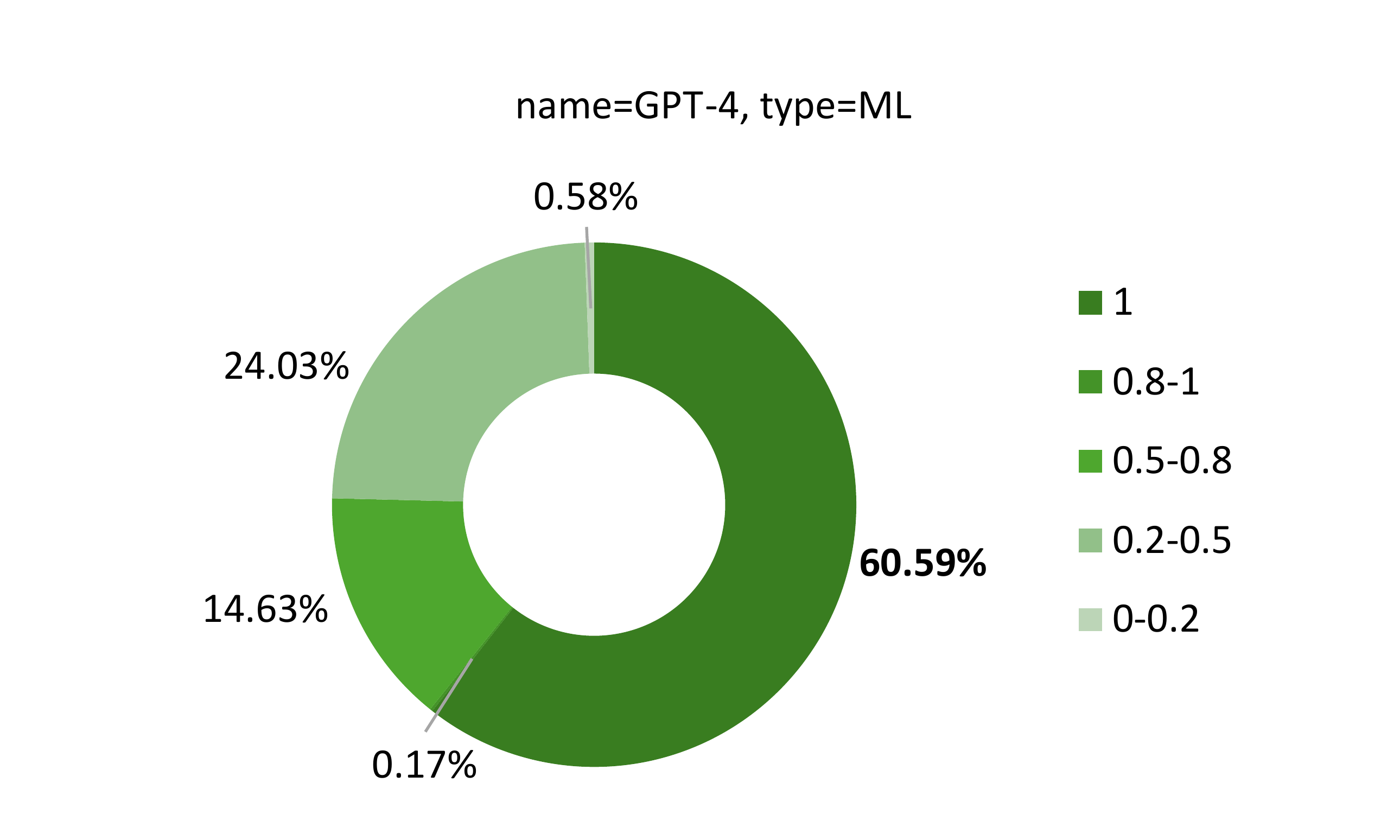}
                        \includegraphics[width=0.5\columnwidth]{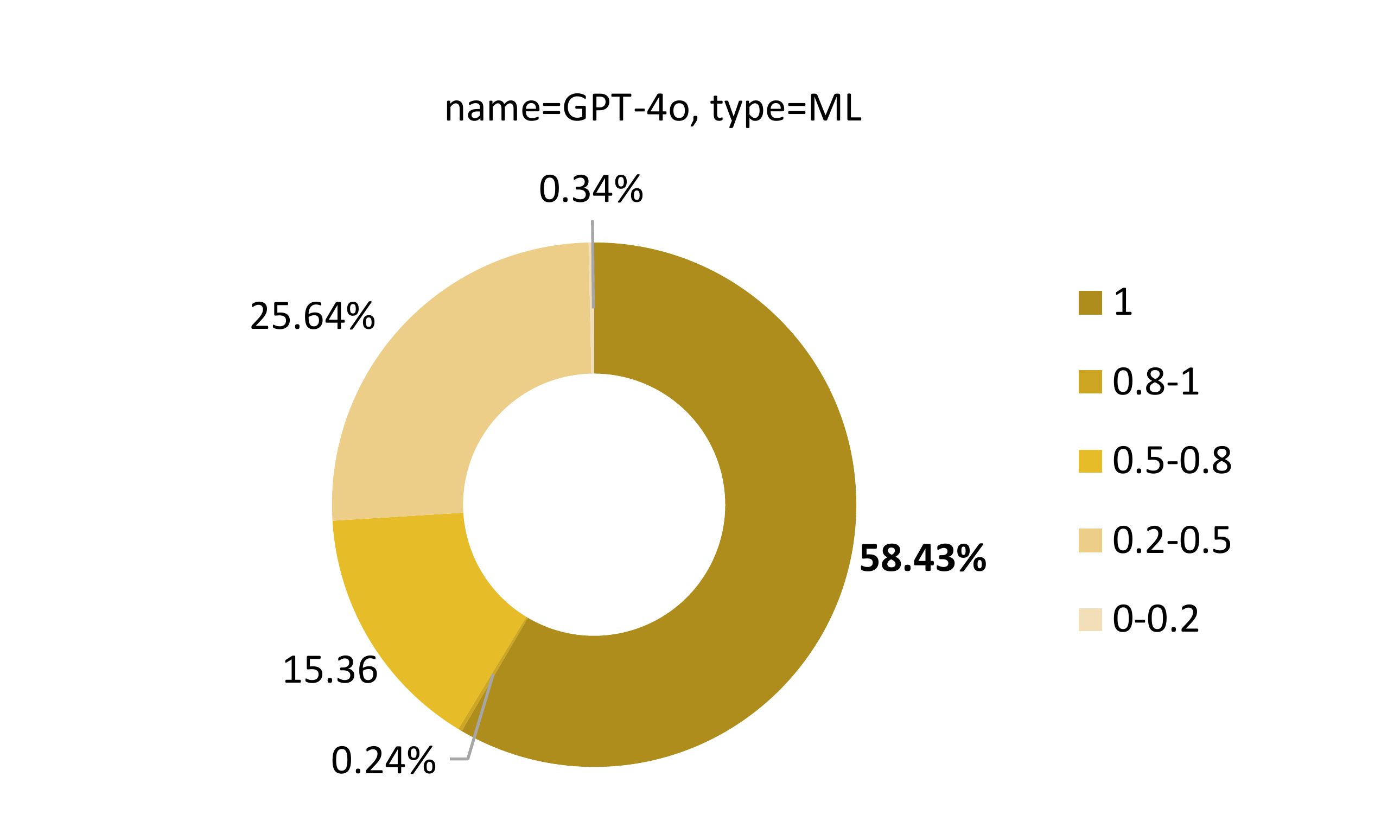}
                        \includegraphics[width=0.5\columnwidth]{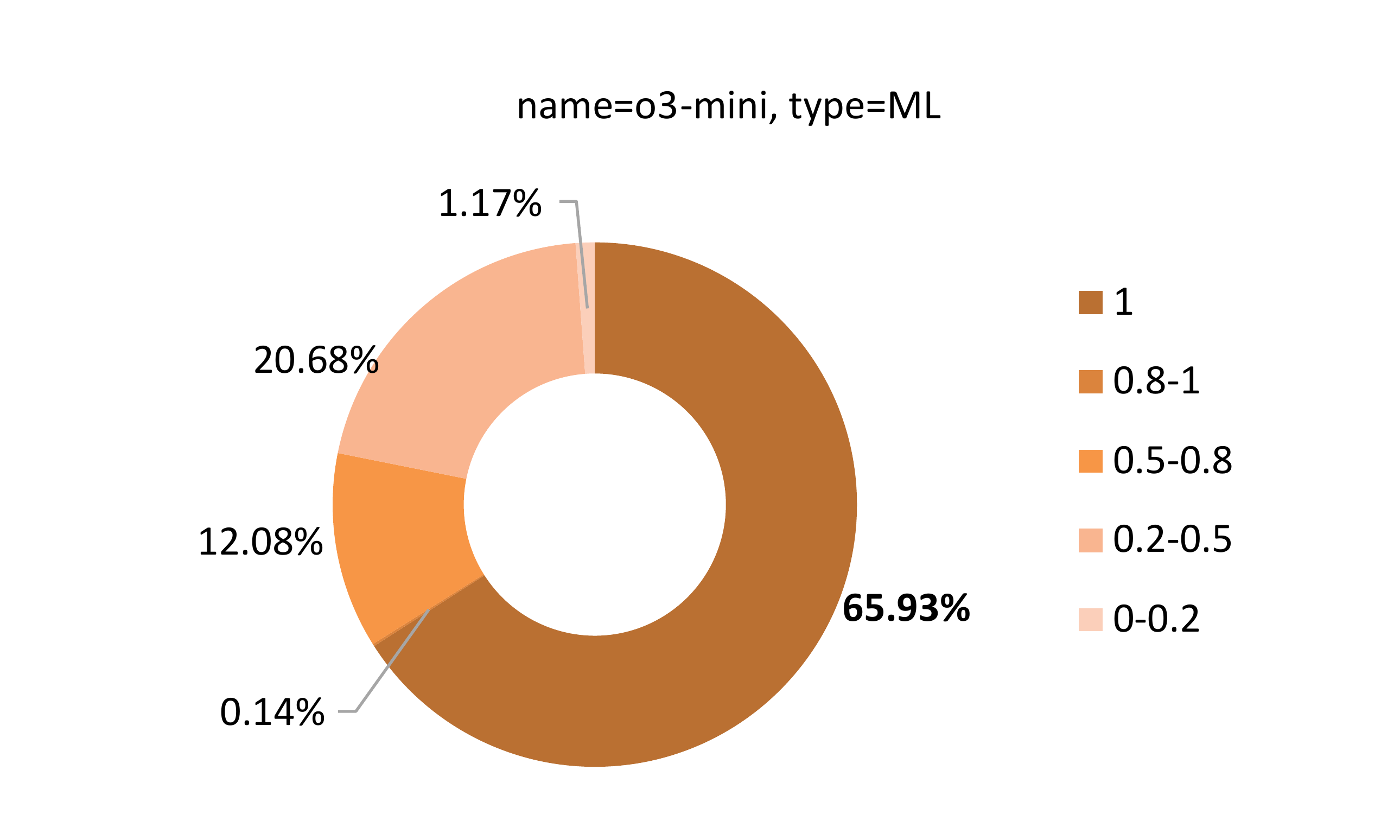}
                        \includegraphics[width=0.5\columnwidth]{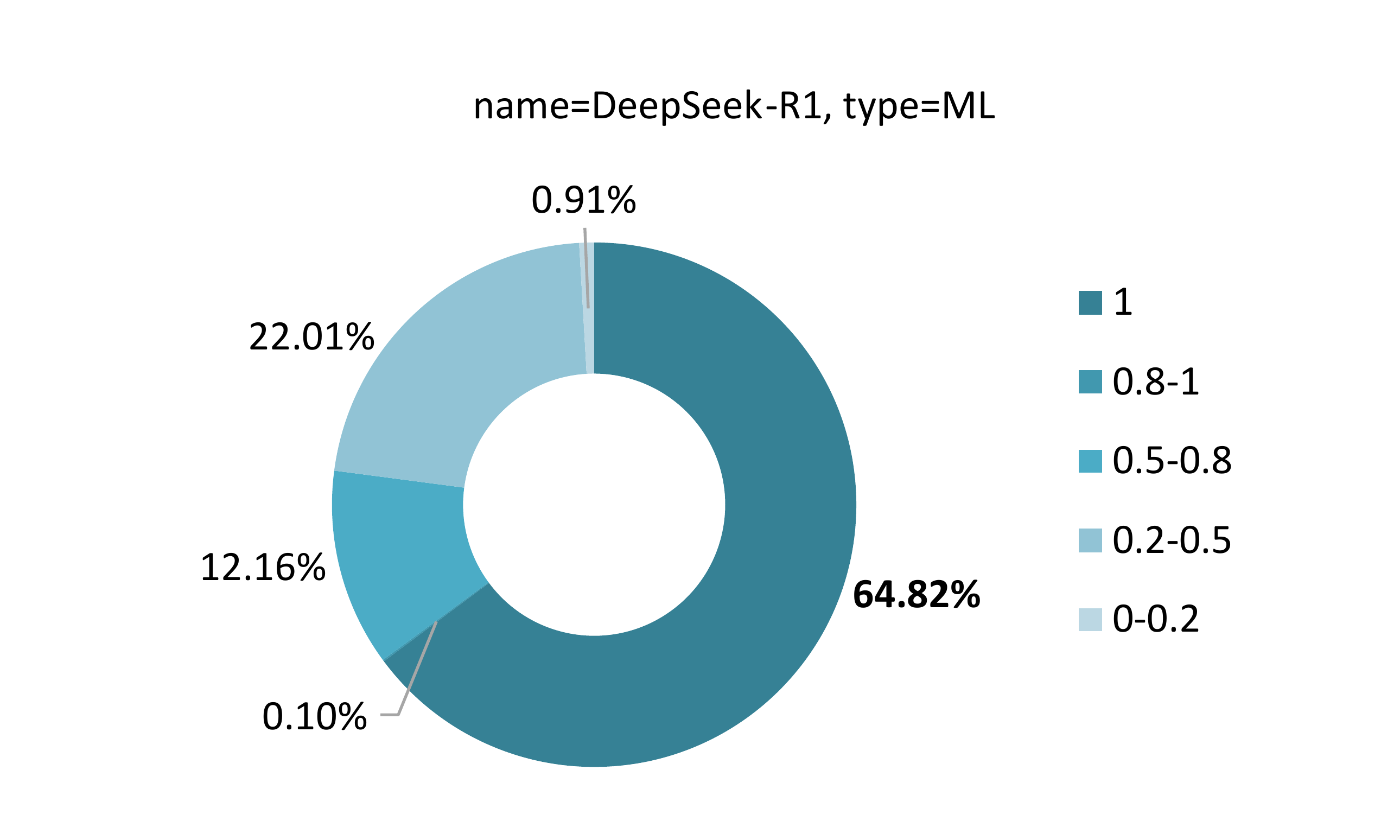}
  \caption{The Anchor Scores distributions of metaphor-literal (ML) word imagination task on every model (the largest portion is in bold and the second largest is underlined).}
  \label{fig:result_imagination3}
\end{figure*}

\begin{figure*}[h]
  \centering
  \includegraphics[width=0.5\columnwidth]{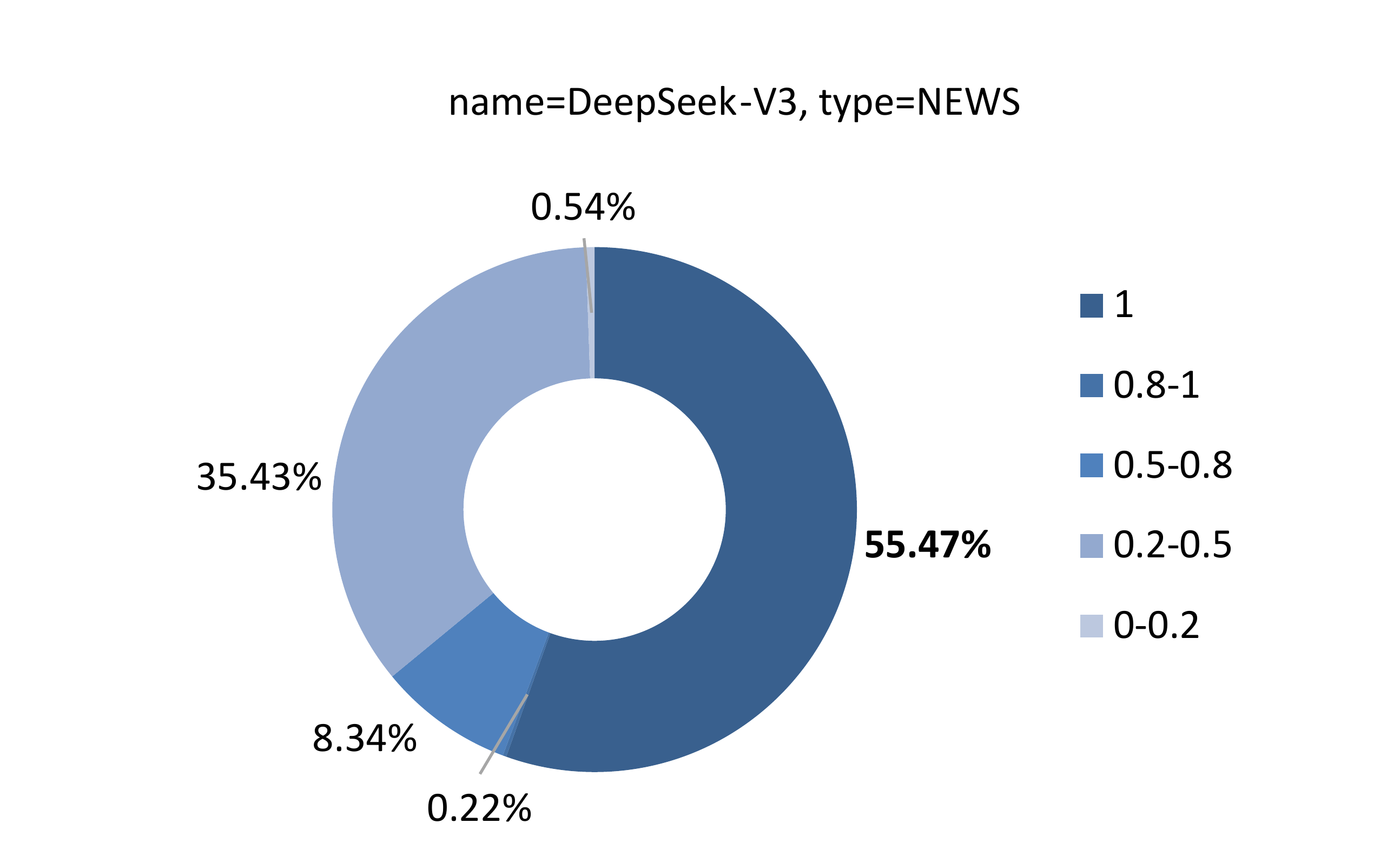}
      \includegraphics[width=0.5\columnwidth]{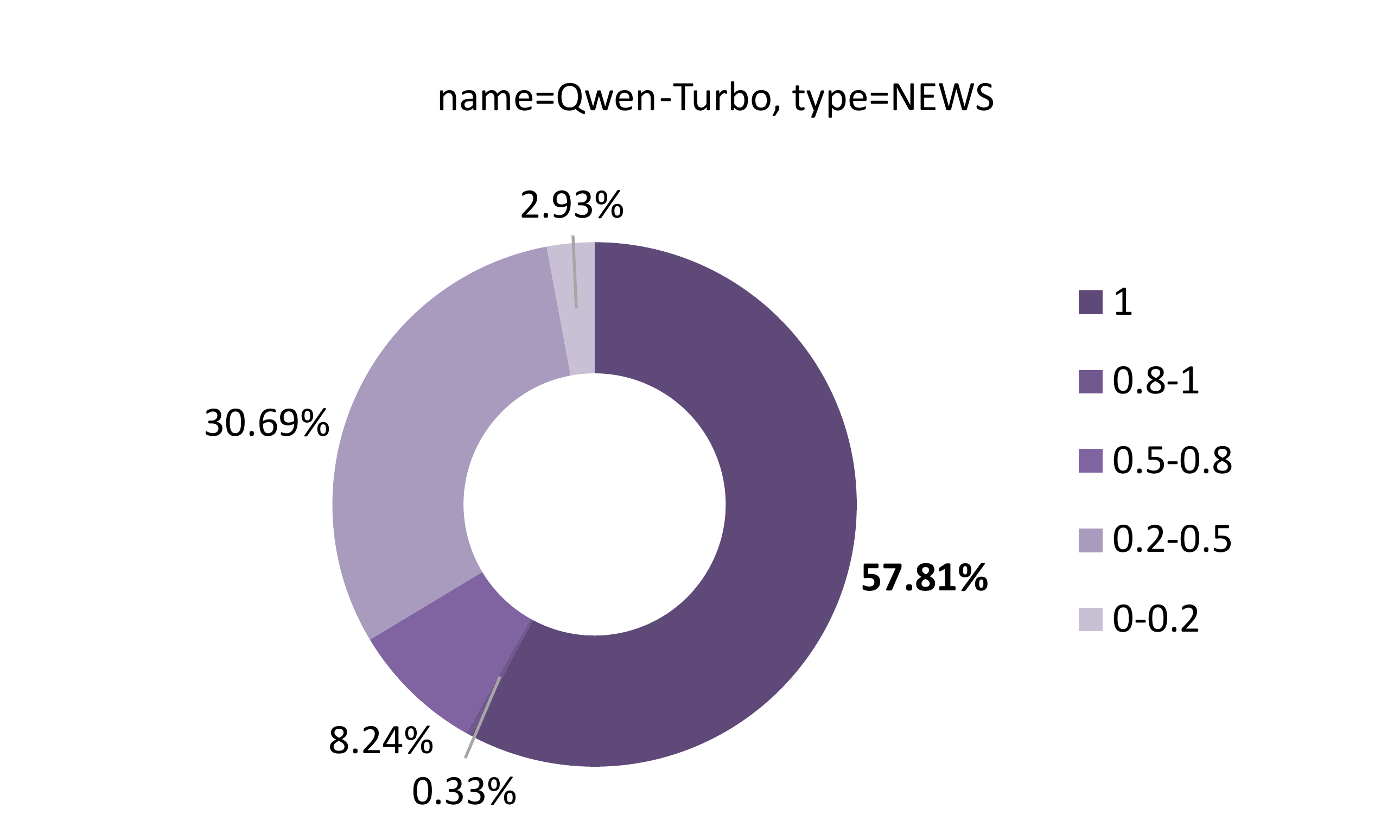}
            \includegraphics[width=0.5\columnwidth]{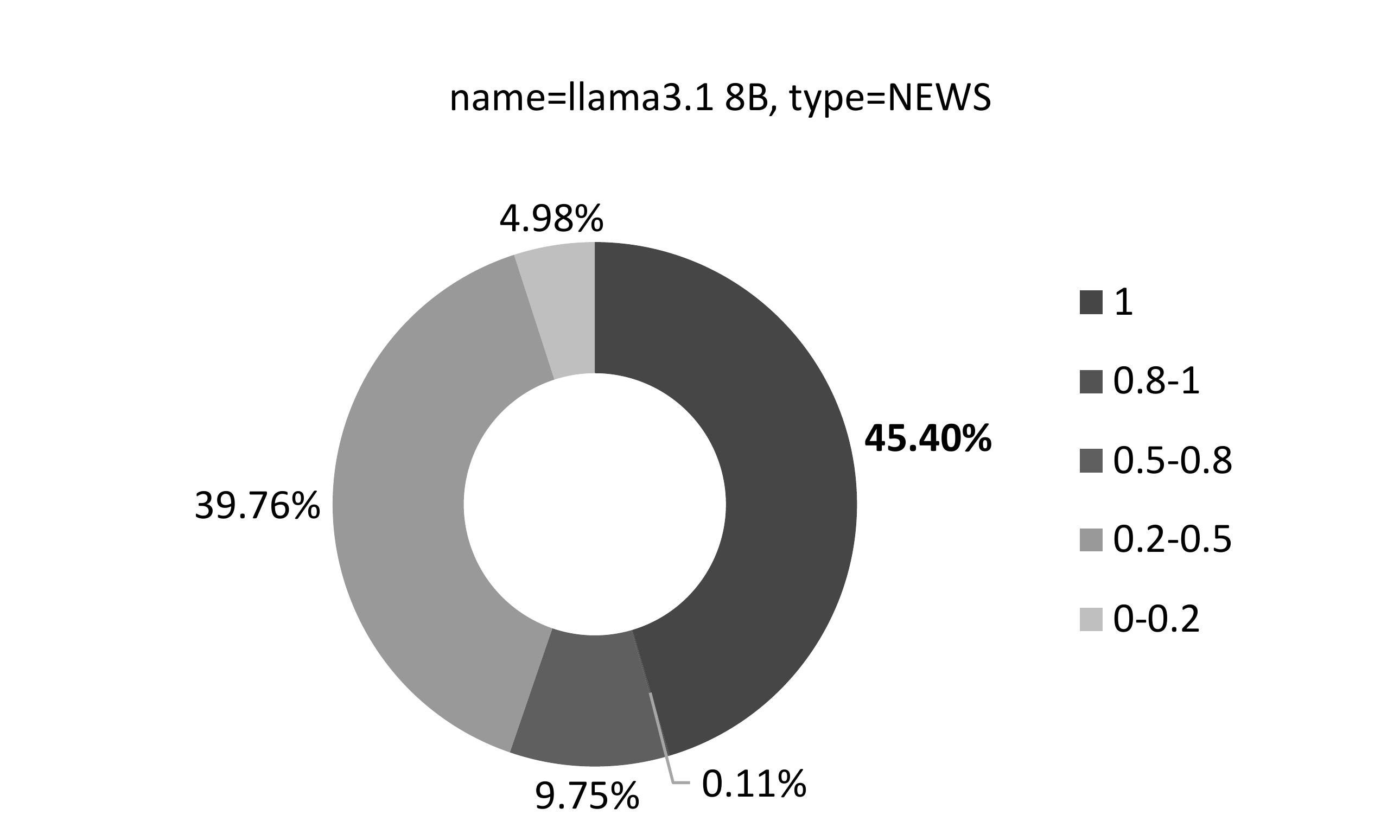}
                  \includegraphics[width=0.5\columnwidth]{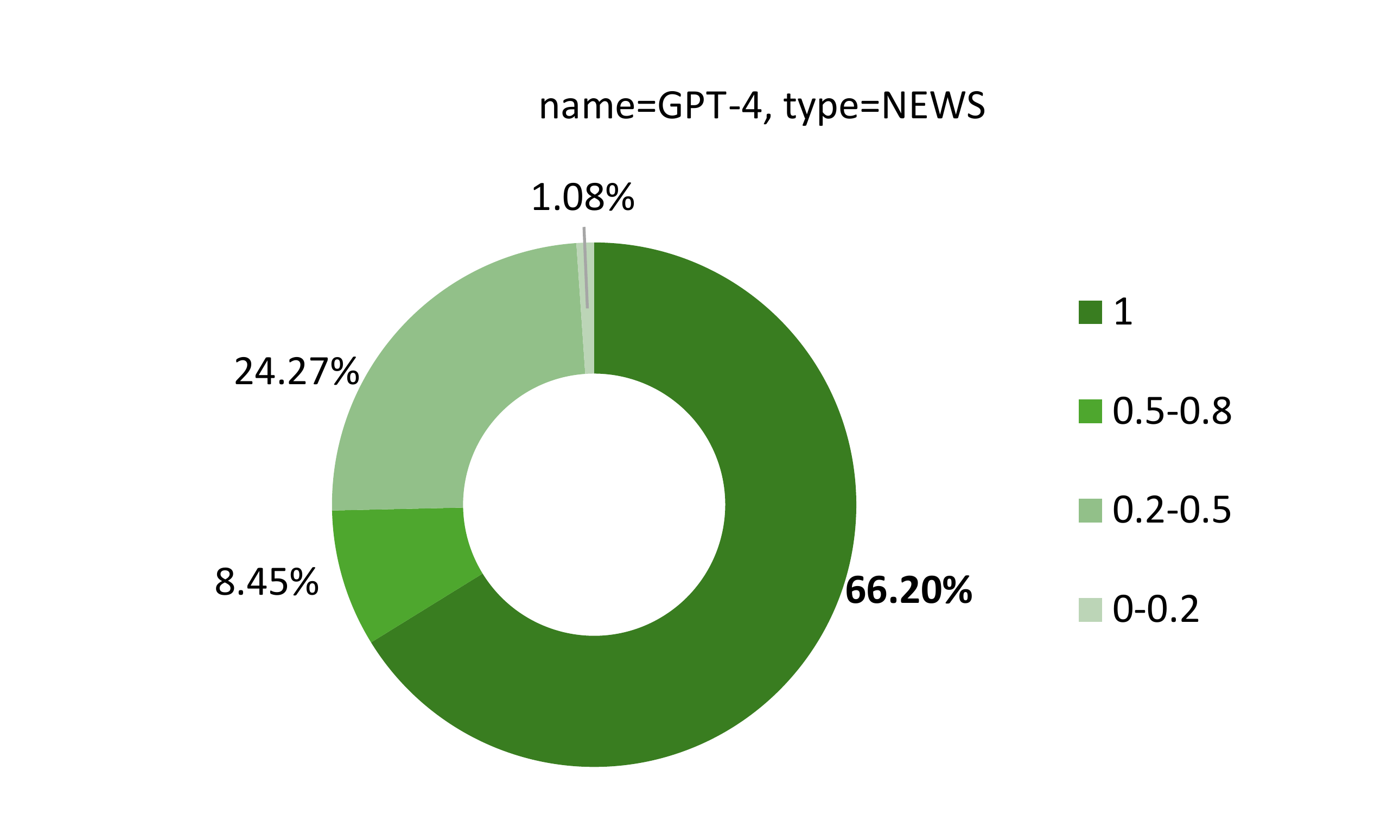}
                        \includegraphics[width=0.5\columnwidth]{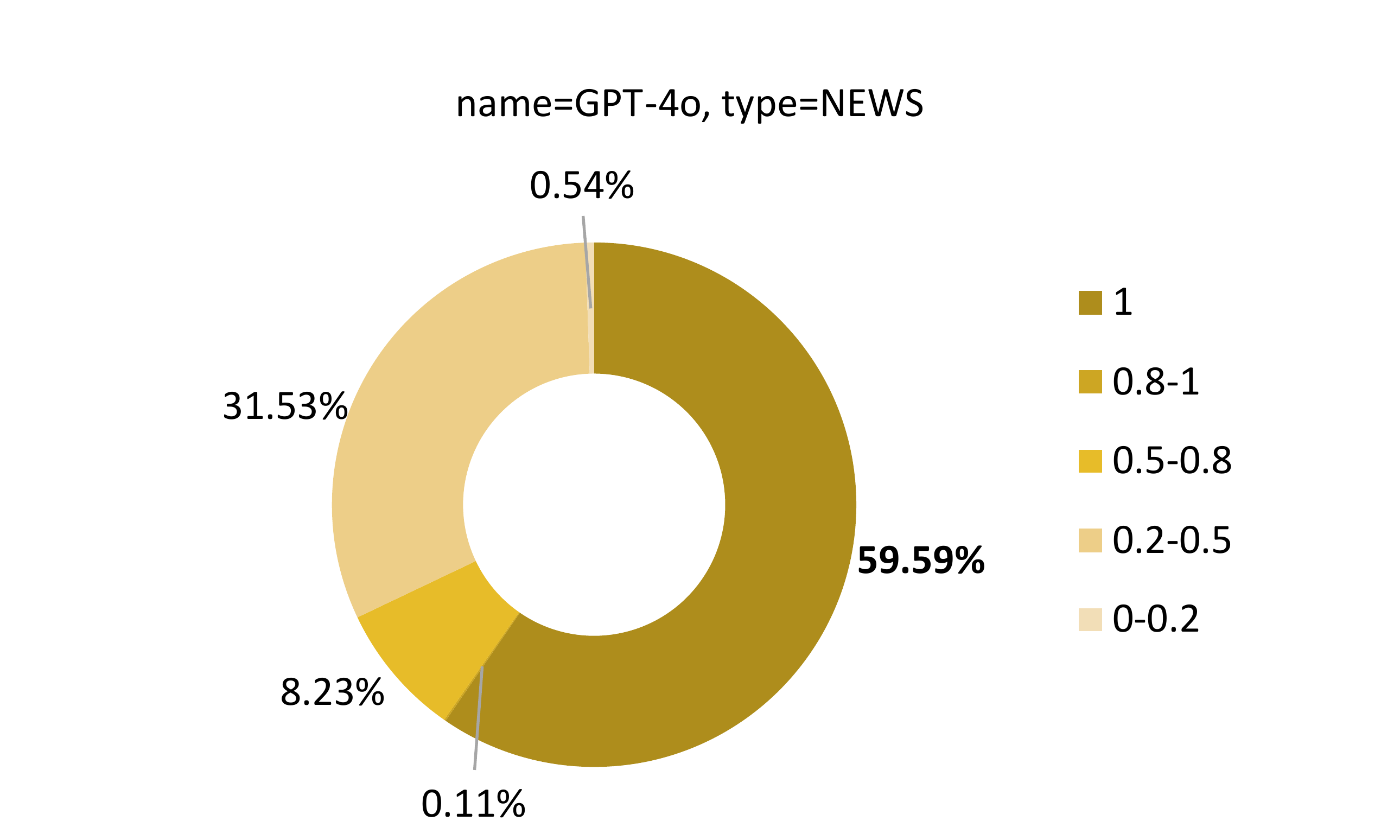}
                        \includegraphics[width=0.5\columnwidth]{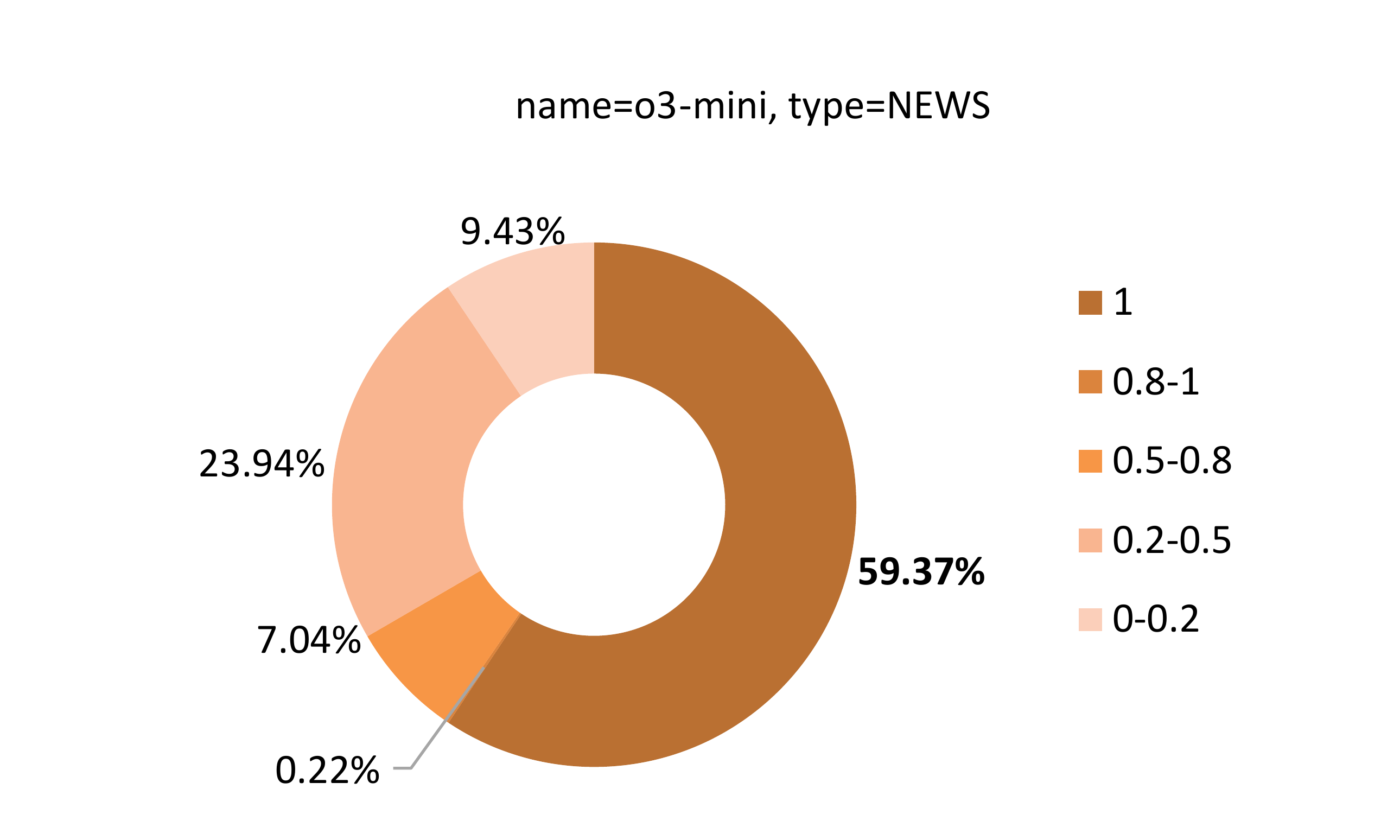}
                        \includegraphics[width=0.5\columnwidth]{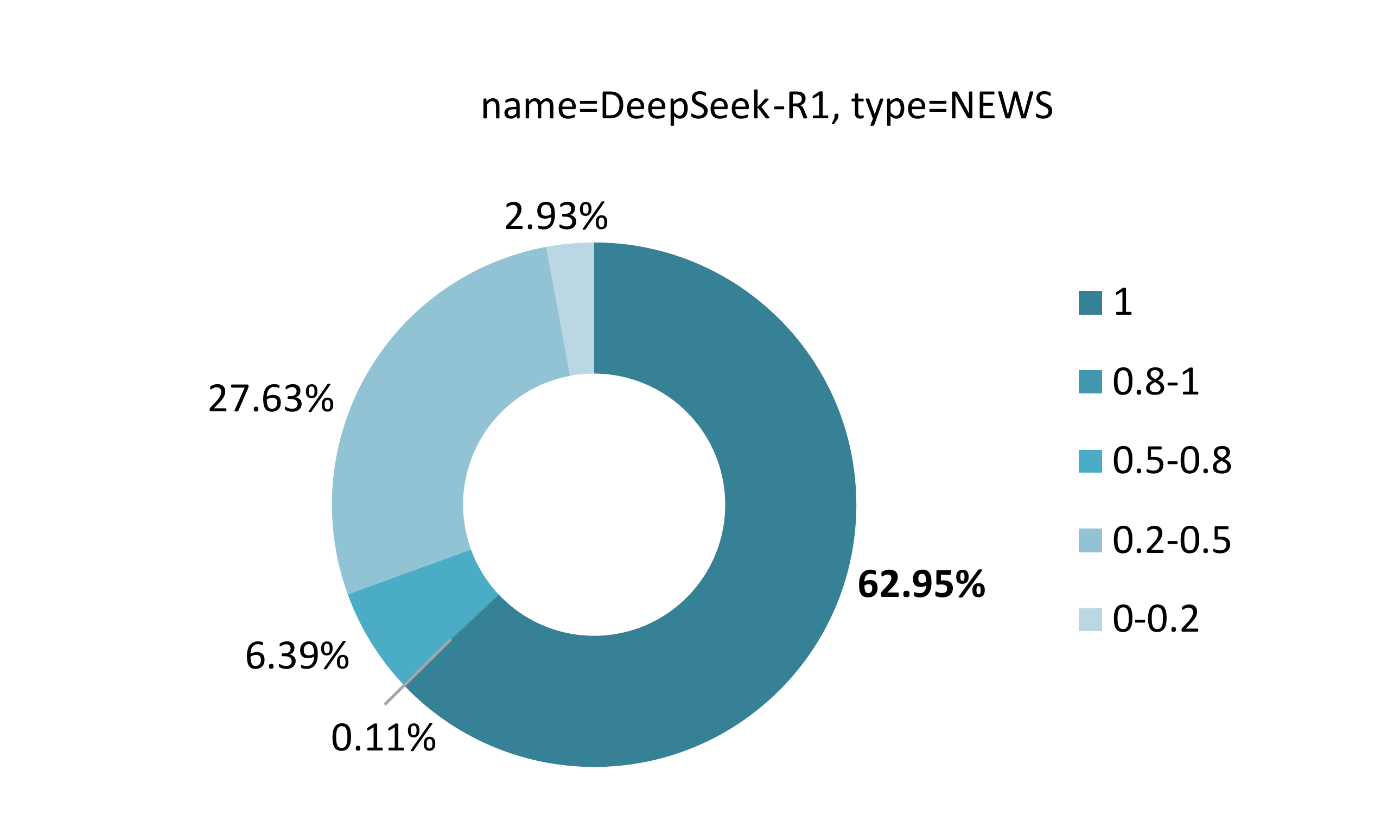}
  \caption{The Anchor Scores distributions of literal-metaphor (LM) word imagination task with sentences in NEWS on every model (the largest portion is in bold and the second largest is underlined).}
  \label{fig:result_imagination4}
\end{figure*}

\begin{figure*}[h]
  \centering
  \includegraphics[width=0.5\columnwidth]{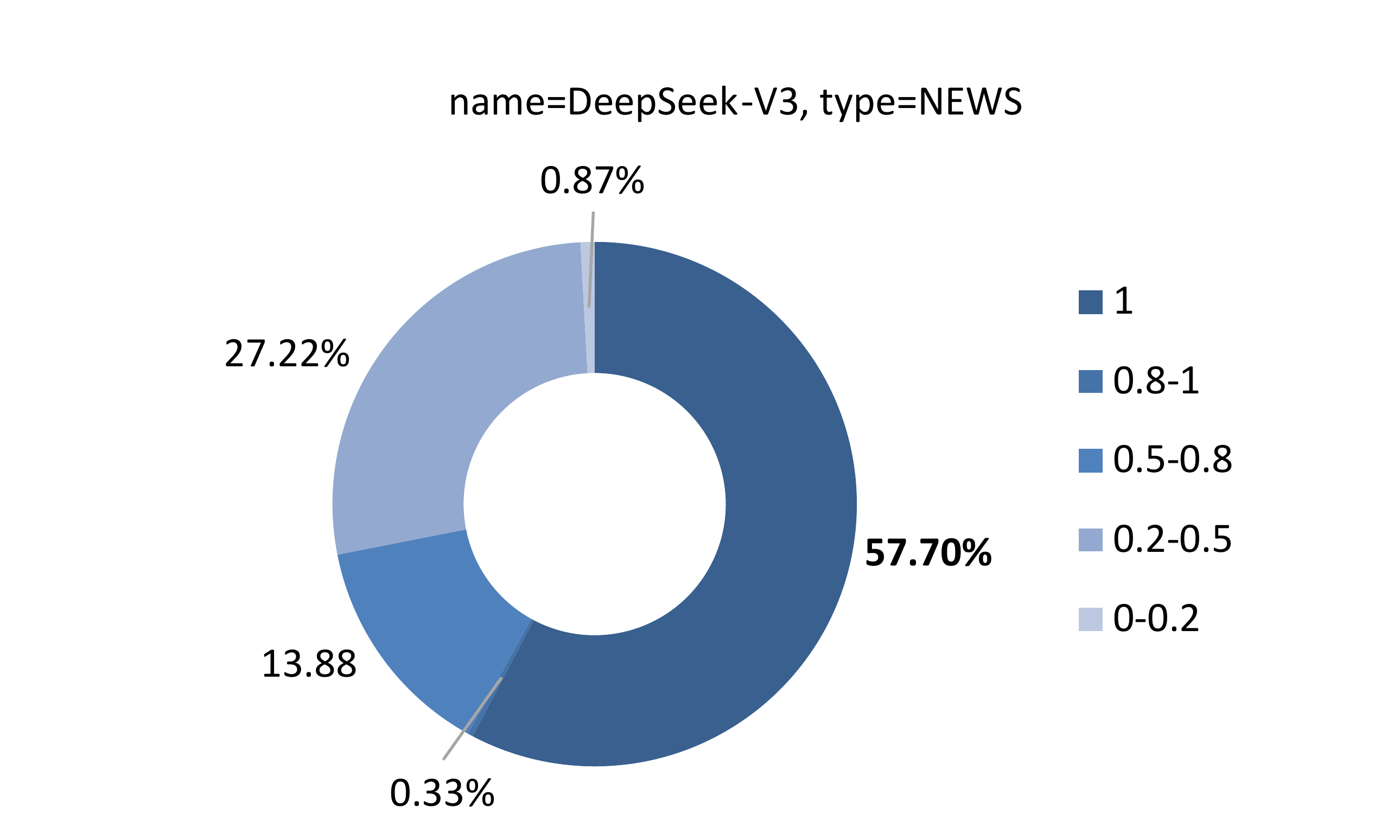}
      \includegraphics[width=0.5\columnwidth]{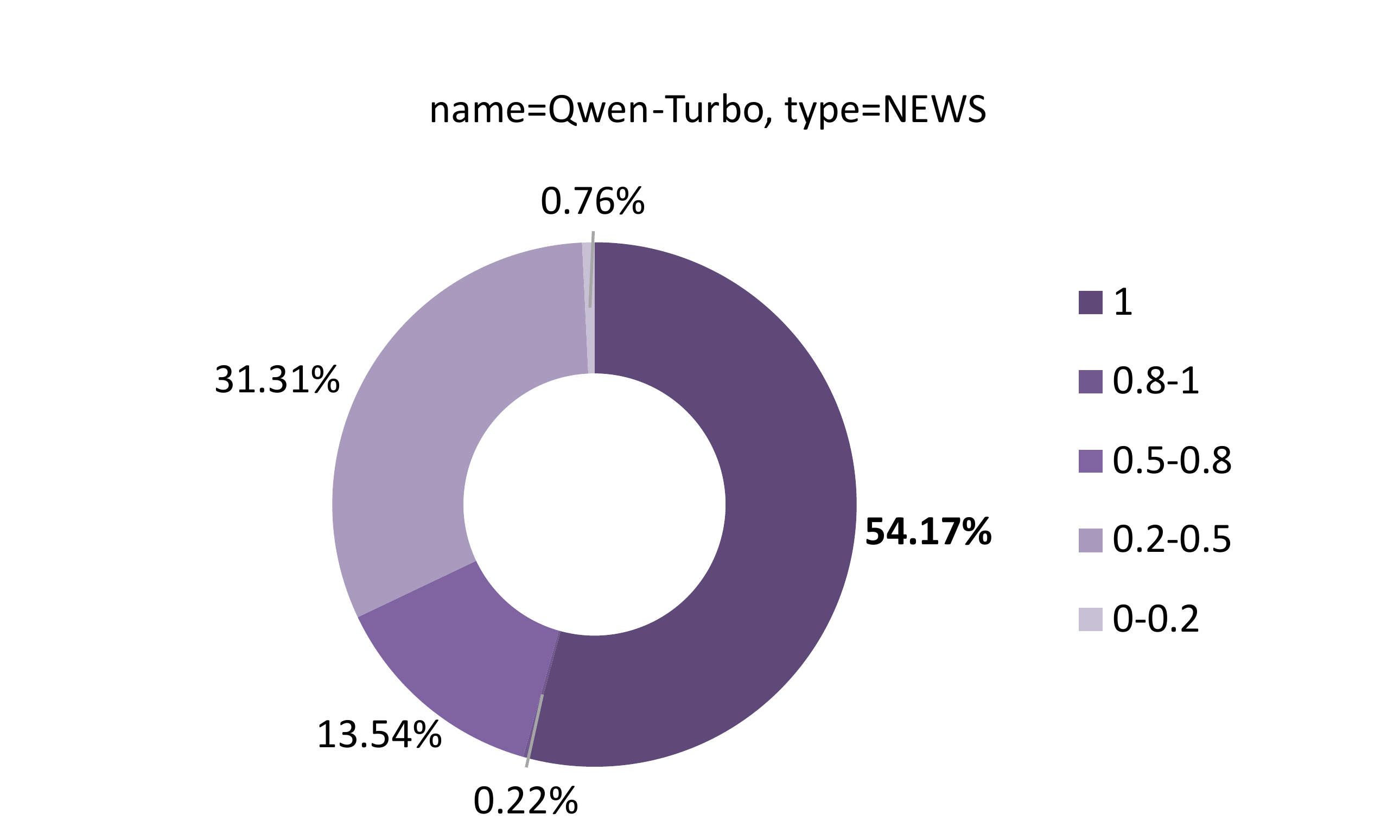}
            \includegraphics[width=0.5\columnwidth]{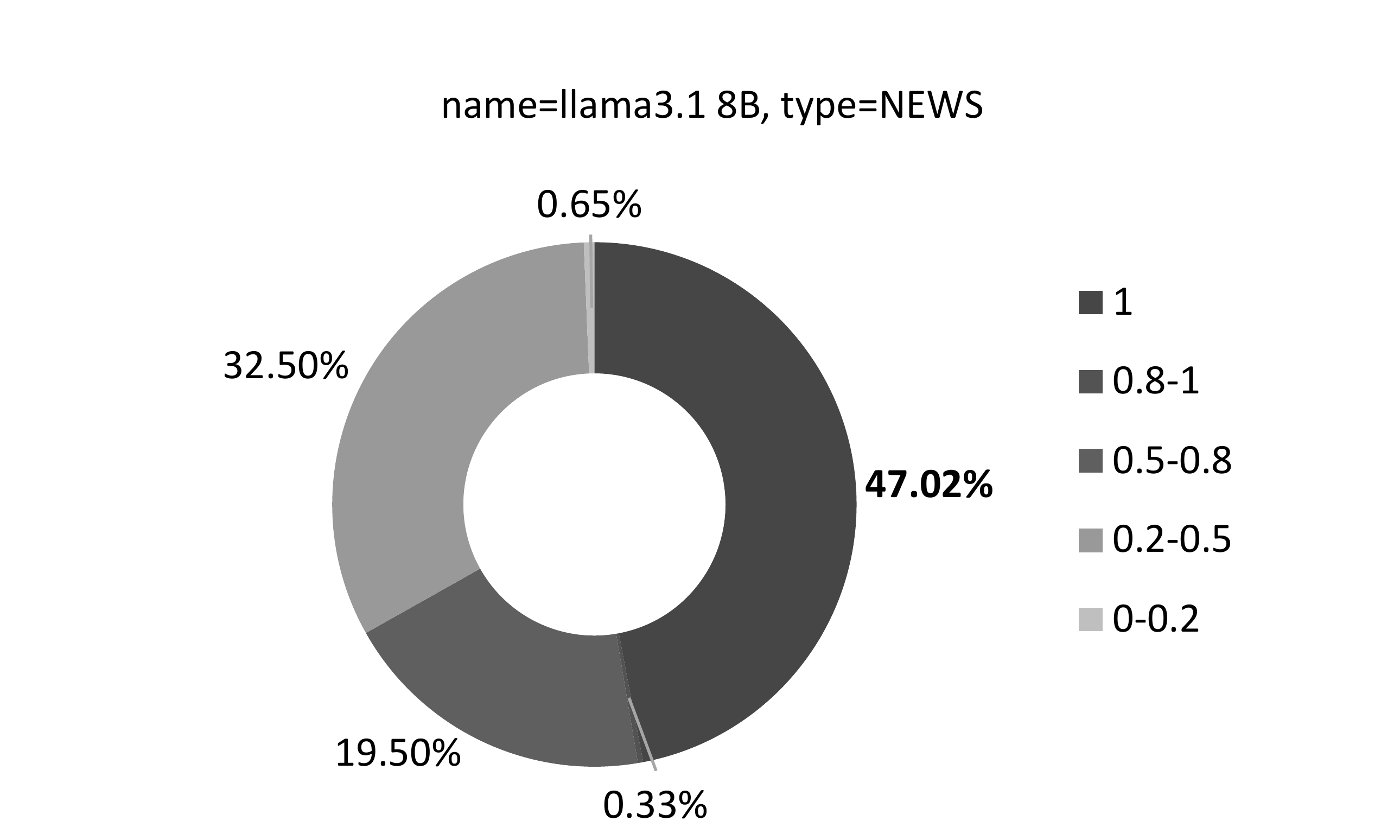}
                  \includegraphics[width=0.5\columnwidth]{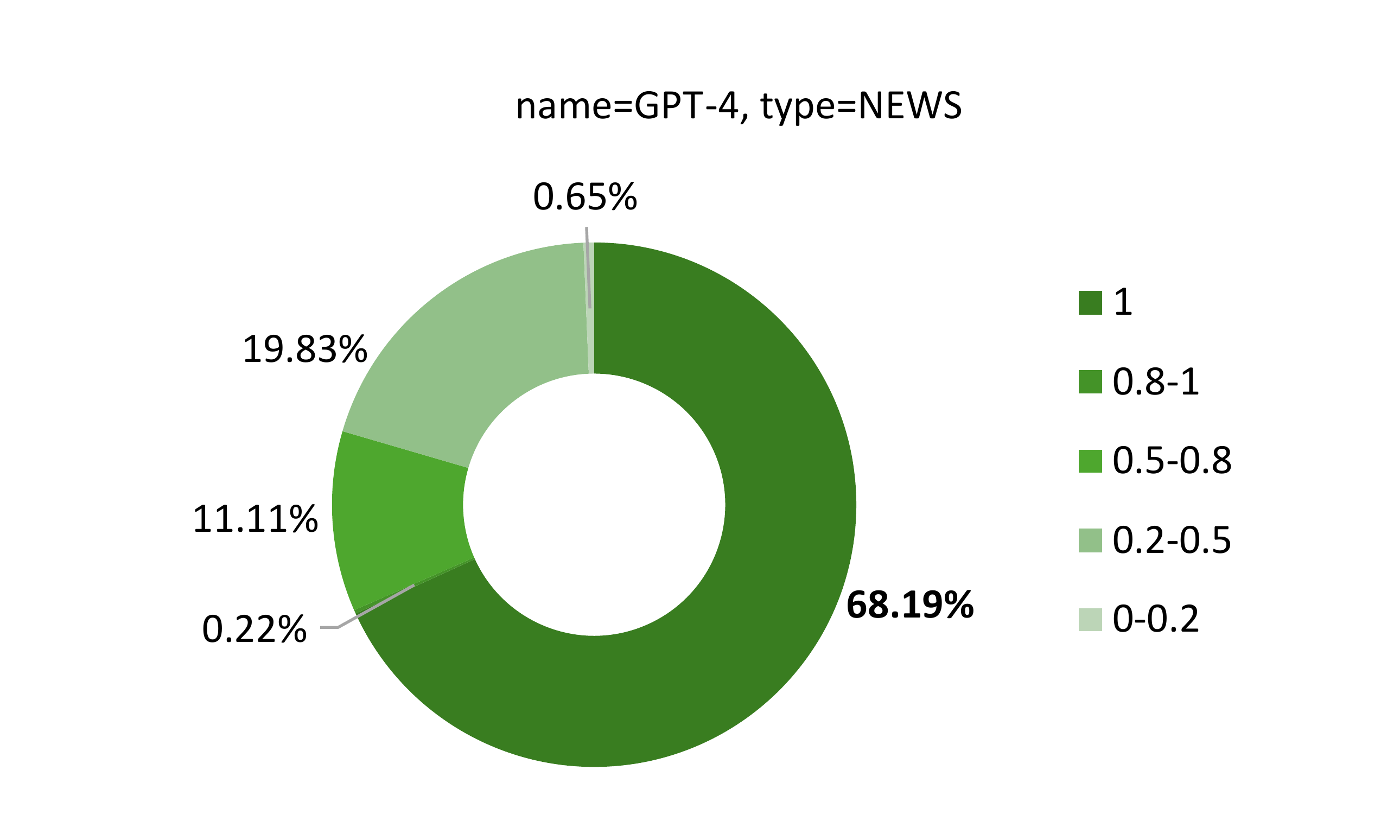}
                        \includegraphics[width=0.5\columnwidth]{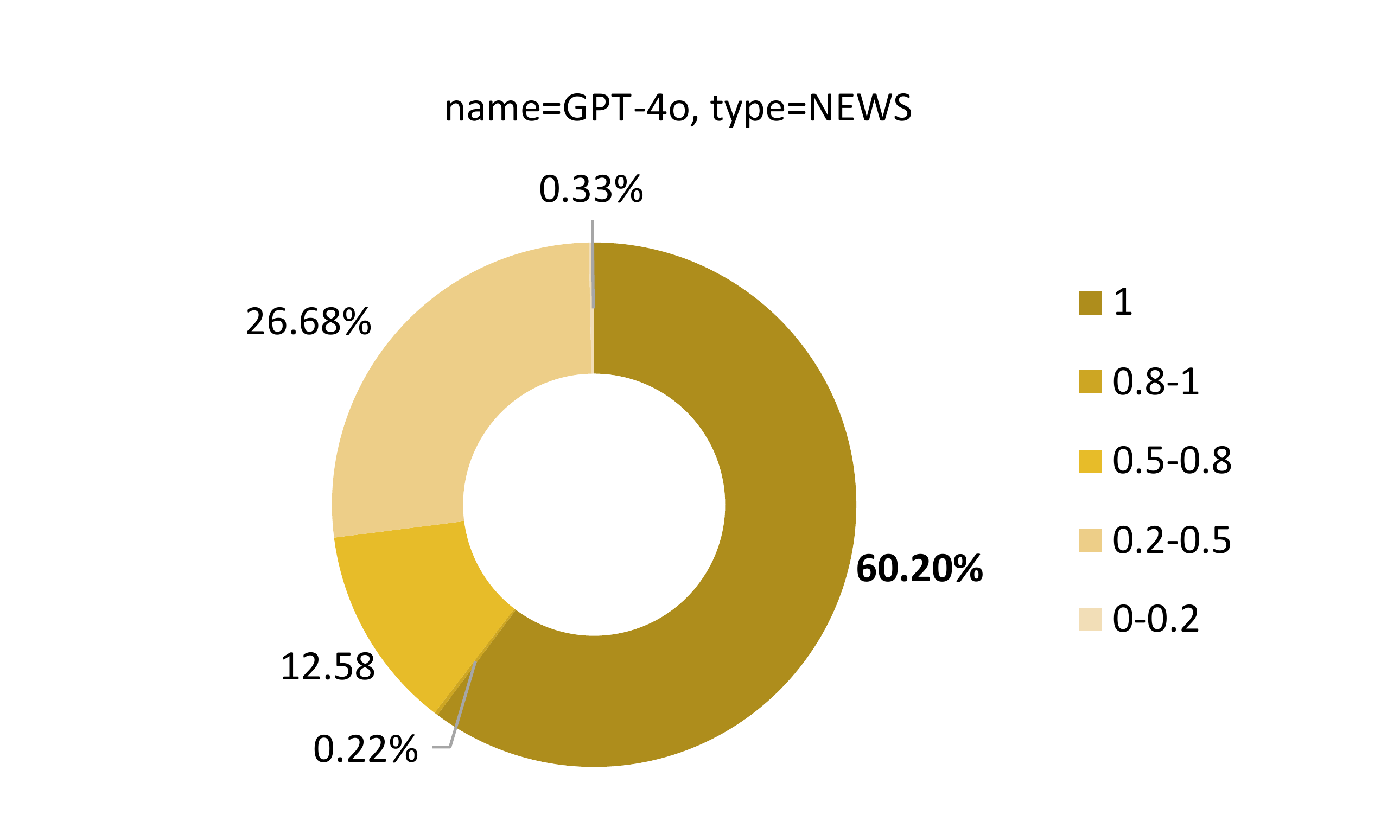}
                        \includegraphics[width=0.5\columnwidth]{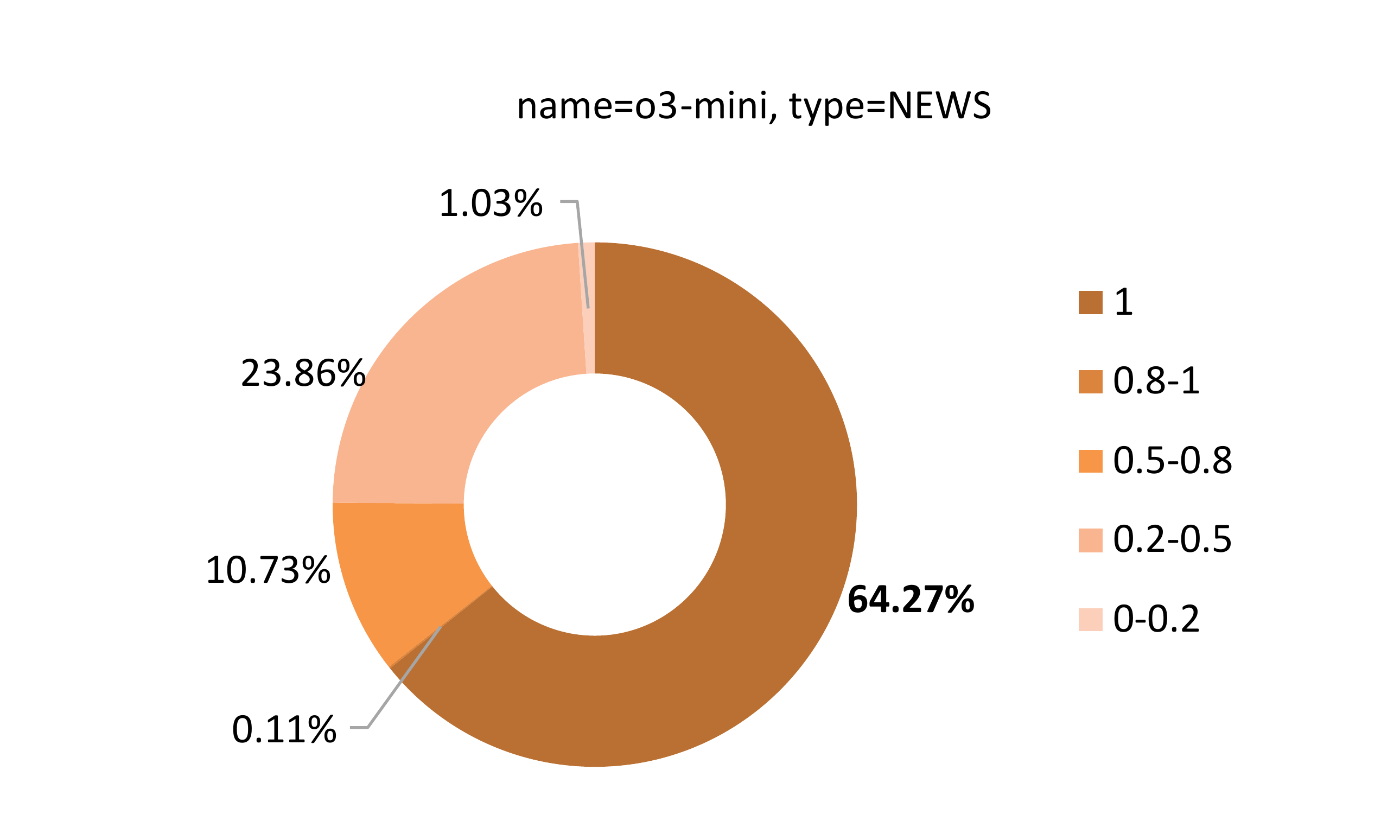}
                        \includegraphics[width=0.5\columnwidth]{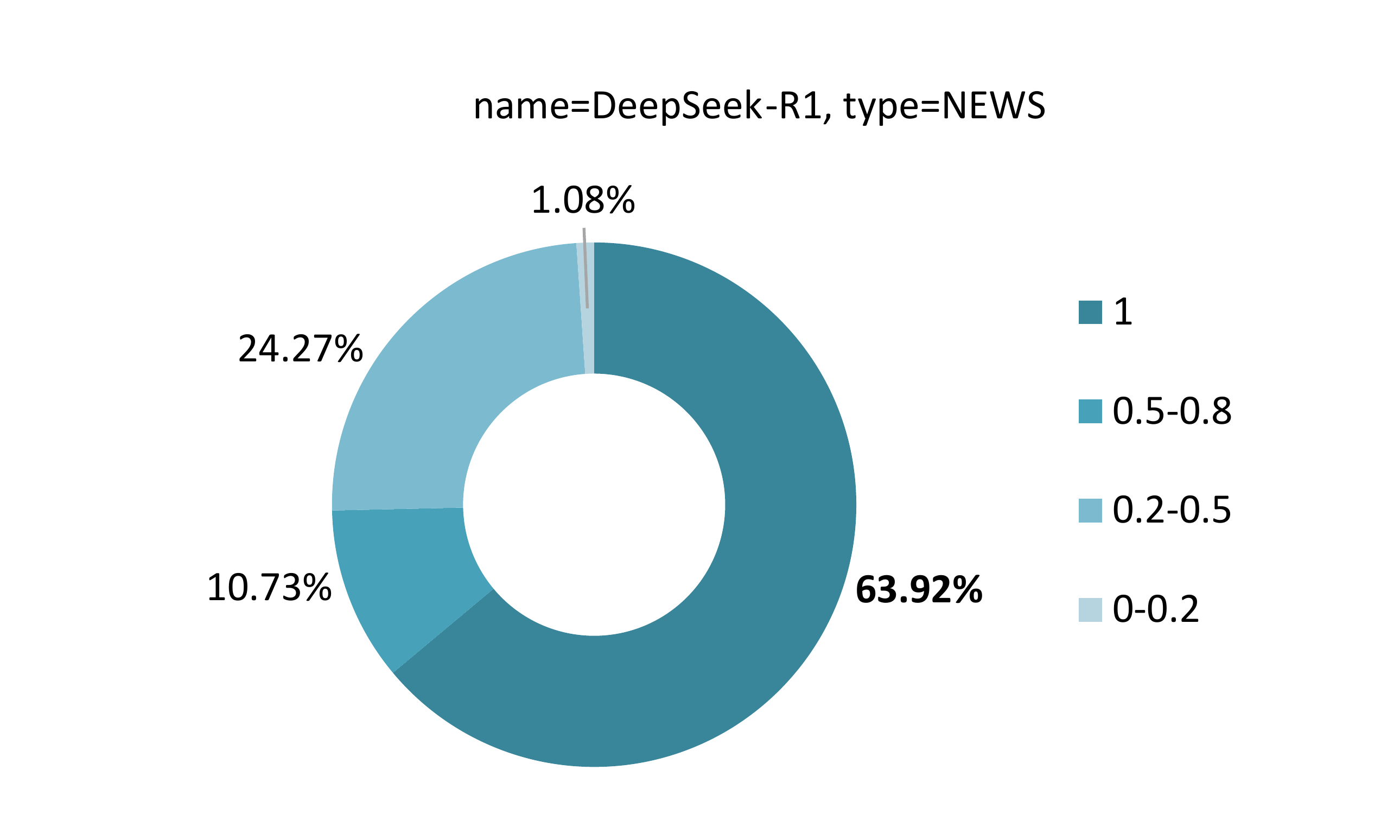}
  \caption{The Anchor Scores distributions of metaphor-literal (ML) word imagination task with sentences in NEWS on every model (the largest portion is in bold and the second largest is underlined).}
  \label{fig:result_imagination5}
\end{figure*}

\begin{figure*}[h]
  \centering
  \includegraphics[width=0.5\columnwidth]{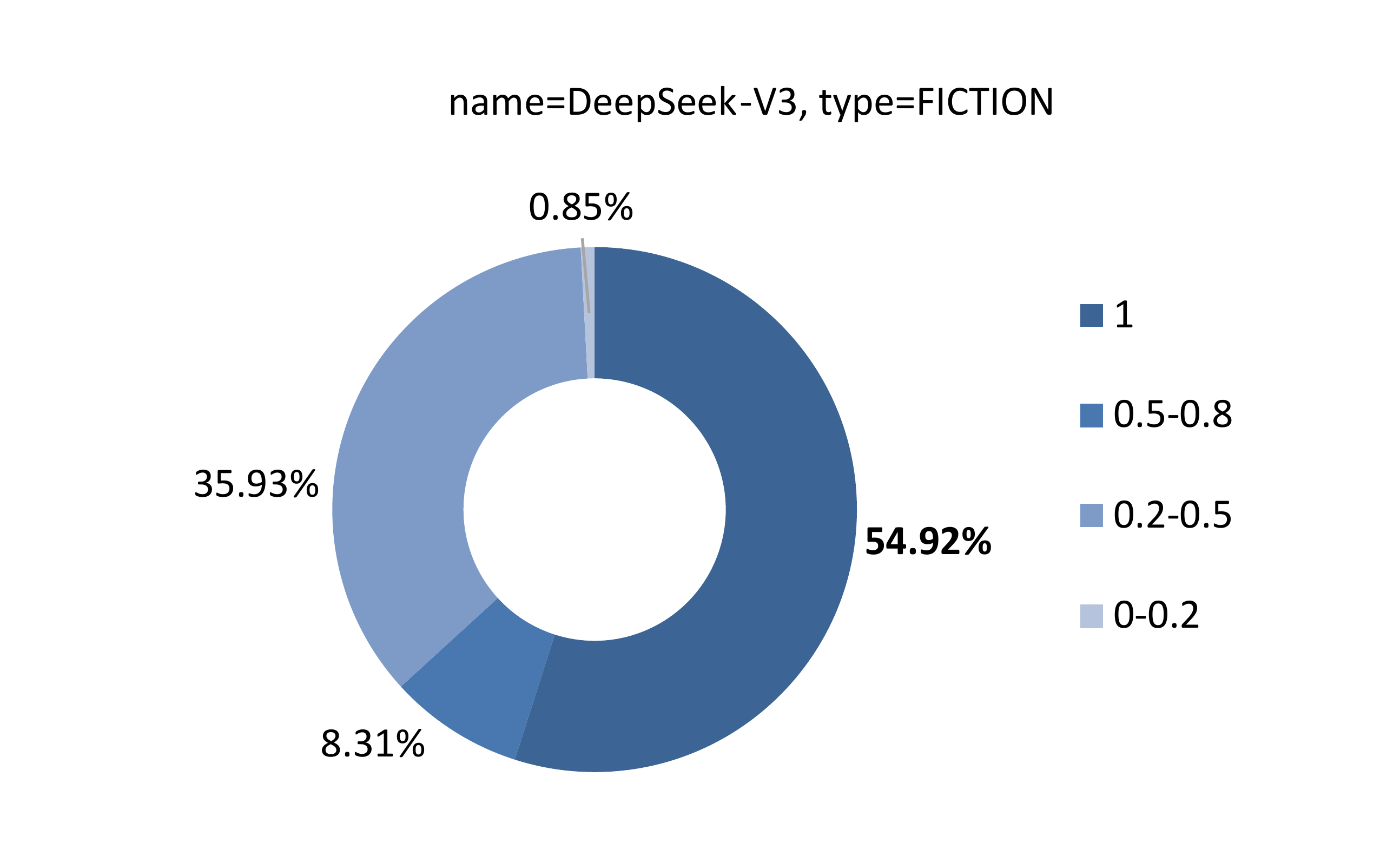}
      \includegraphics[width=0.5\columnwidth]{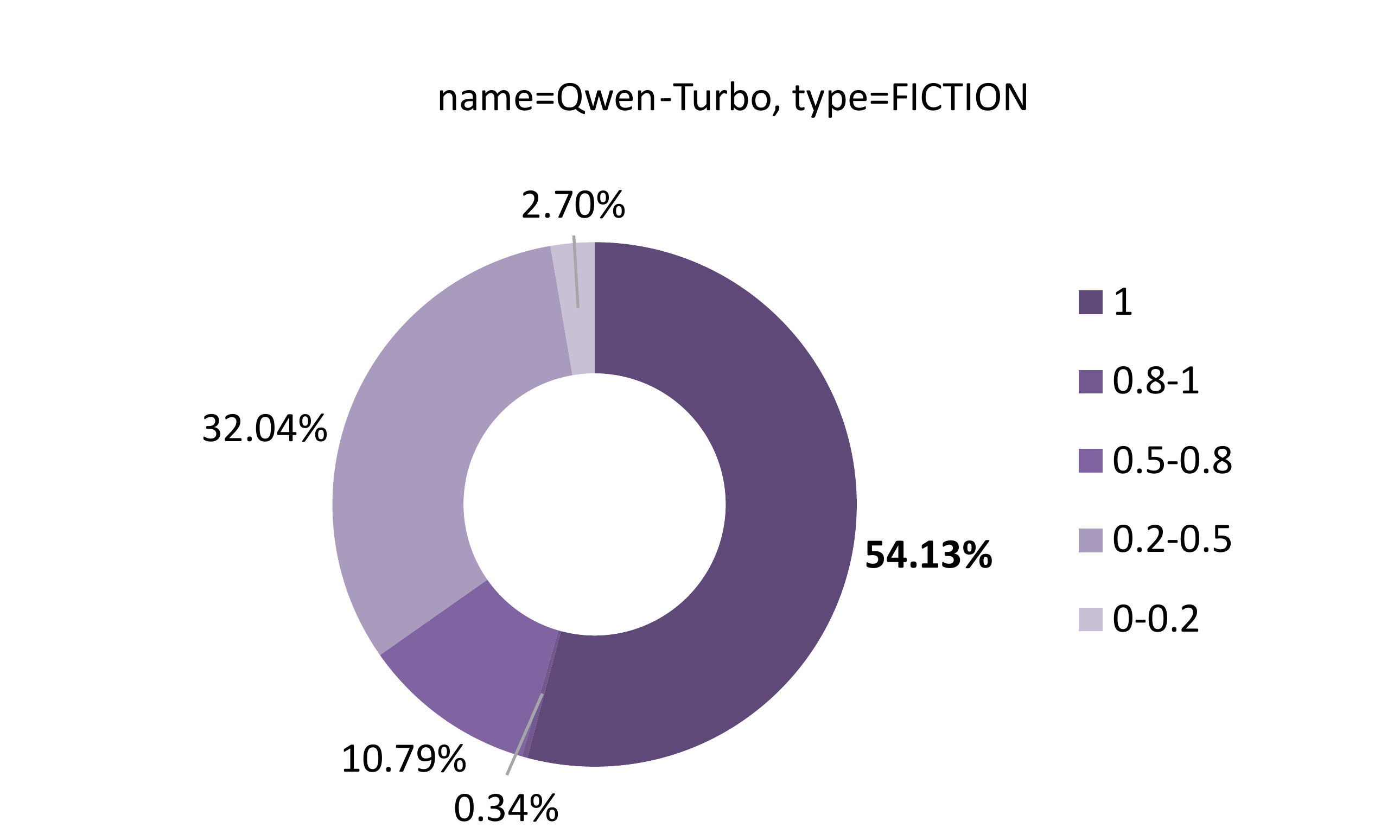}
            \includegraphics[width=0.5\columnwidth]{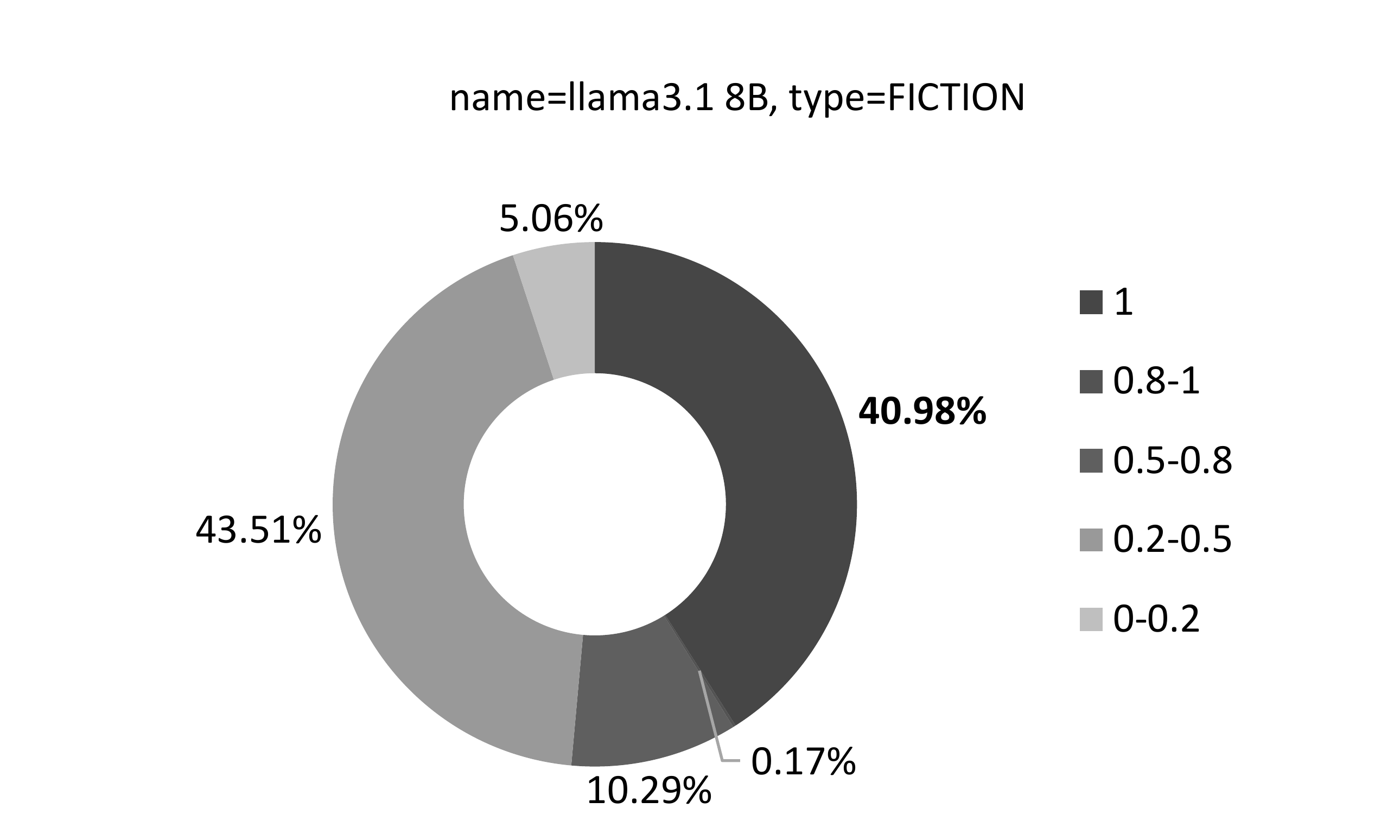}
                  \includegraphics[width=0.5\columnwidth]{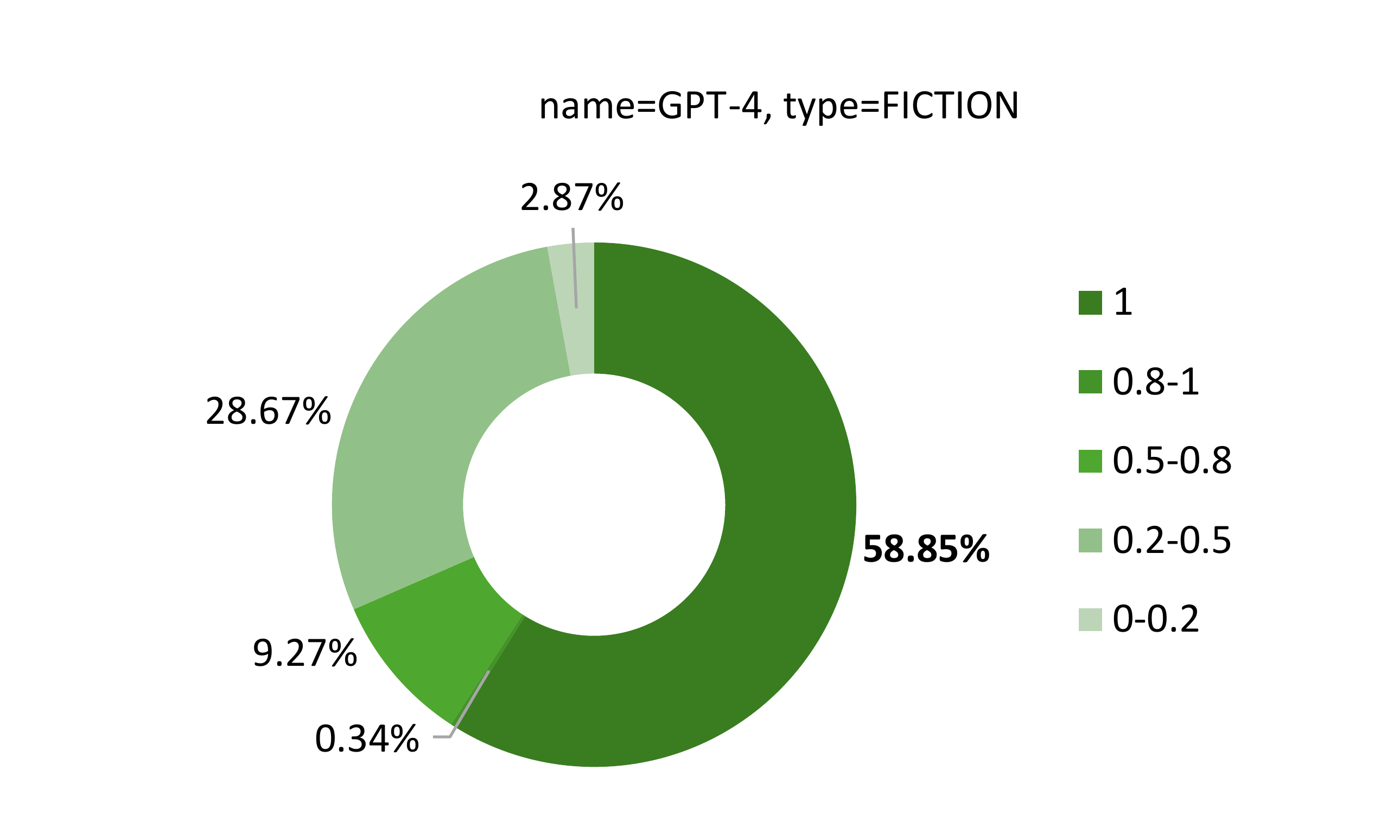}
                        \includegraphics[width=0.5\columnwidth]{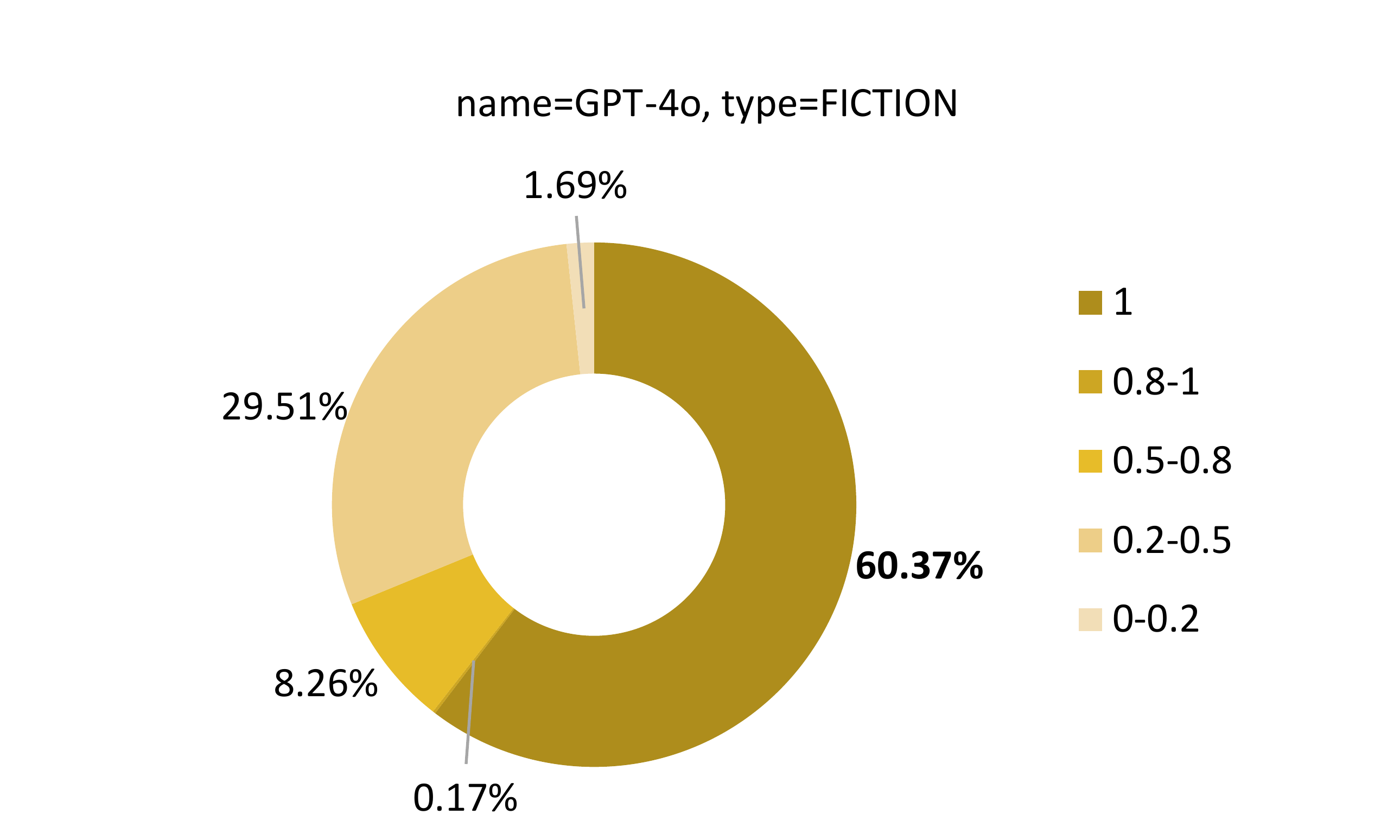}
                        \includegraphics[width=0.5\columnwidth]{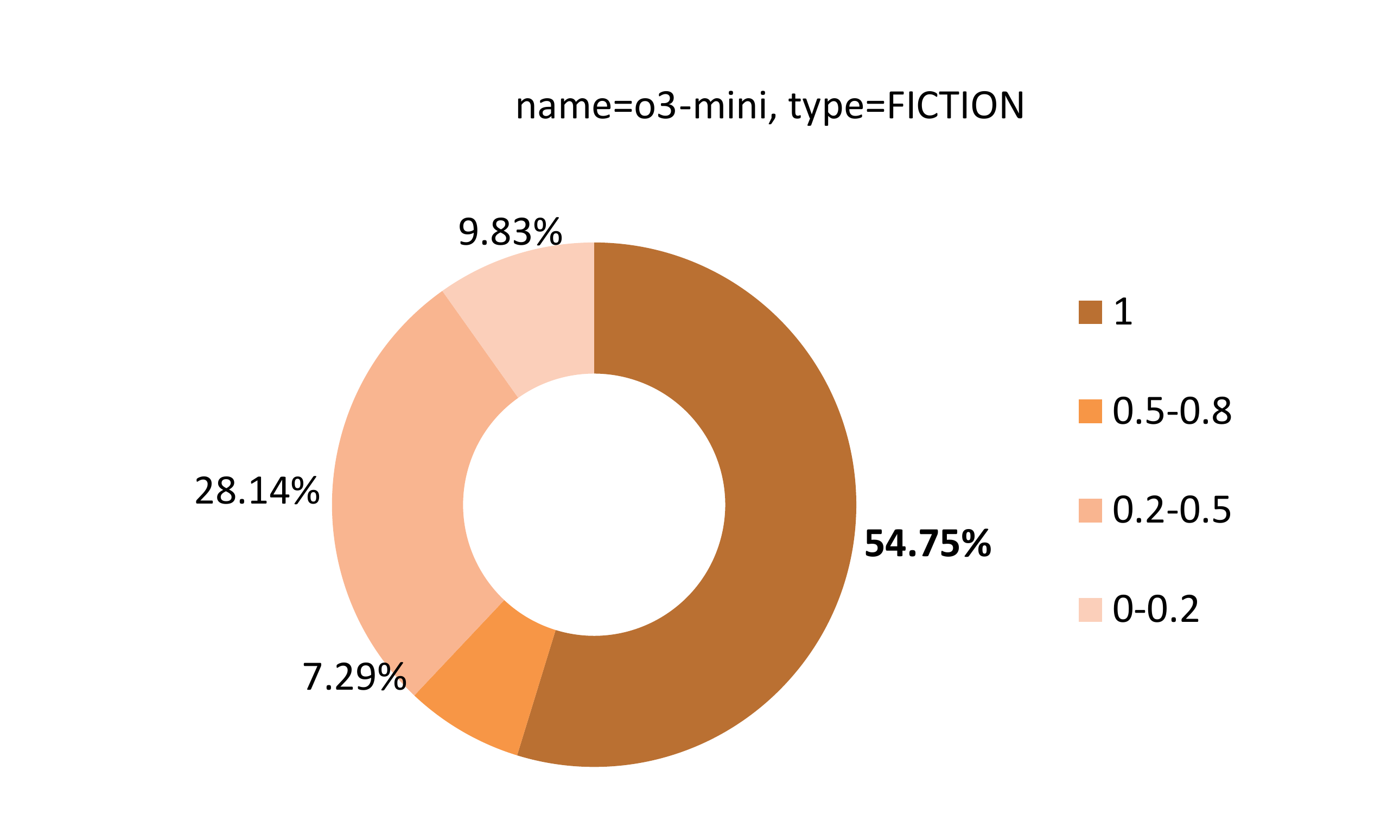}
                        \includegraphics[width=0.5\columnwidth]{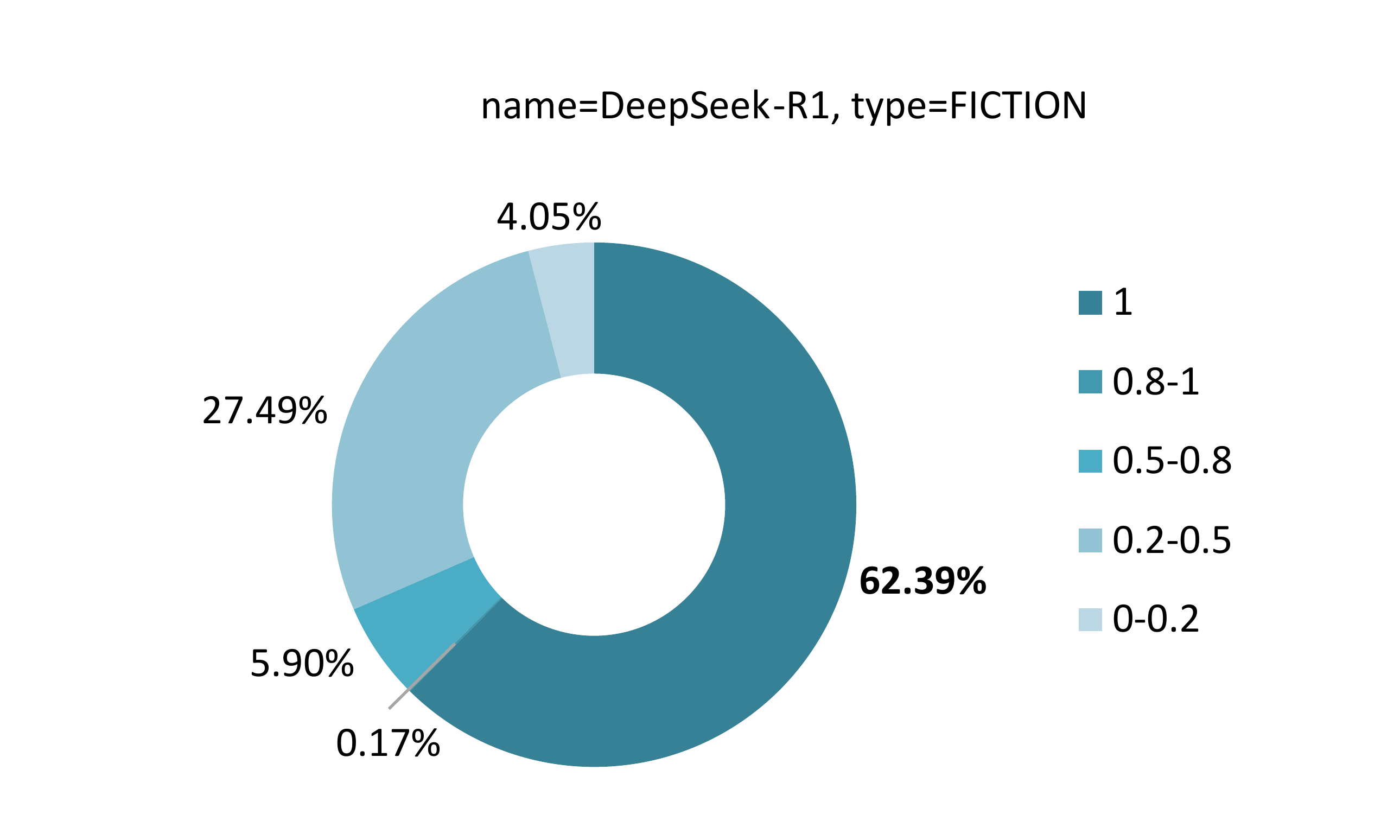}
  \caption{The Anchor Scores distributions of literal-metaphor (LM) word imagination task with sentences in FICTION on every model (the largest portion is in bold and the second largest is underlined).}
  \label{fig:result_imagination6}
\end{figure*}

\begin{figure*}[h]
  \centering
  \includegraphics[width=0.5\columnwidth]{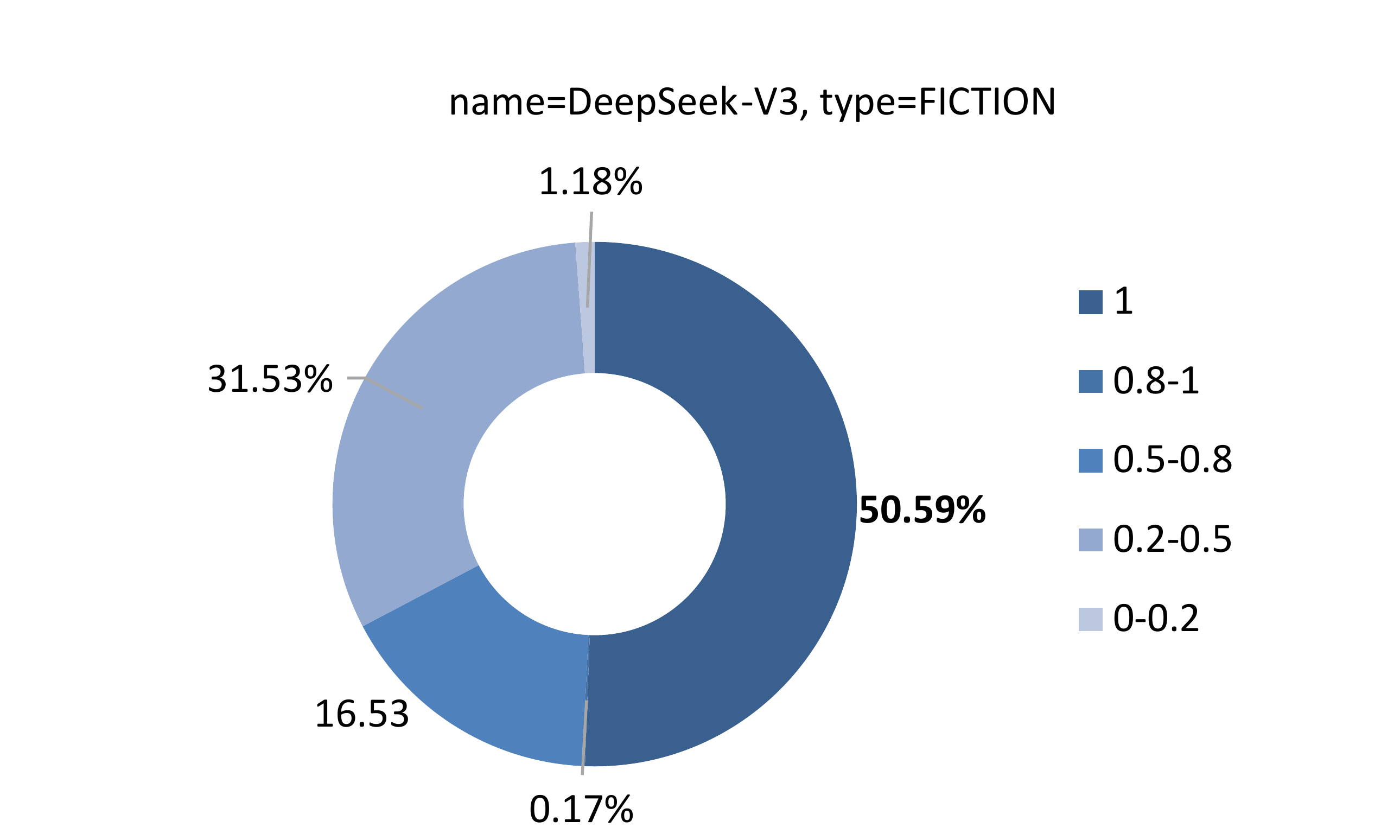}
      \includegraphics[width=0.5\columnwidth]{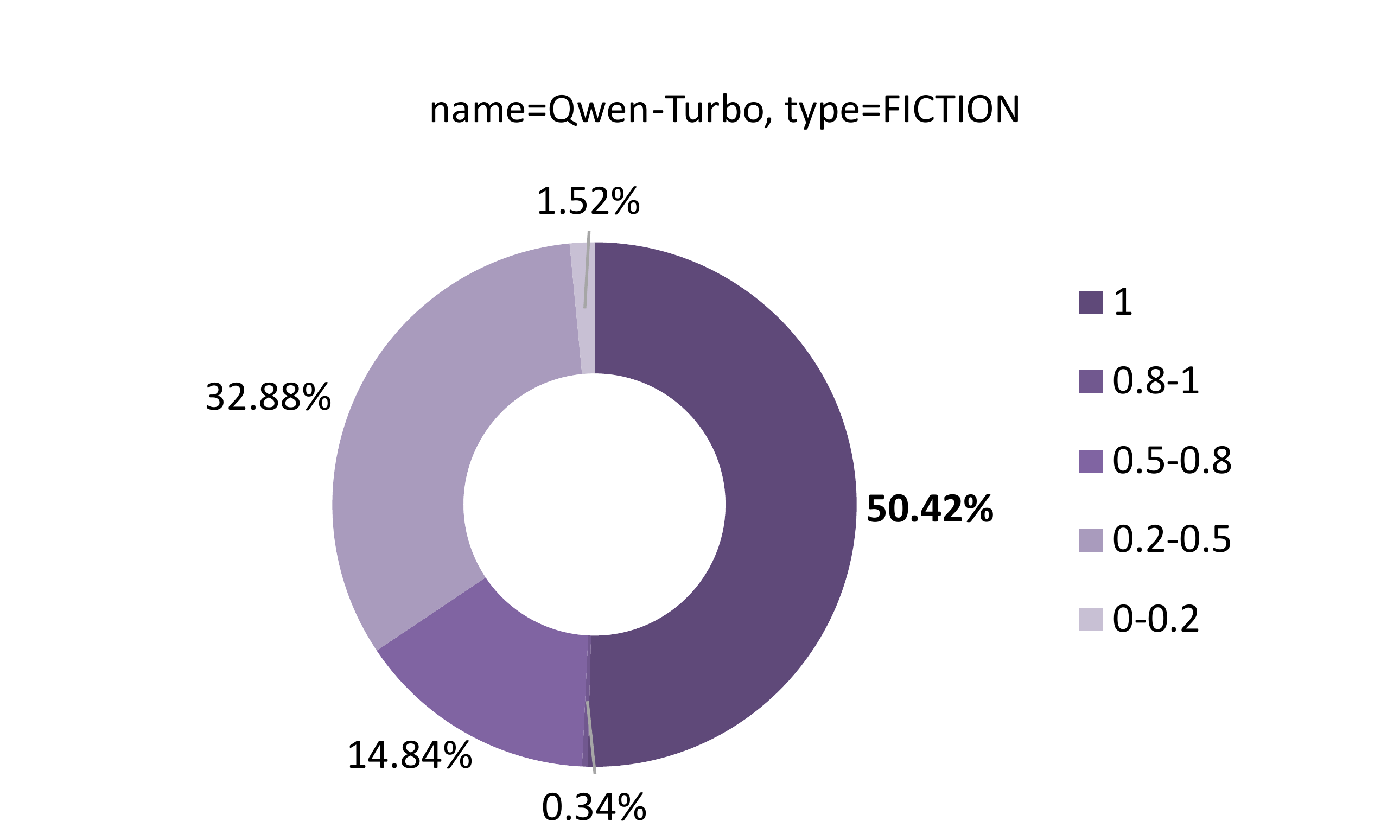}
            \includegraphics[width=0.5\columnwidth]{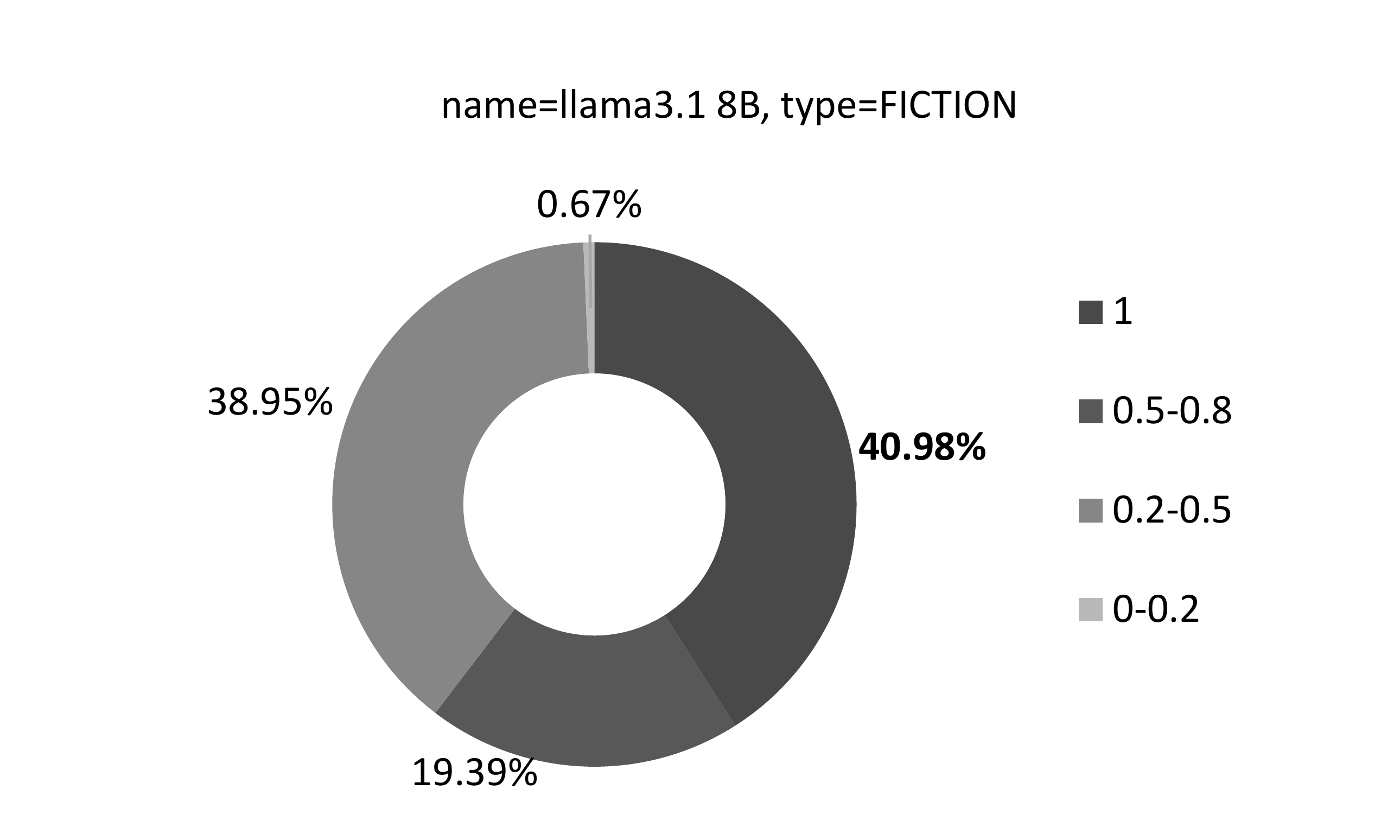}
                  \includegraphics[width=0.5\columnwidth]{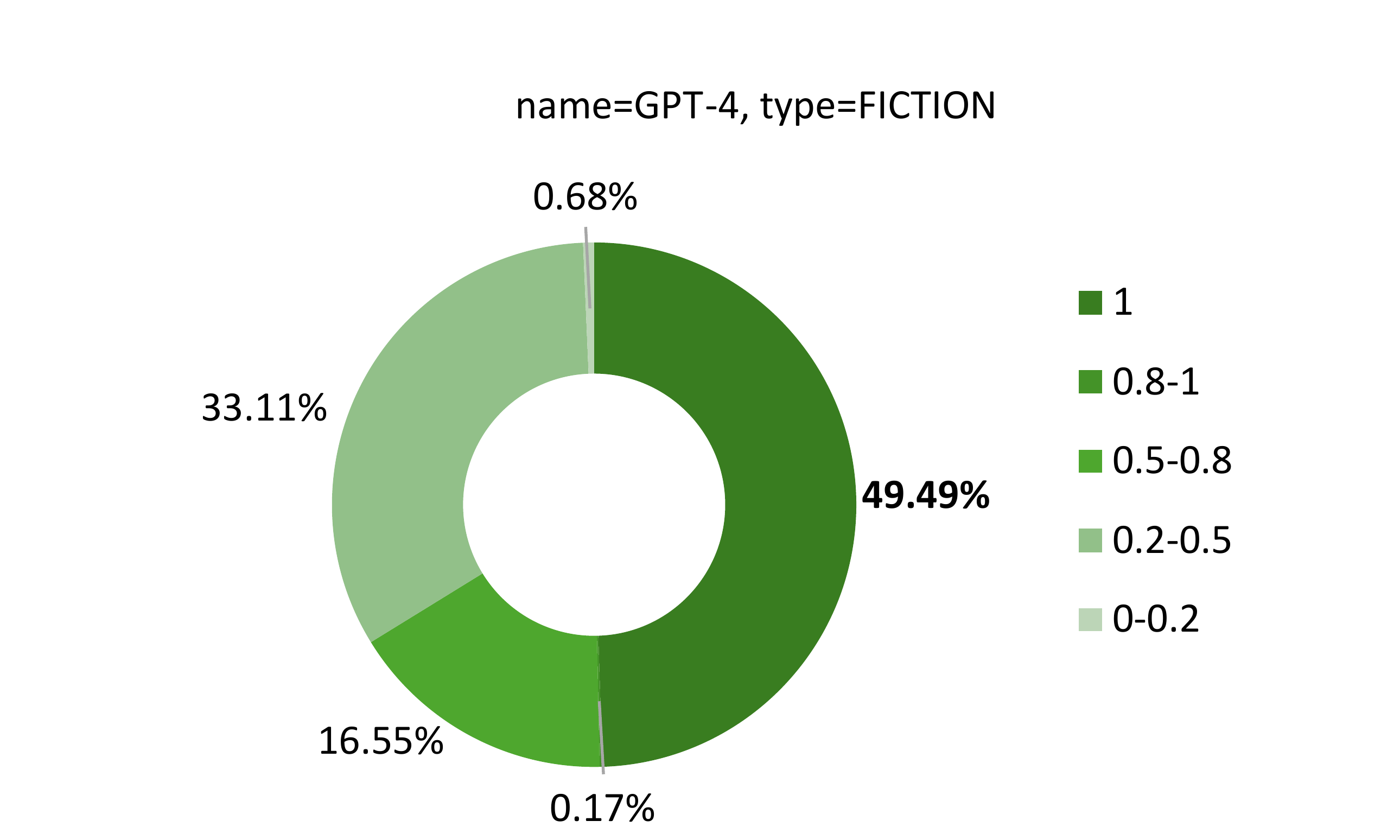}
                        \includegraphics[width=0.5\columnwidth]{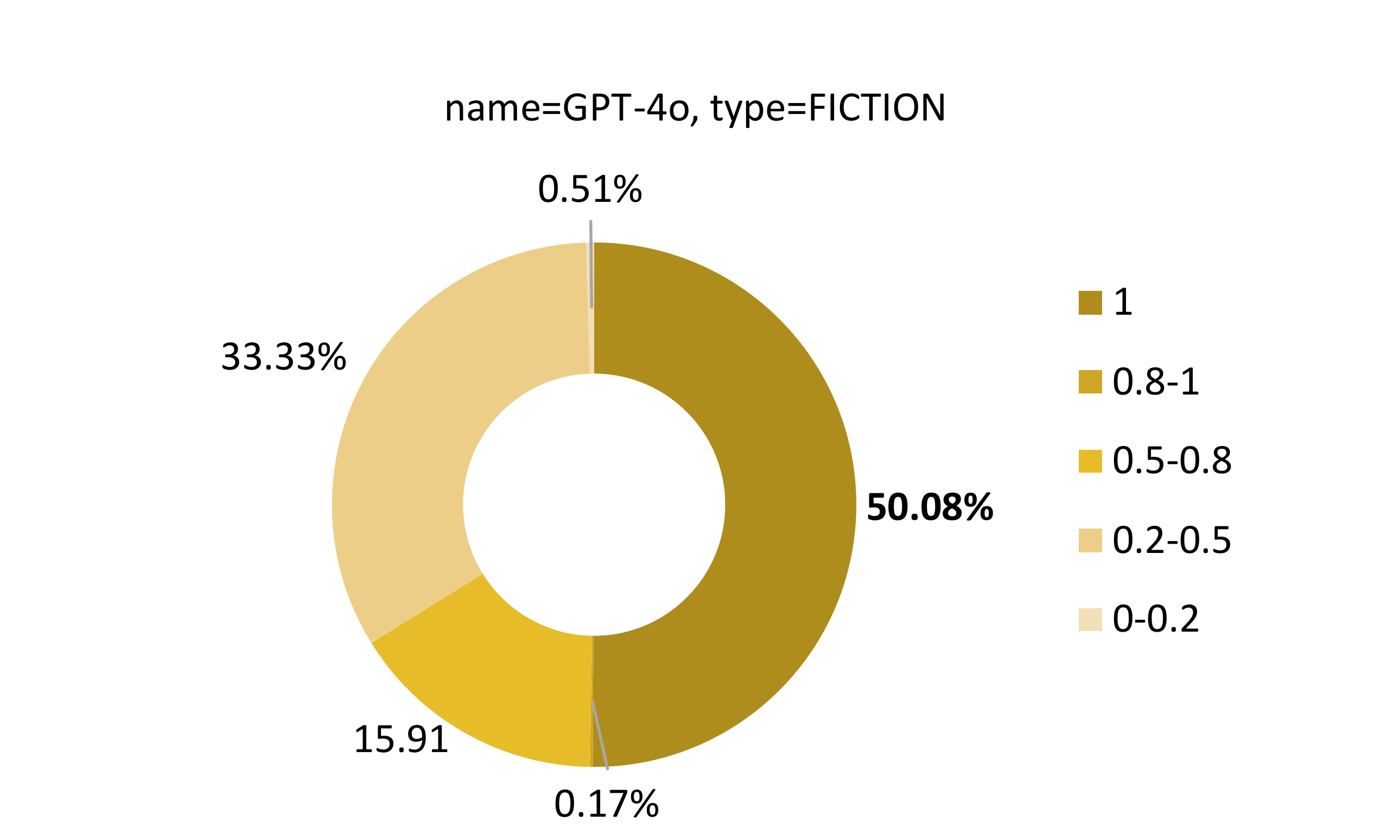}
                        \includegraphics[width=0.5\columnwidth]{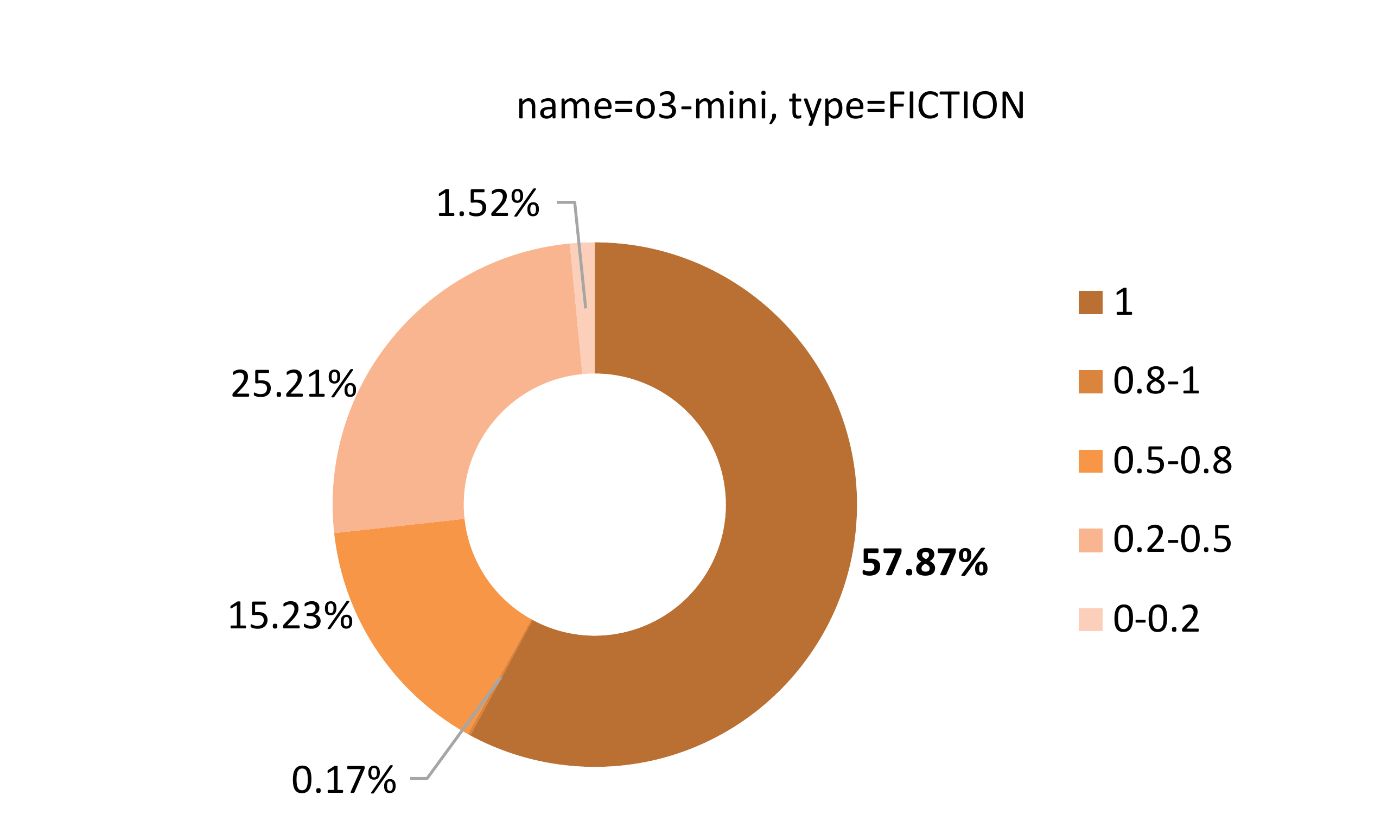}
                        \includegraphics[width=0.5\columnwidth]{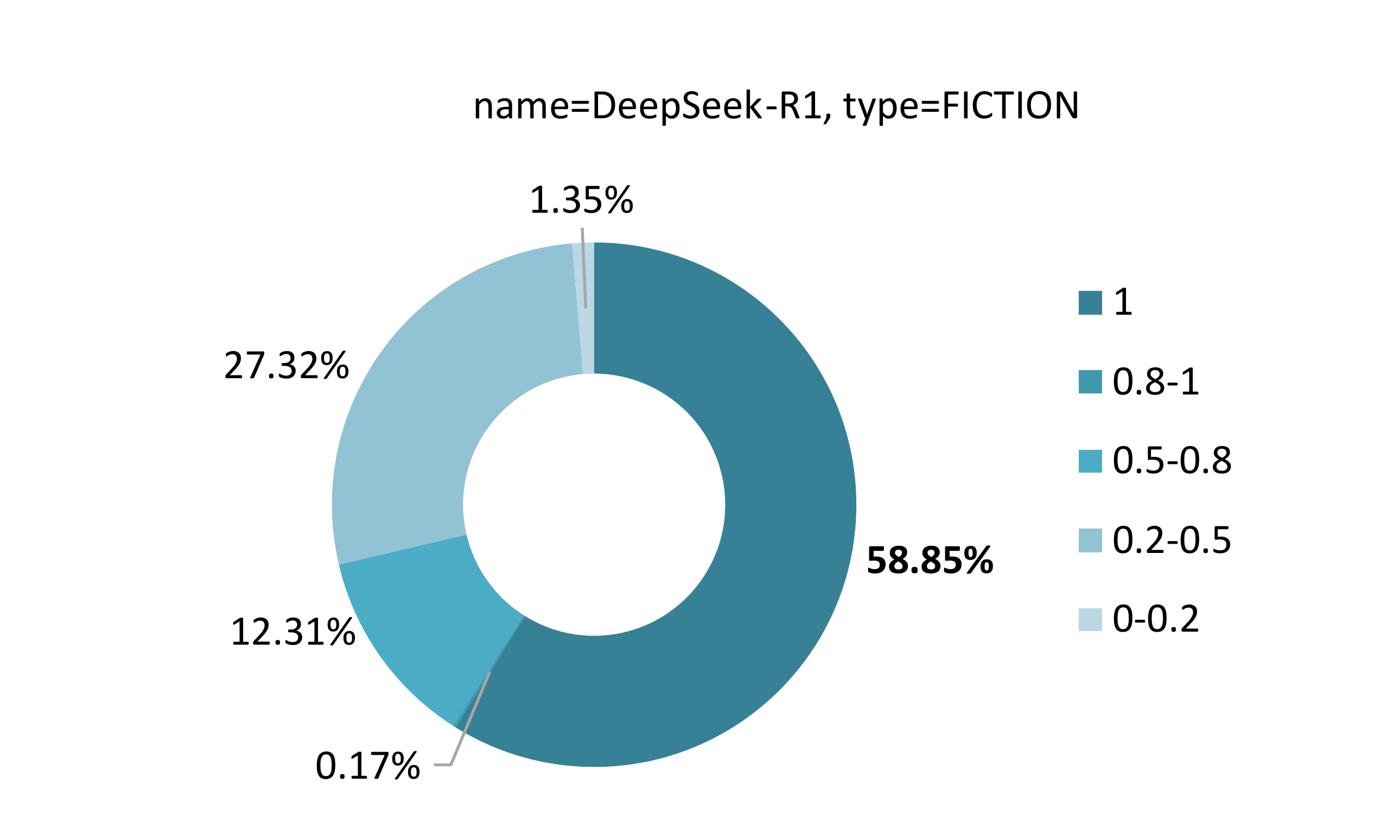}
  \caption{The Anchor Scores distributions of metaphor-literal (ML) word imagination task with sentences in FICTION on every model (the largest portion is in bold and the second largest is underlined).}
  \label{fig:result_imagination7}
\end{figure*}

\begin{figure*}[h]
  \centering
  \includegraphics[width=0.5\columnwidth]{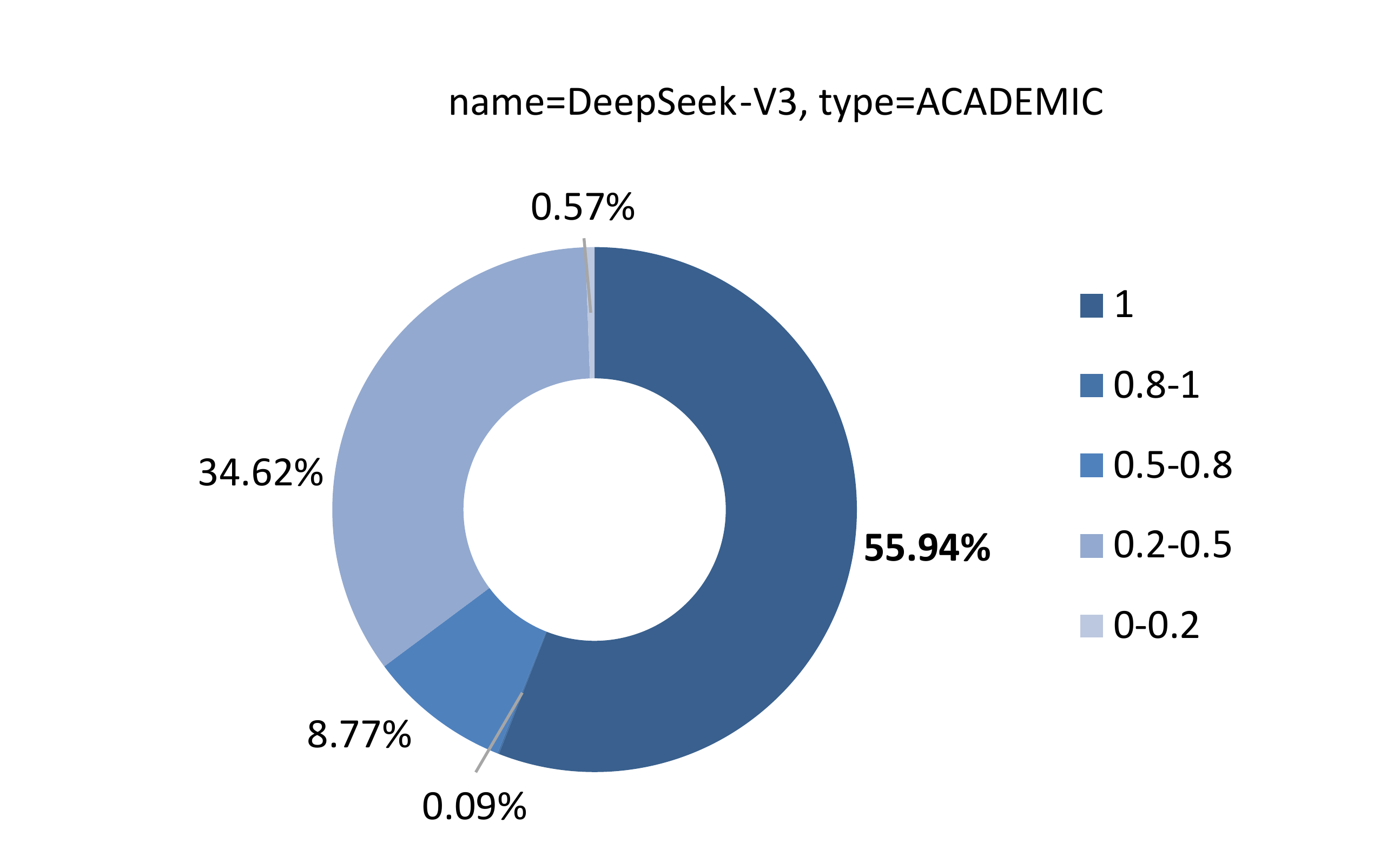}
      \includegraphics[width=0.5\columnwidth]{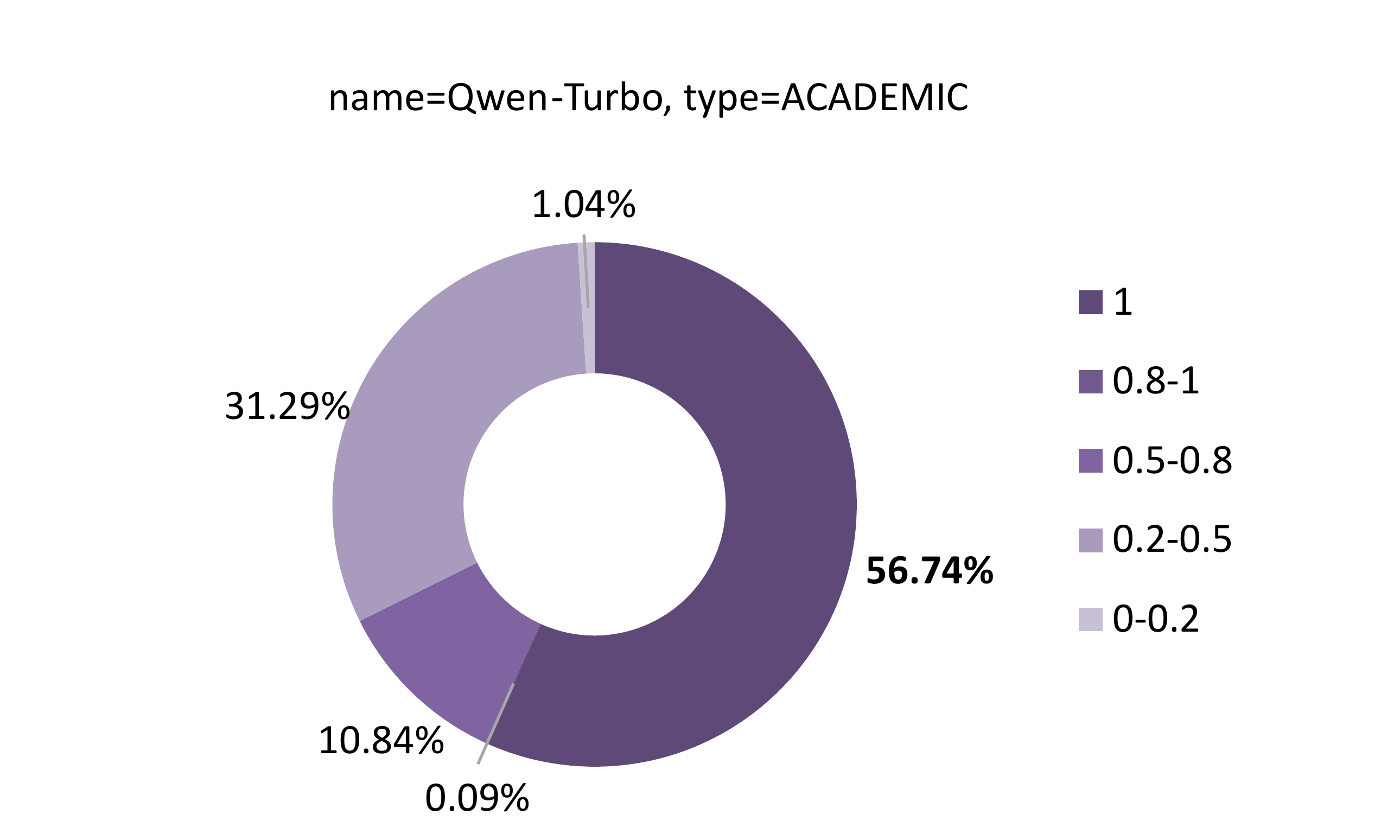}
            \includegraphics[width=0.5\columnwidth]{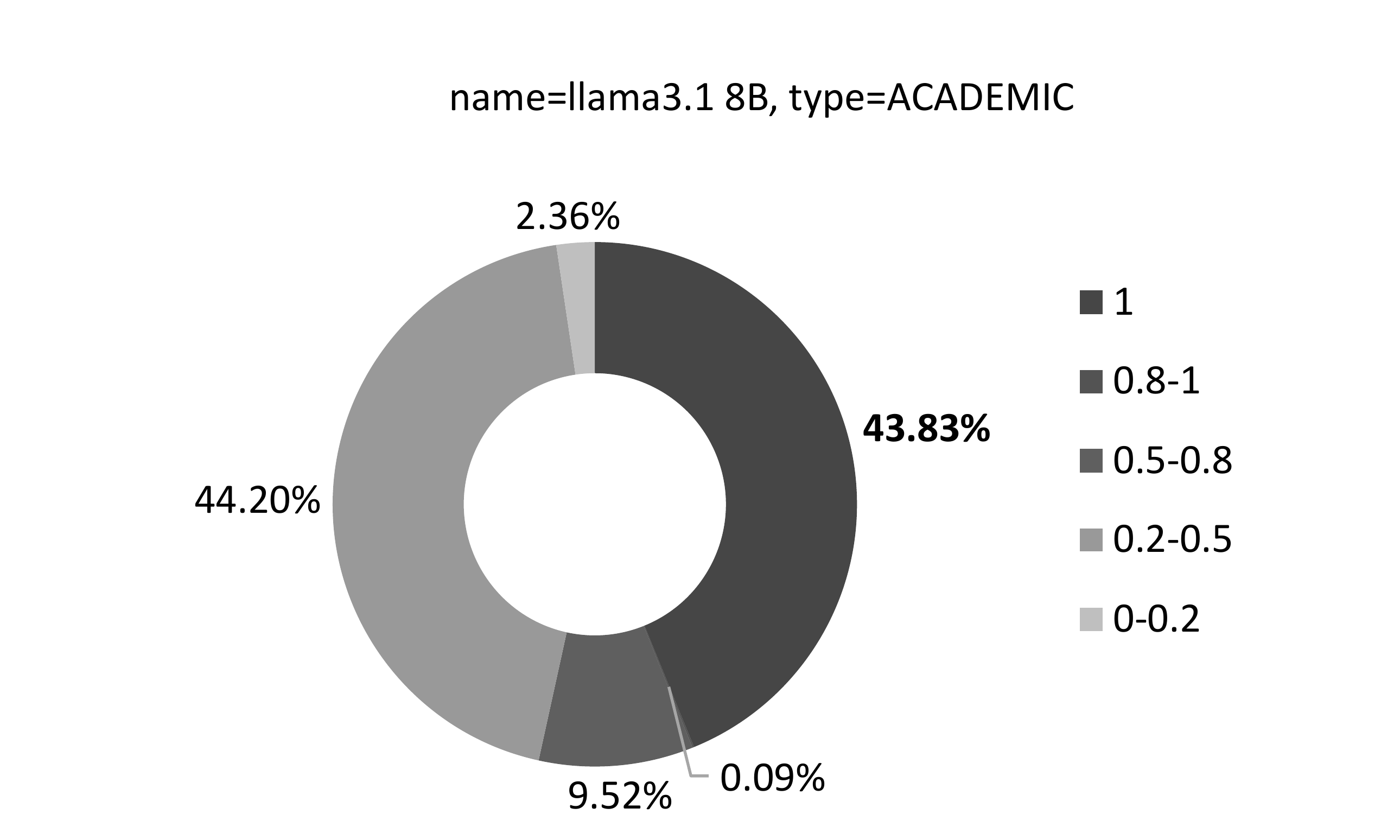}
                  \includegraphics[width=0.5\columnwidth]{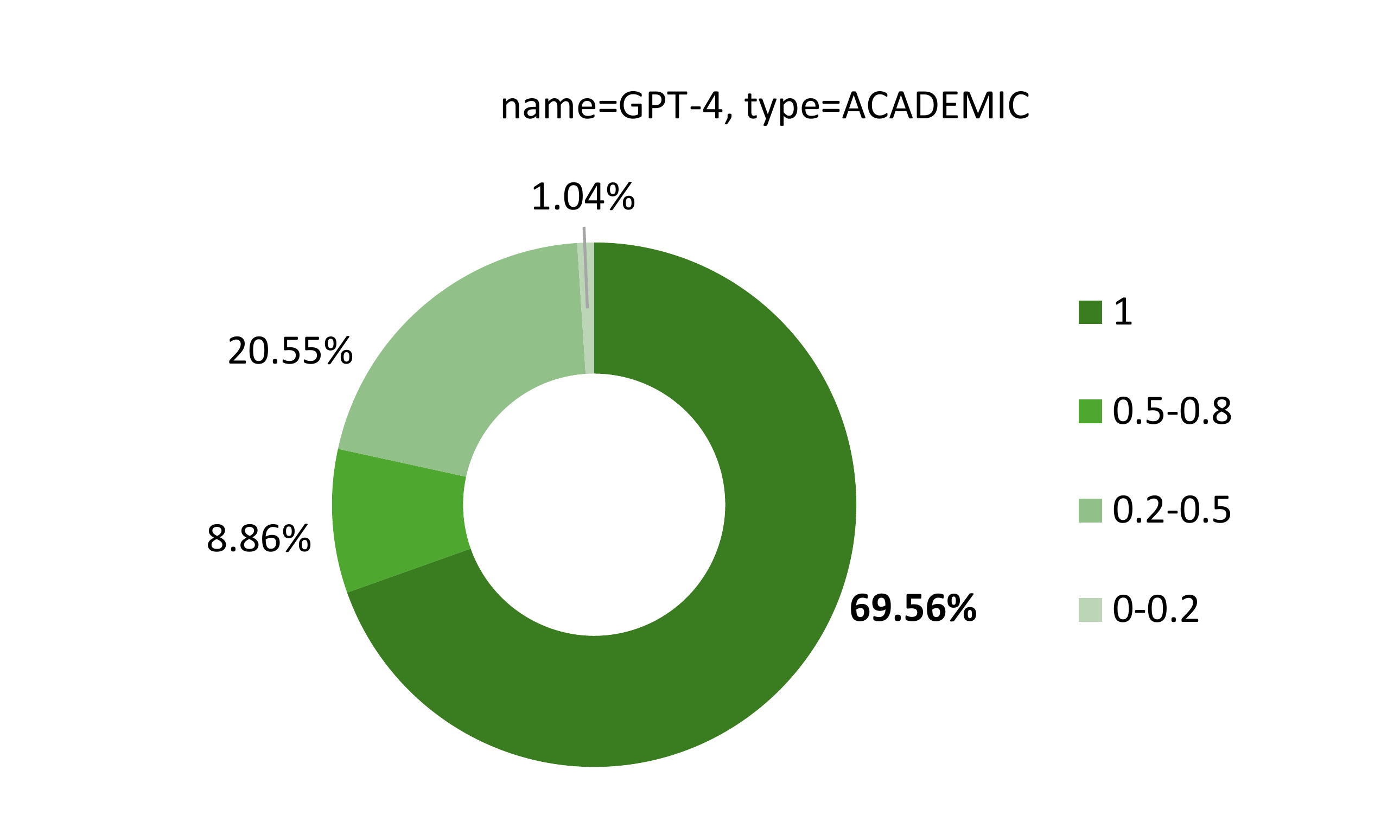}
                        \includegraphics[width=0.5\columnwidth]{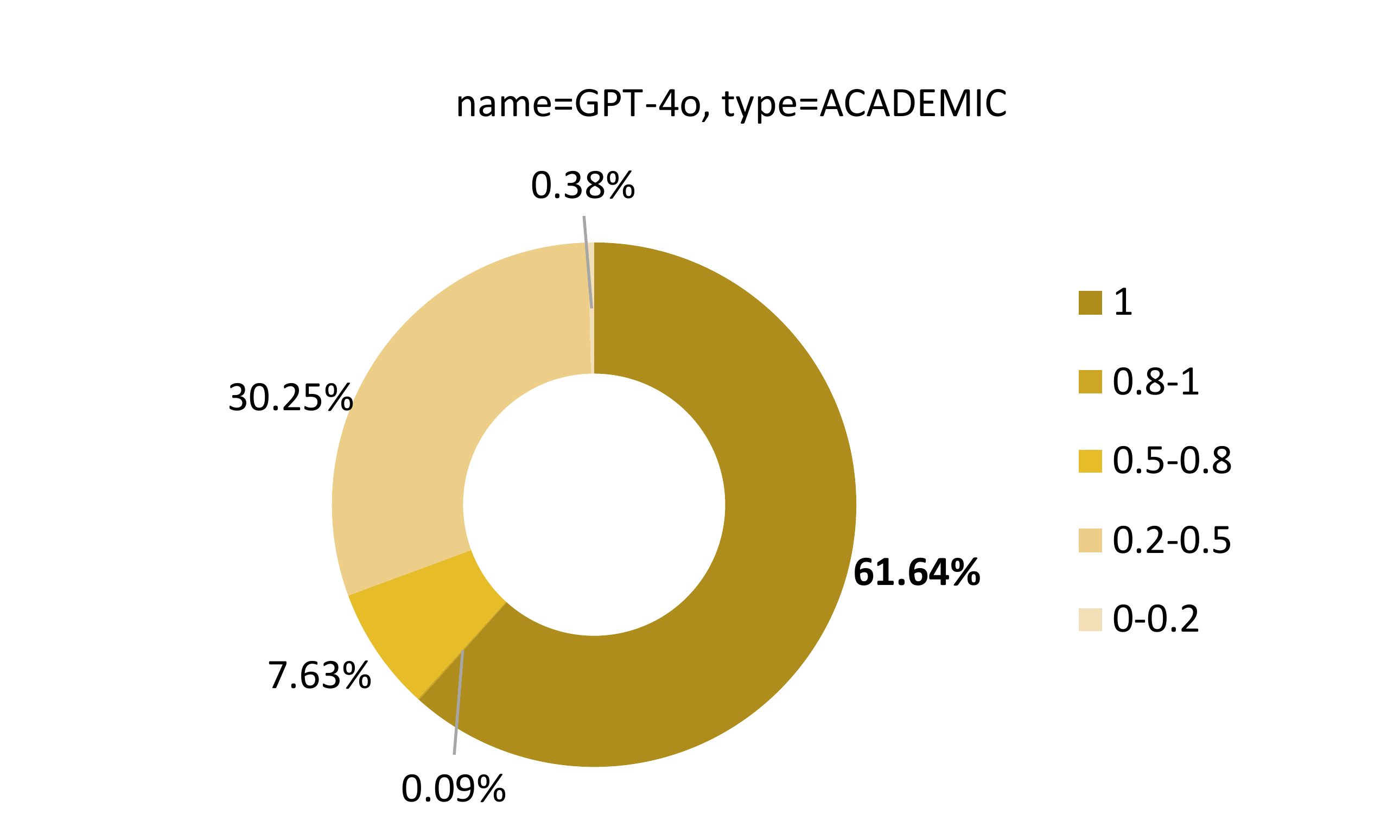}
                        \includegraphics[width=0.5\columnwidth]{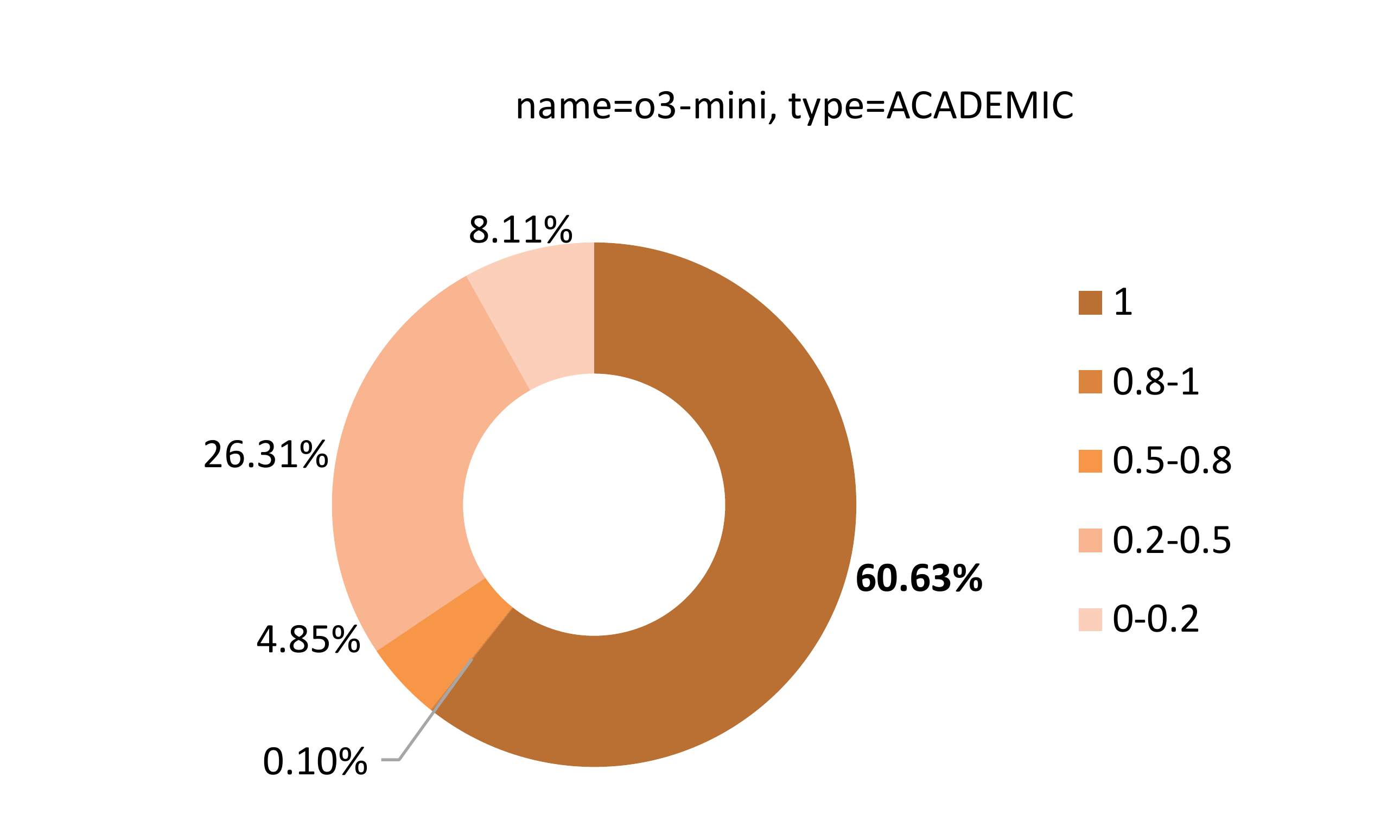}
                        \includegraphics[width=0.5\columnwidth]{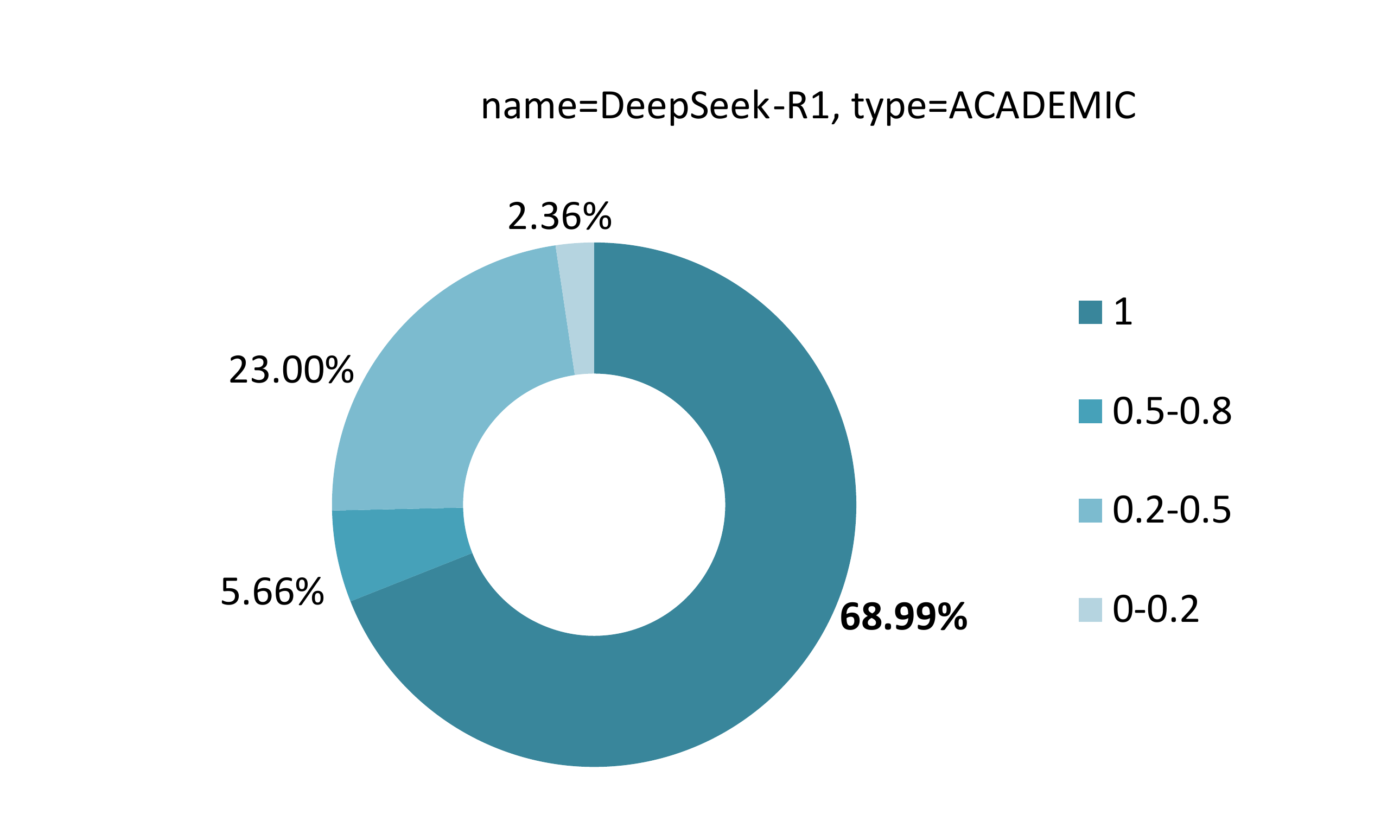}
  \caption{The Anchor Scores distributions of literal-metaphor (LM) word imagination task with sentences in ACADEMIC on every model (the largest portion is in bold and the second largest is underlined).}
  \label{fig:result_imagination8}
\end{figure*}

\begin{figure*}[h]
  \centering
  \includegraphics[width=0.5\columnwidth]{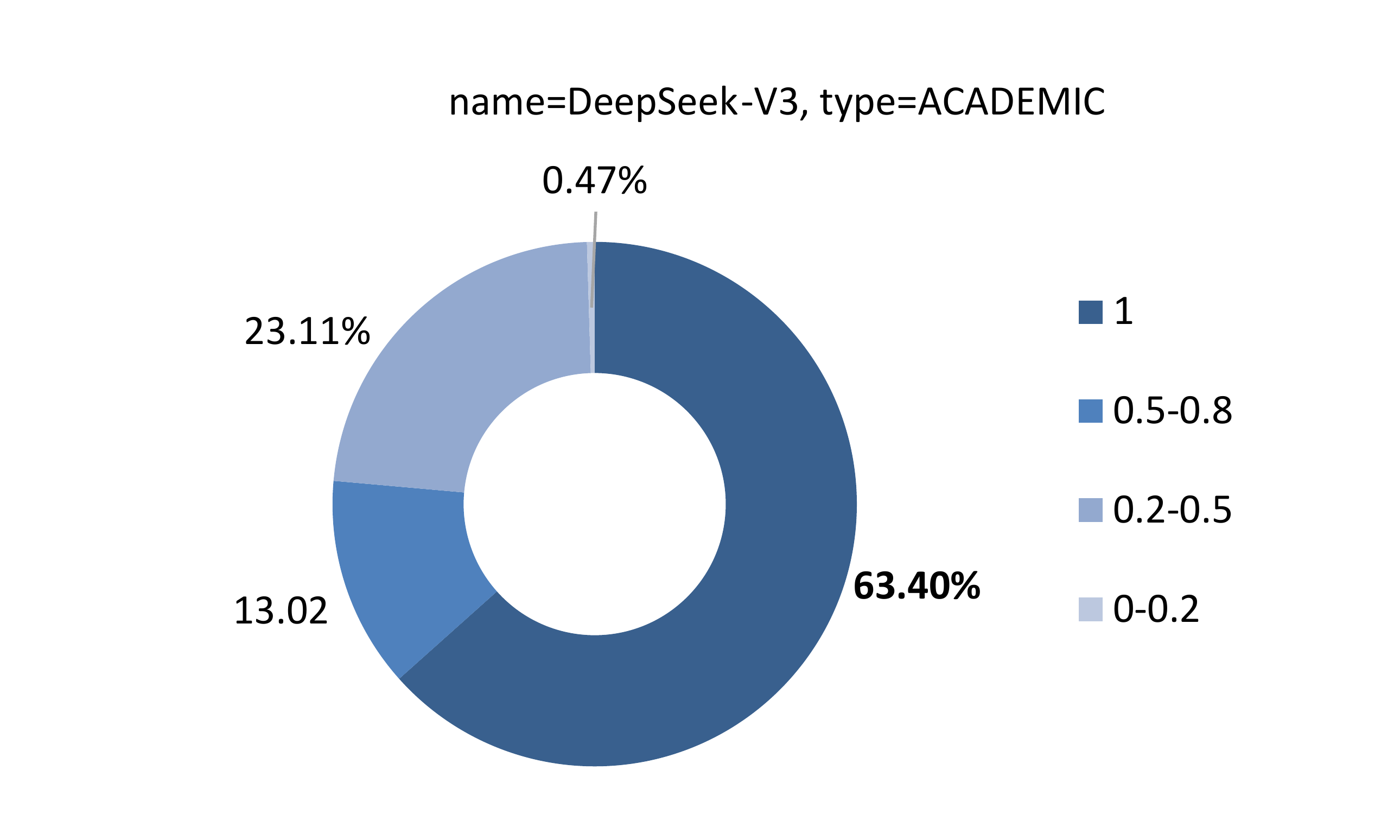}
      \includegraphics[width=0.5\columnwidth]{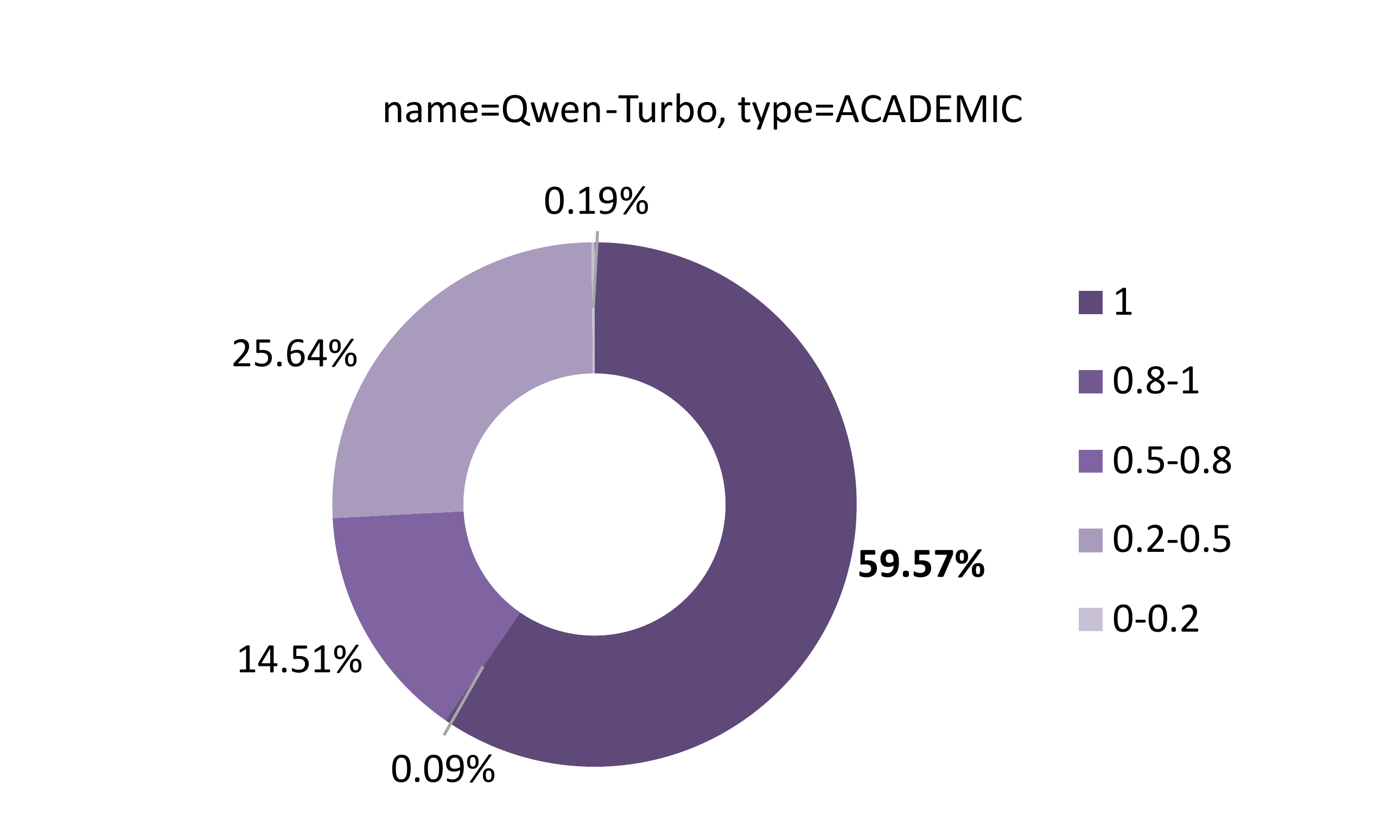}
            \includegraphics[width=0.5\columnwidth]{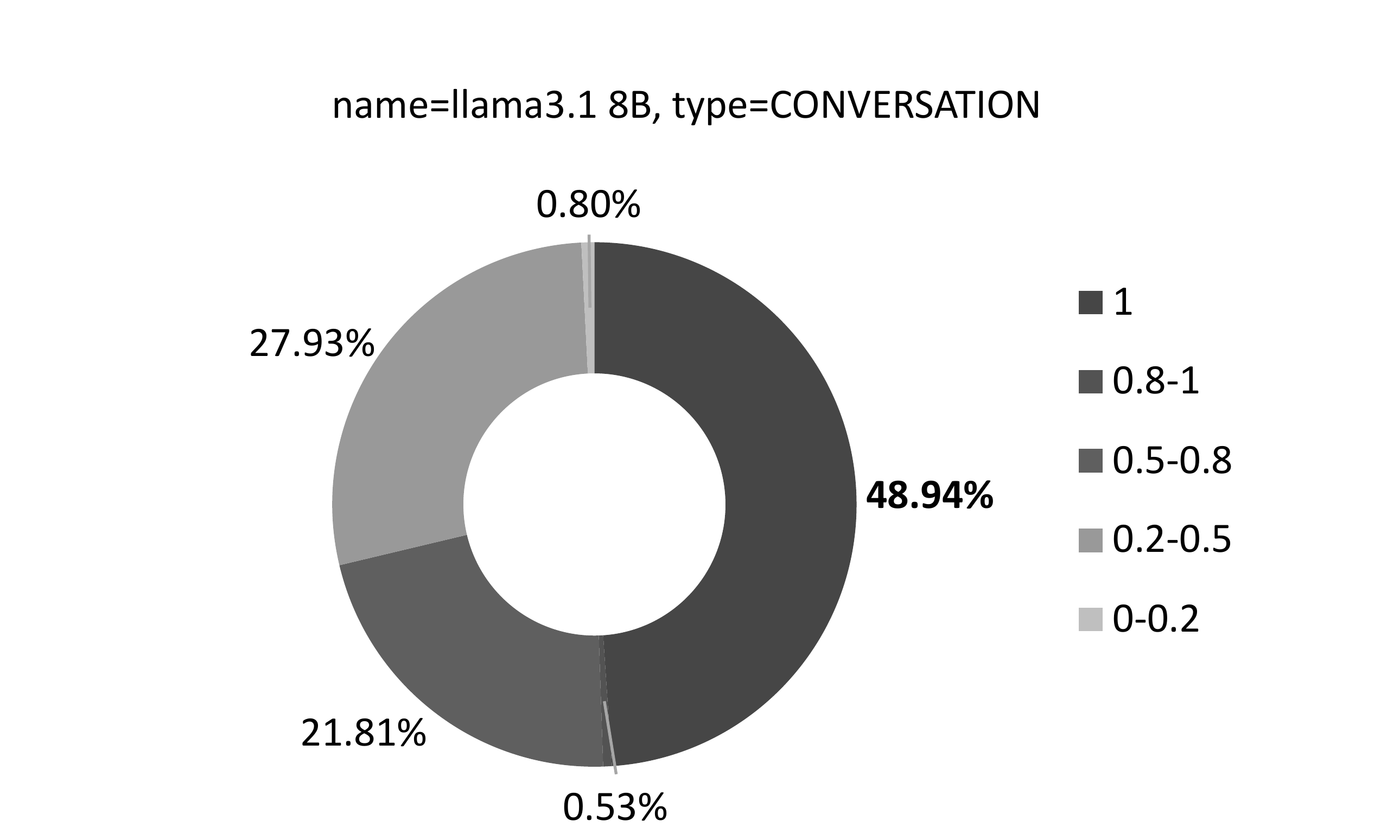}
                  \includegraphics[width=0.5\columnwidth]{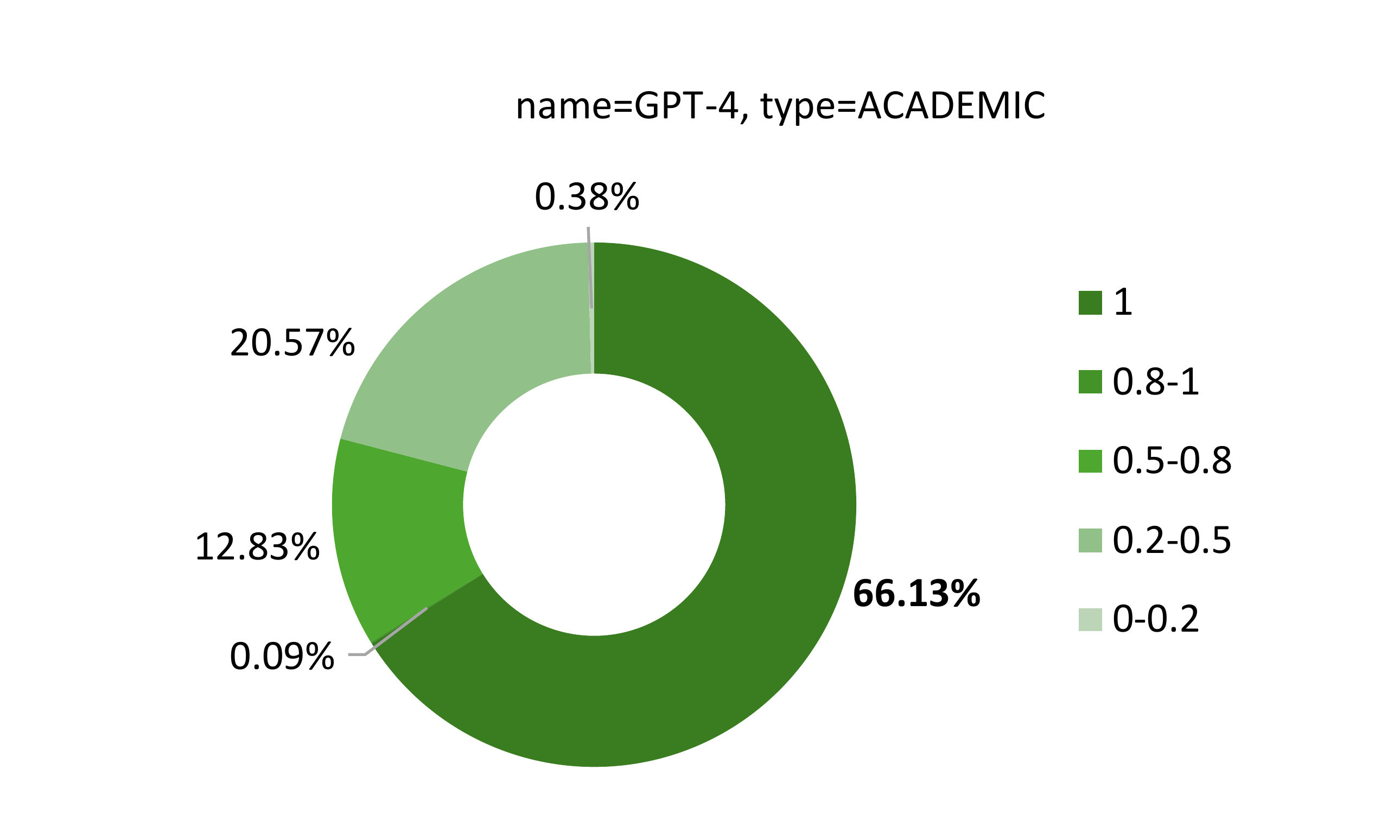}
                        \includegraphics[width=0.5\columnwidth]{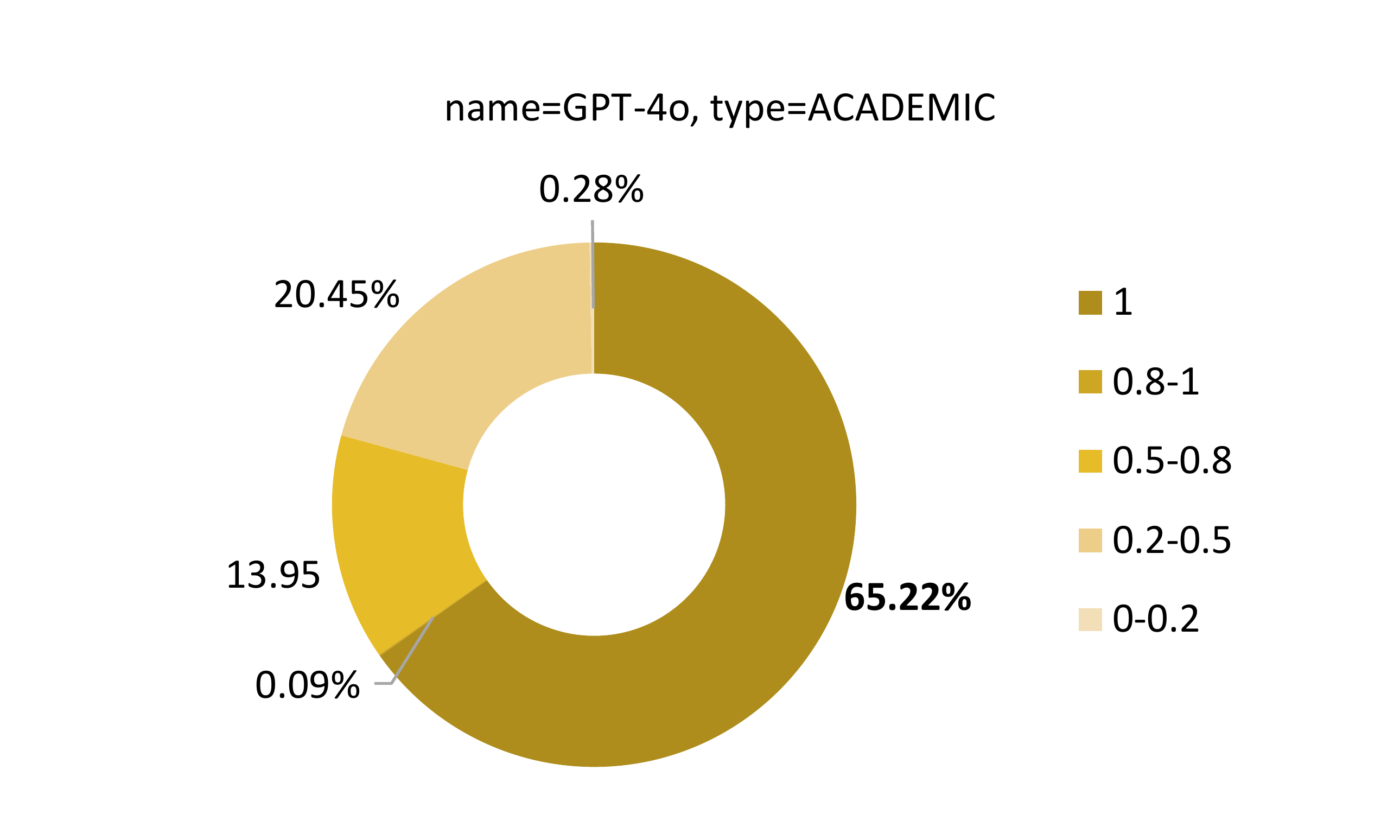}
                        \includegraphics[width=0.5\columnwidth]{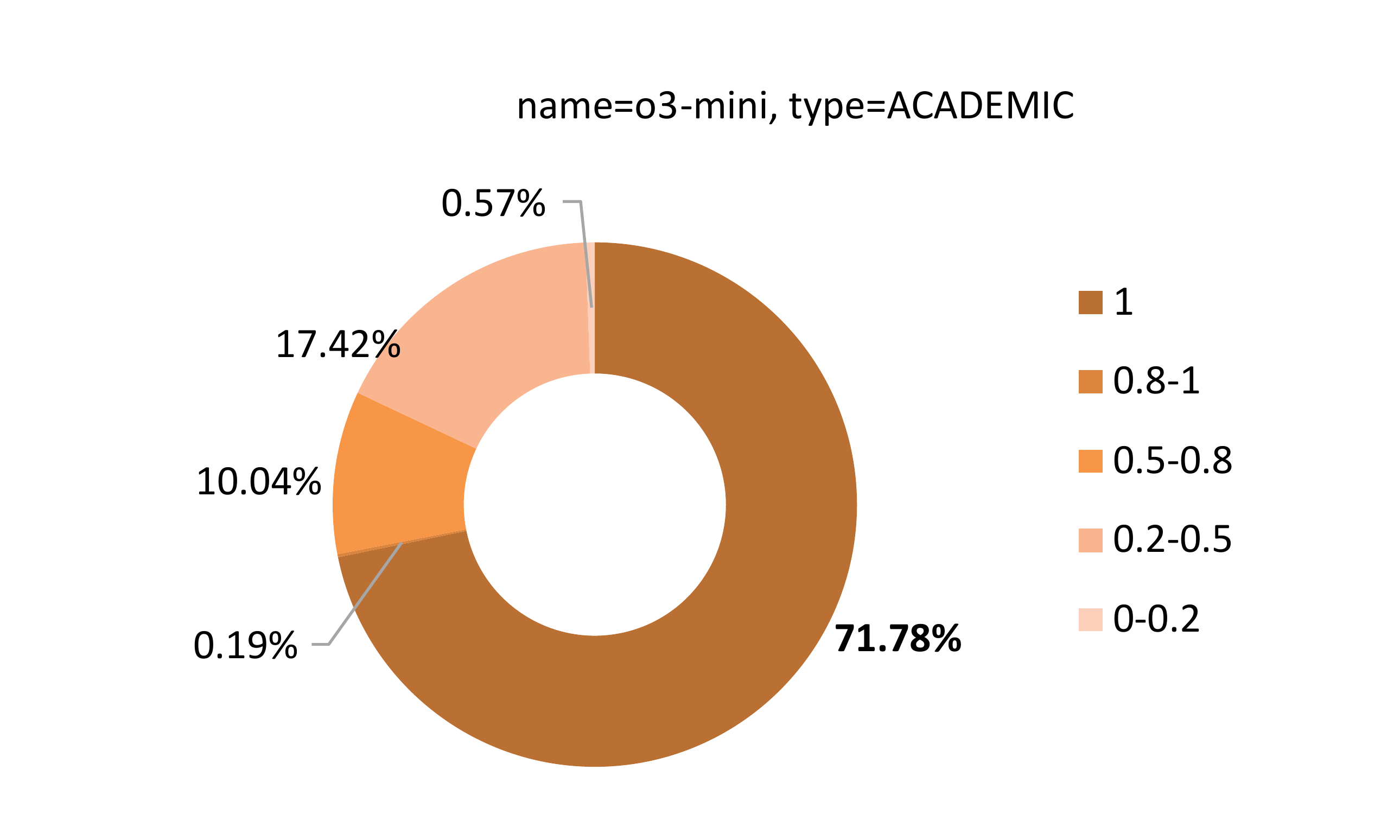}
                        \includegraphics[width=0.5\columnwidth]{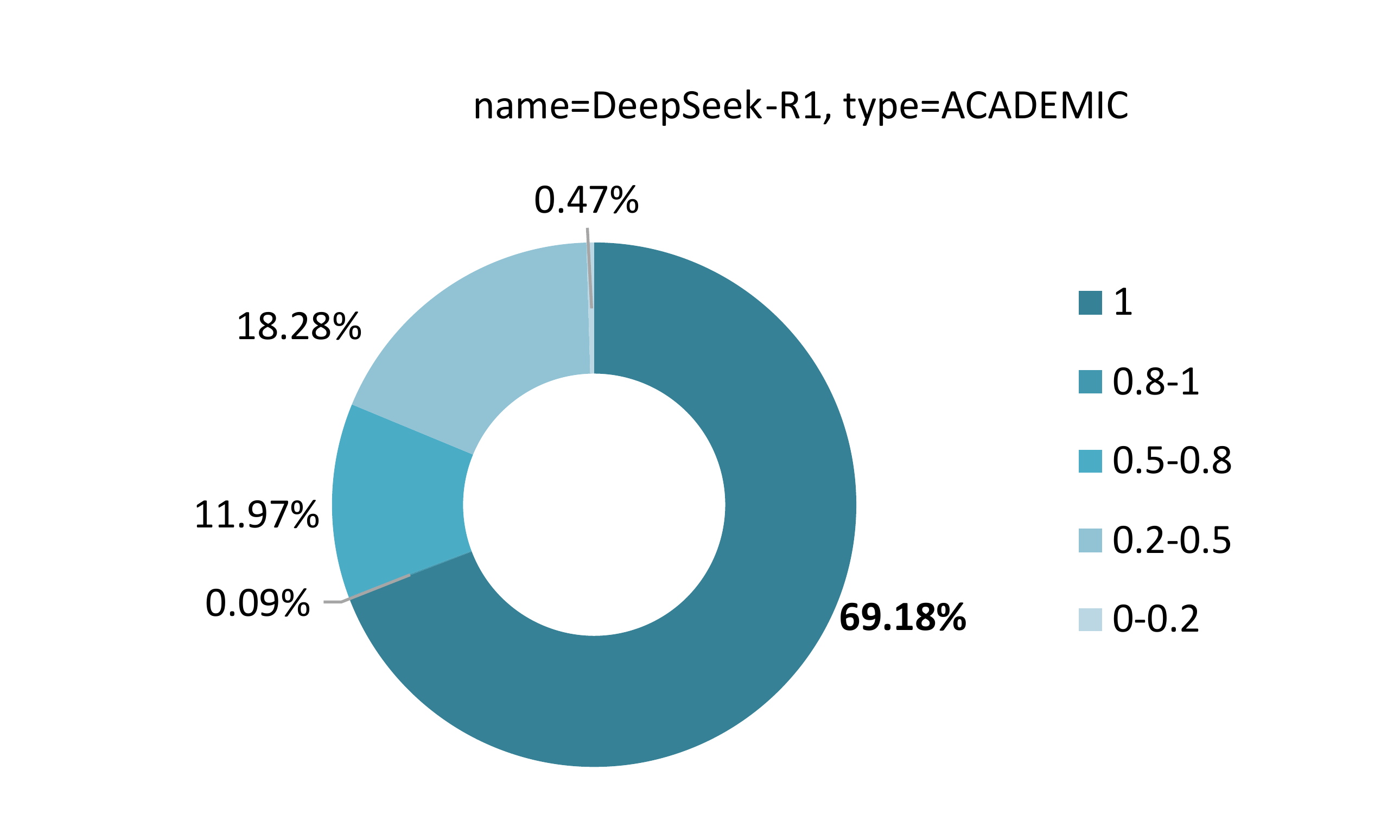}
  \caption{The Anchor Scores distributions of metaphor-literal (ML) word imagination task with sentences in ACADEMIC on every model (the largest portion is in bold and the second largest is underlined).}
  \label{fig:result_imagination9}
\end{figure*}

\begin{figure*}[h]
  \centering
  \includegraphics[width=0.5\columnwidth]{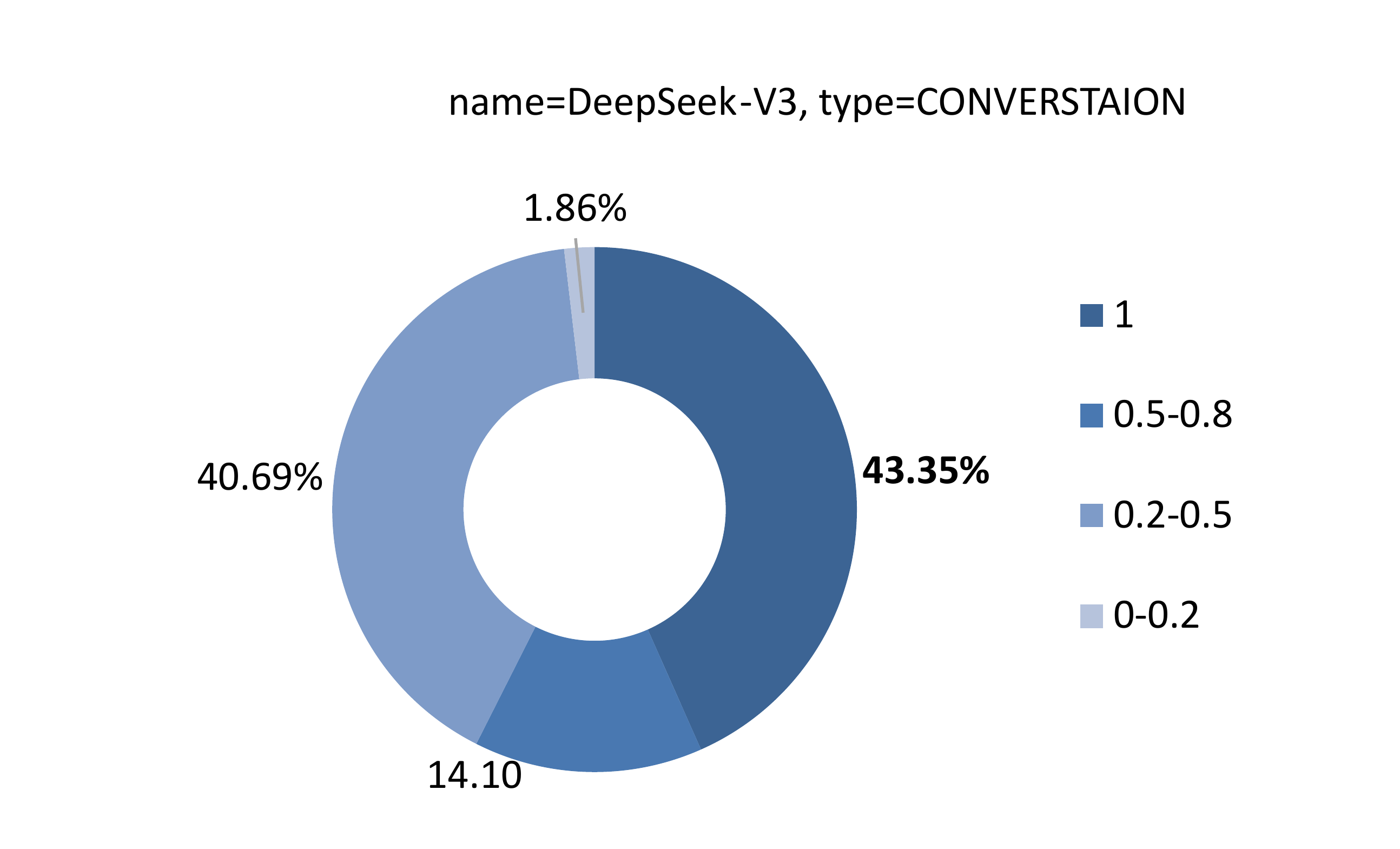}
      \includegraphics[width=0.5\columnwidth]{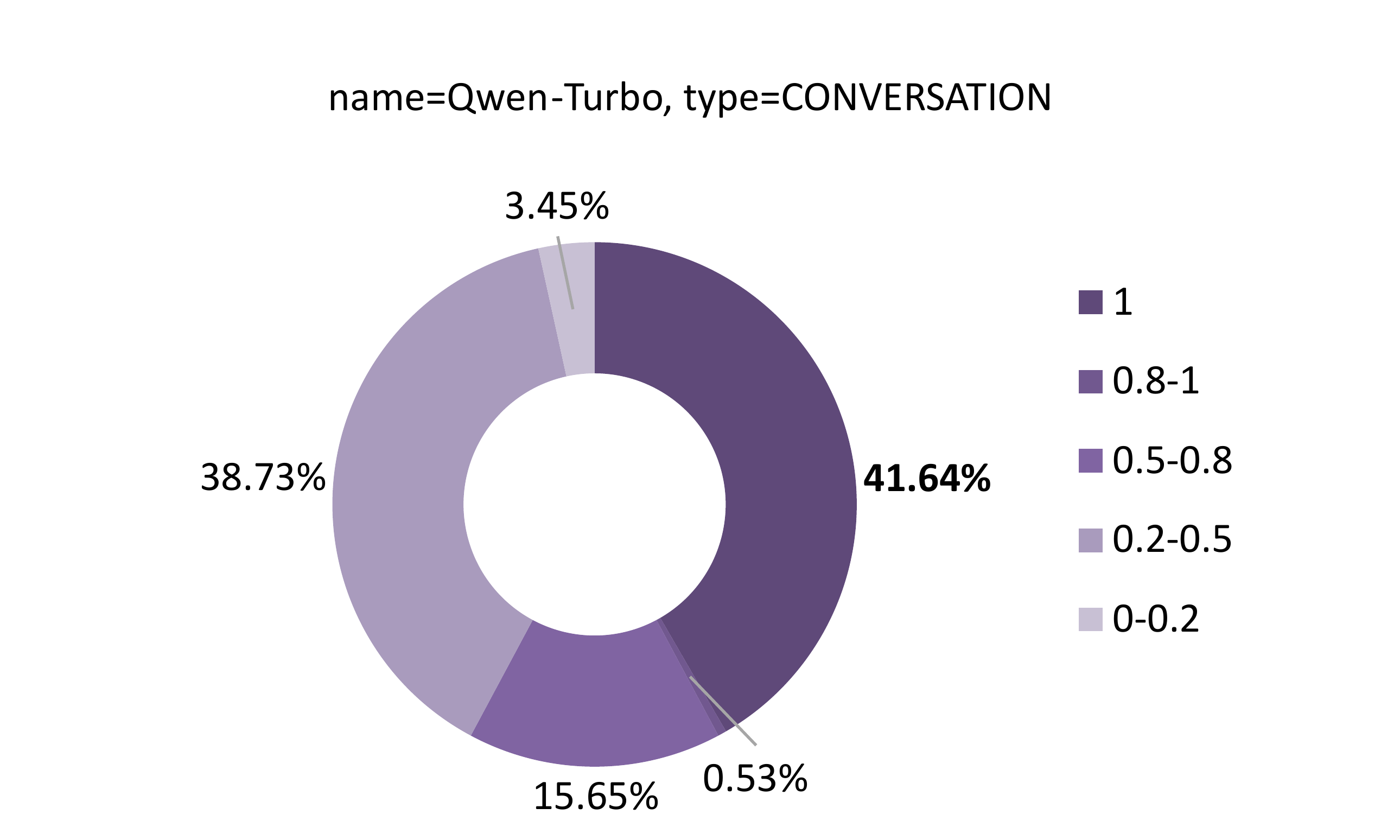}
            \includegraphics[width=0.5\columnwidth]{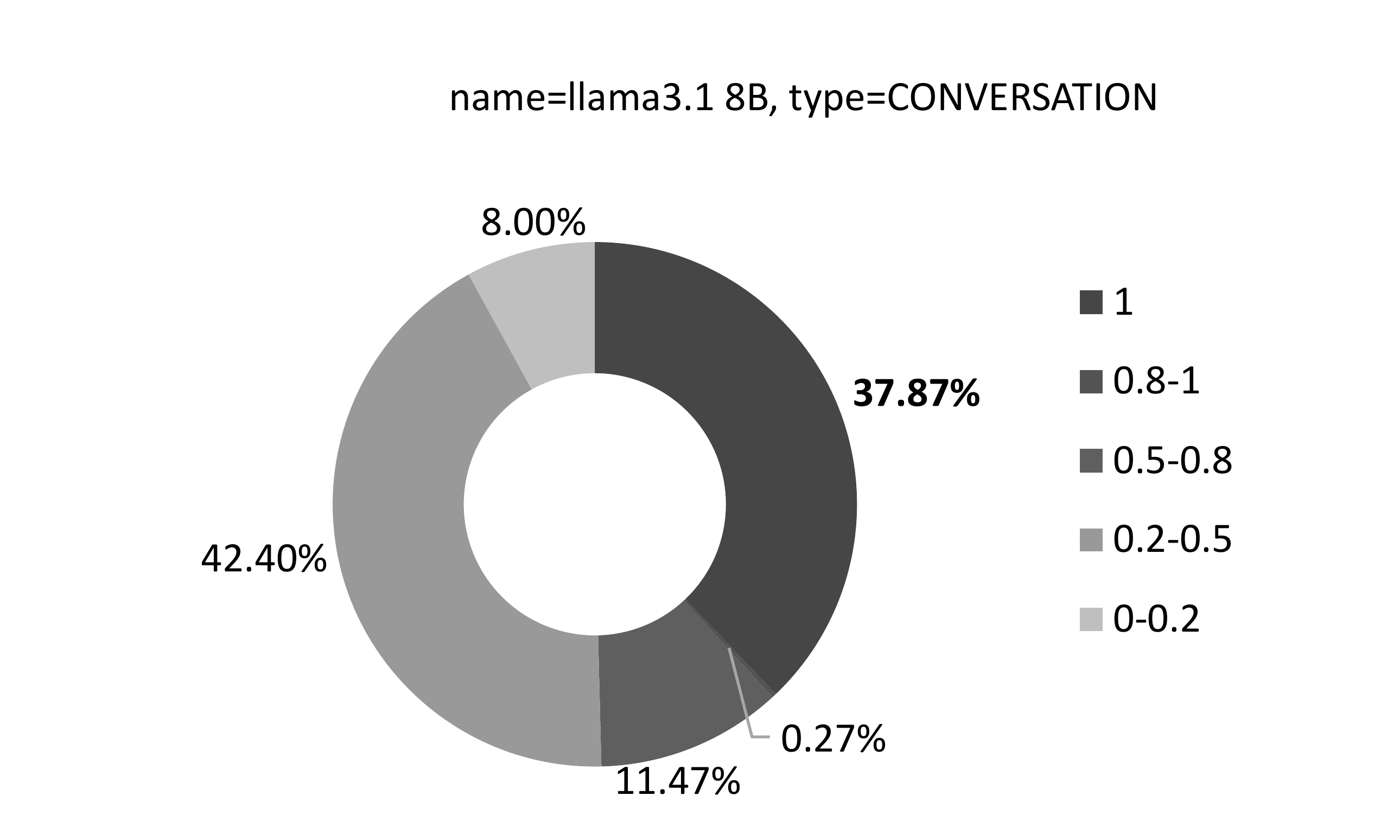}
                  \includegraphics[width=0.5\columnwidth]{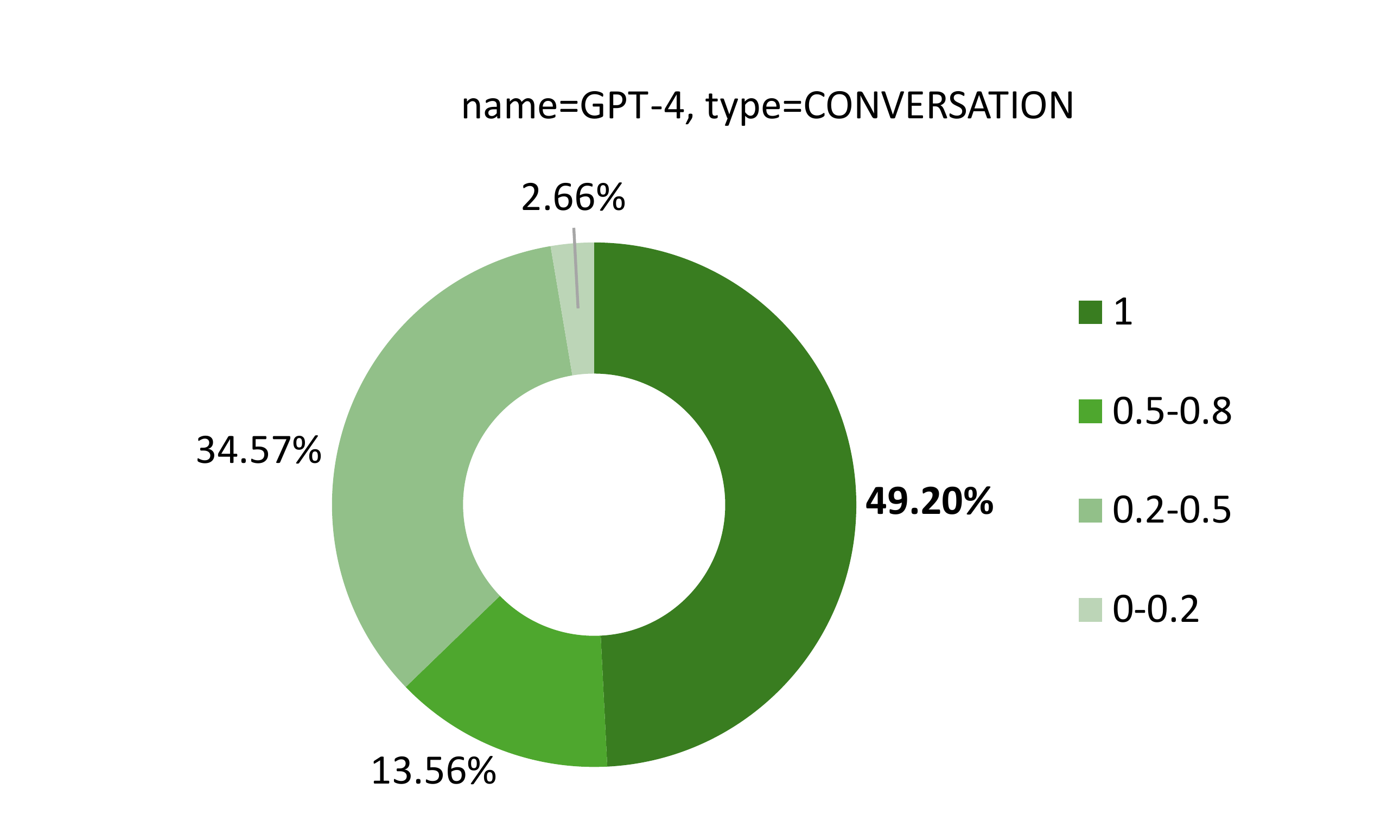}
                        \includegraphics[width=0.5\columnwidth]{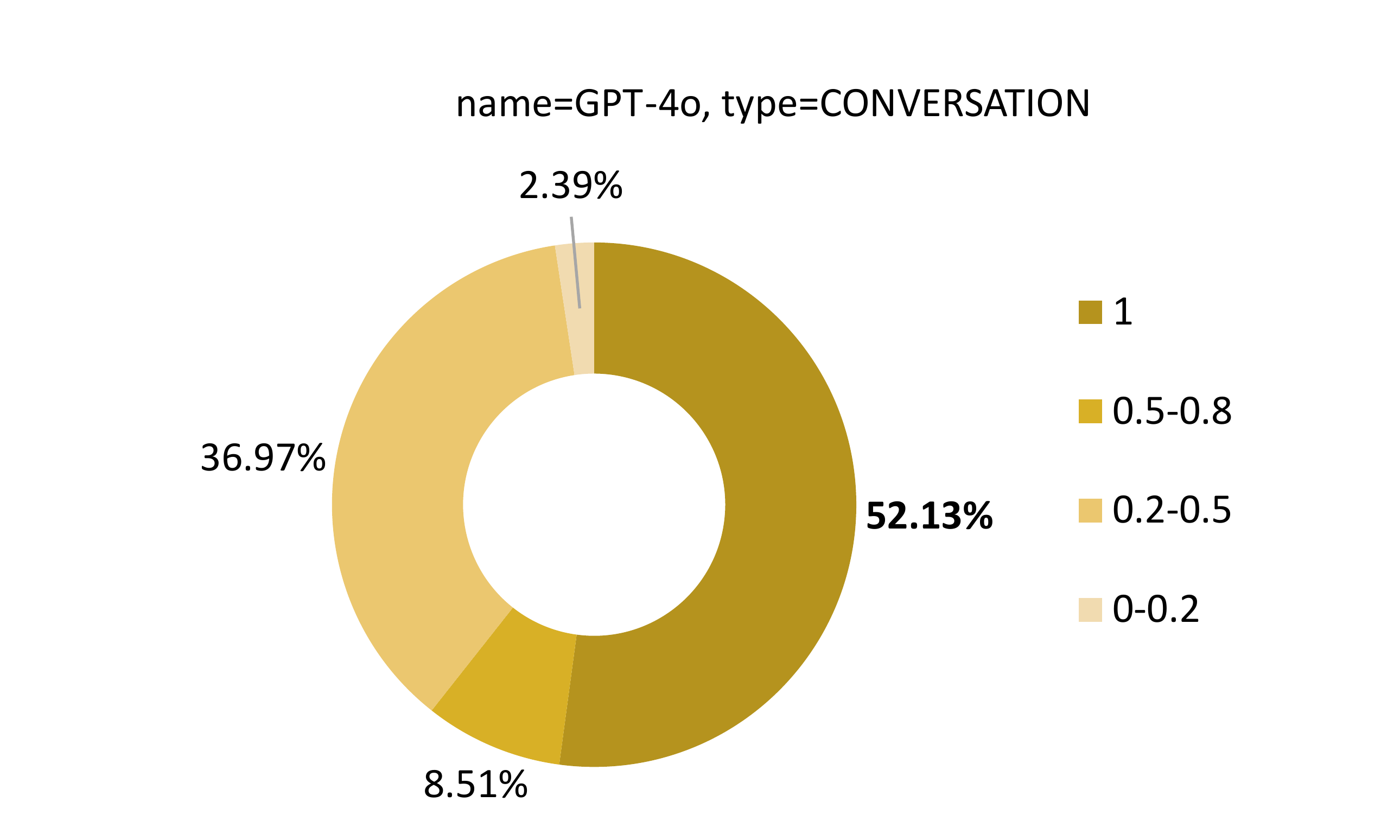}
                        \includegraphics[width=0.5\columnwidth]{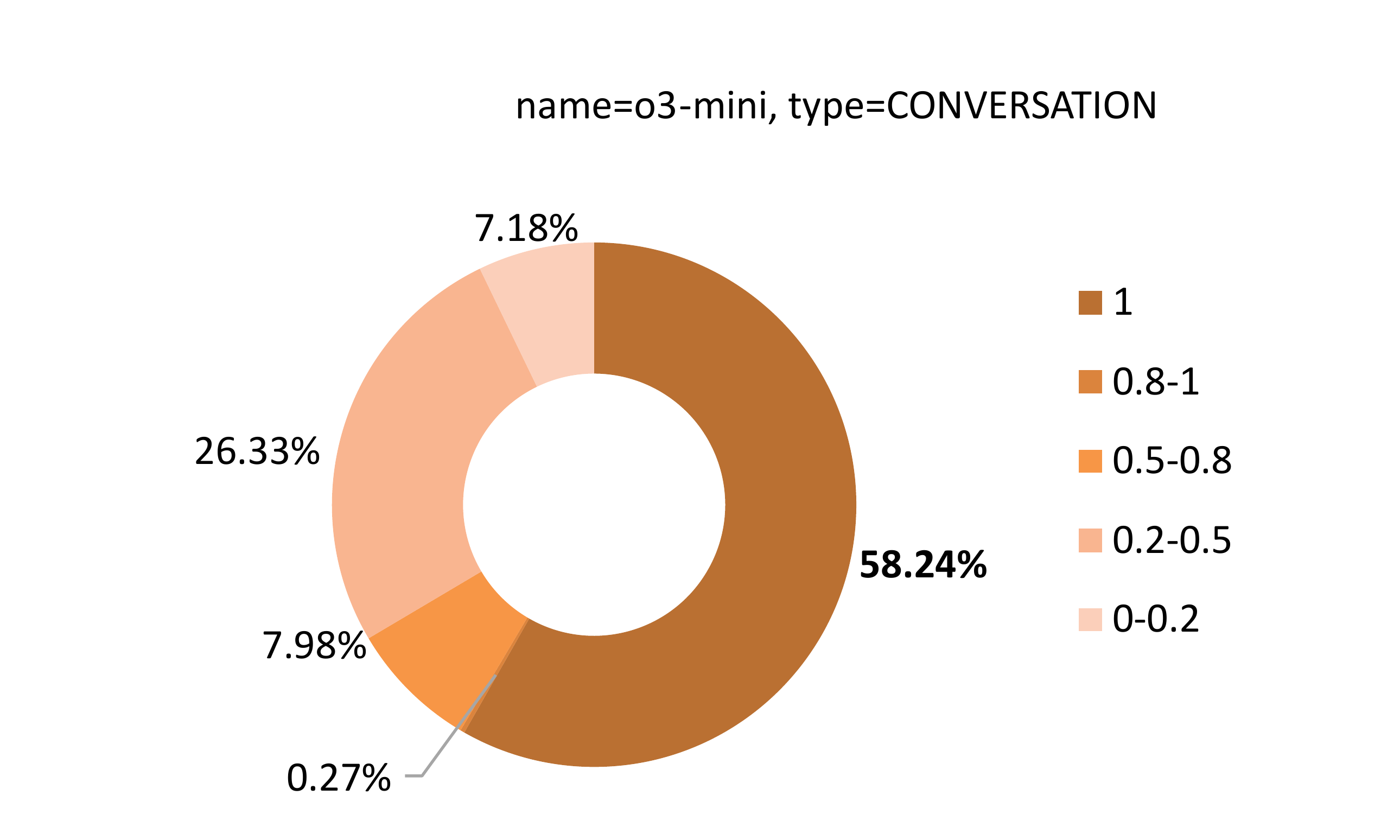}
                        \includegraphics[width=0.5\columnwidth]{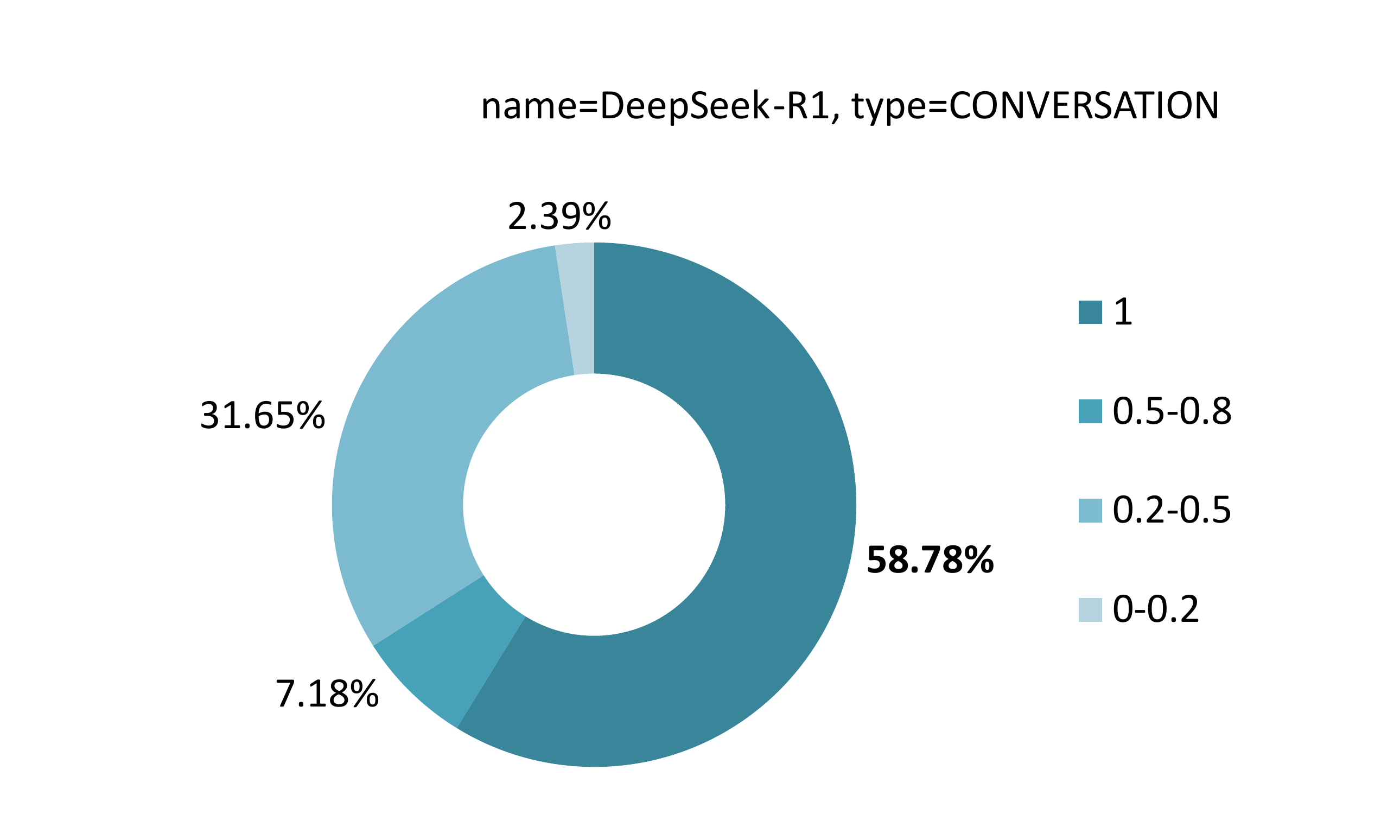}
  \caption{The Anchor Scores distributions of literal-metaphor (LM) word imagination task with sentences in CONVERSATION on every model (the largest portion is in bold and the second largest is underlined).}
  \label{fig:result_imagination10}
\end{figure*}

\begin{figure*}[h]
  \centering
  \includegraphics[width=0.5\columnwidth]{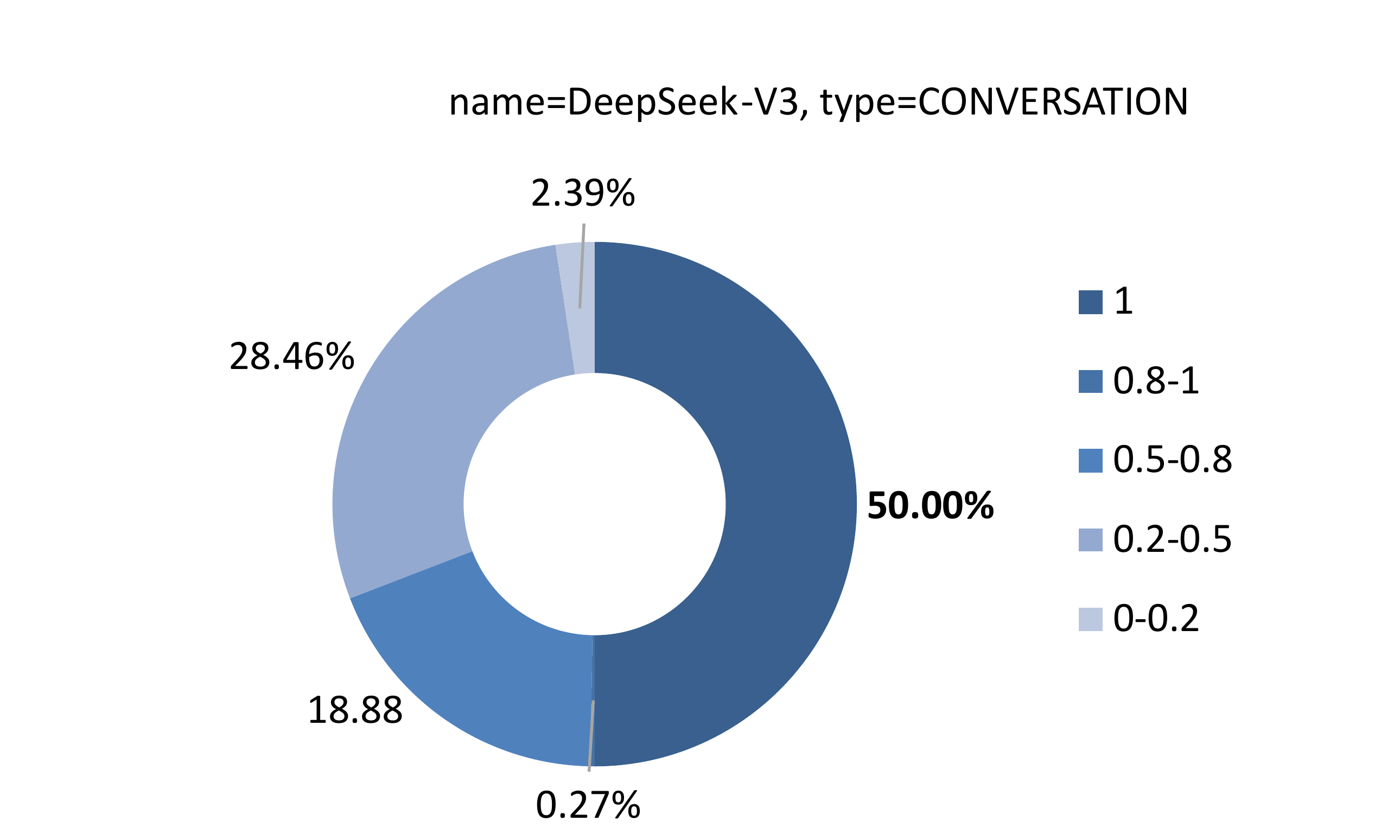}
      \includegraphics[width=0.5\columnwidth]{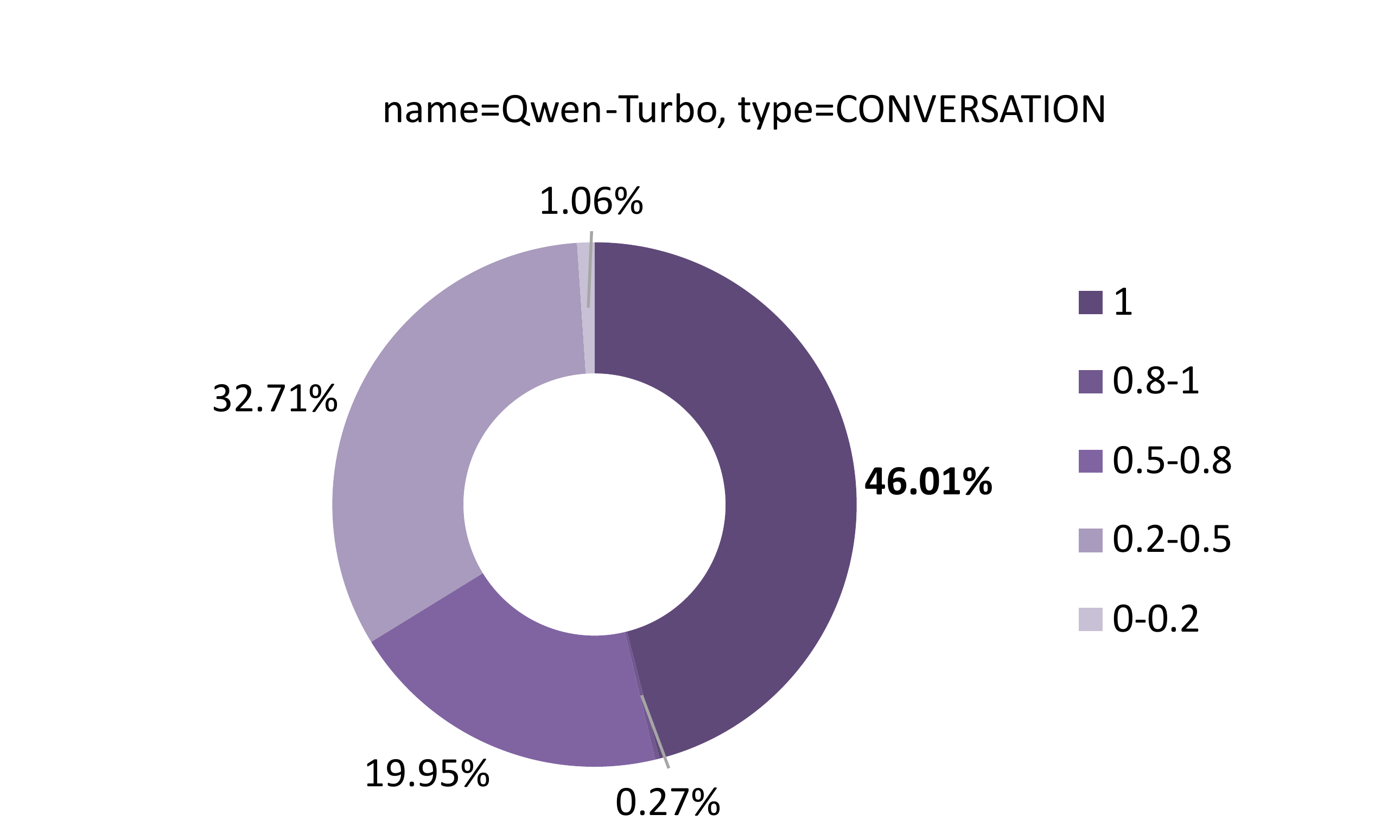}
            \includegraphics[width=0.5\columnwidth]{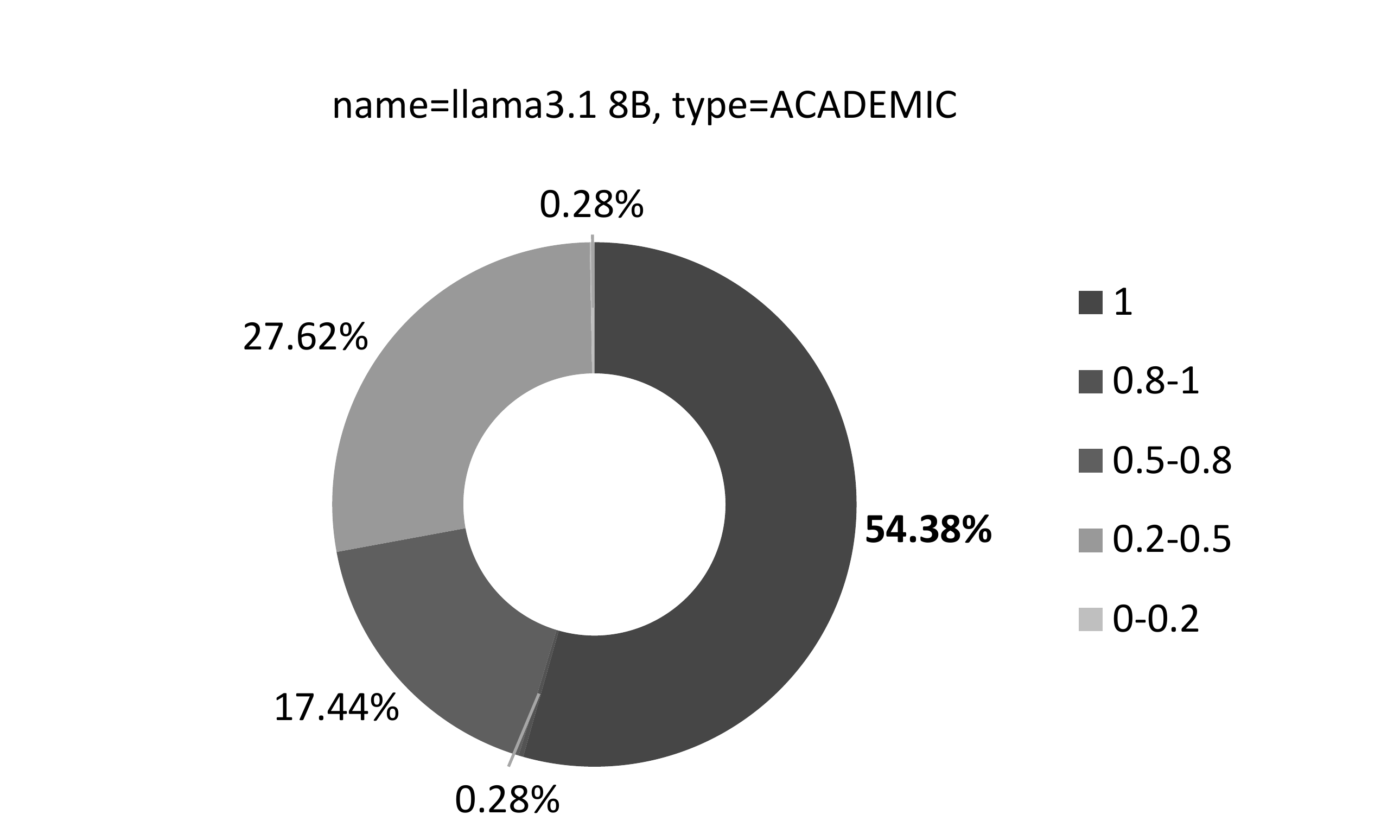}
                  \includegraphics[width=0.5\columnwidth]{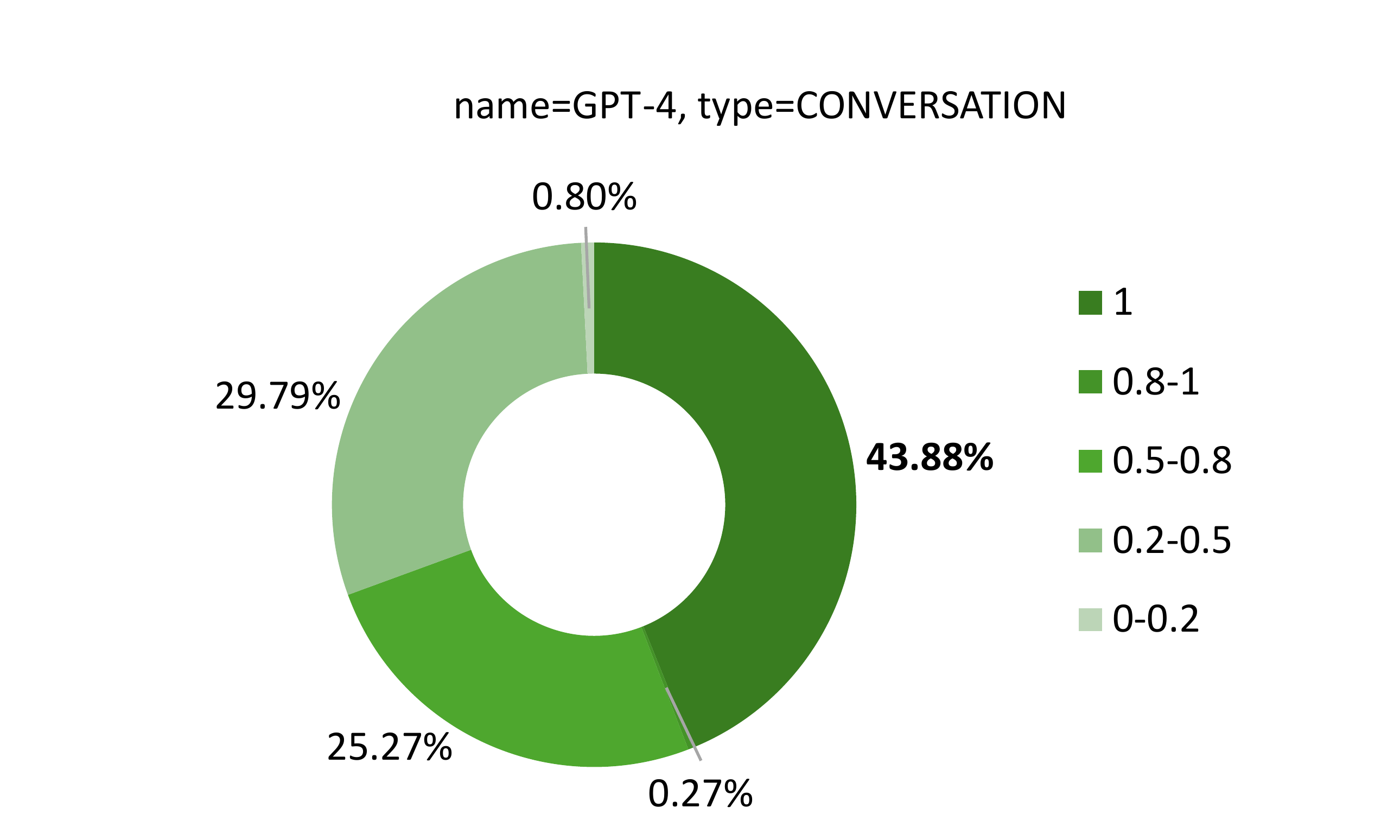}
                        \includegraphics[width=0.5\columnwidth]{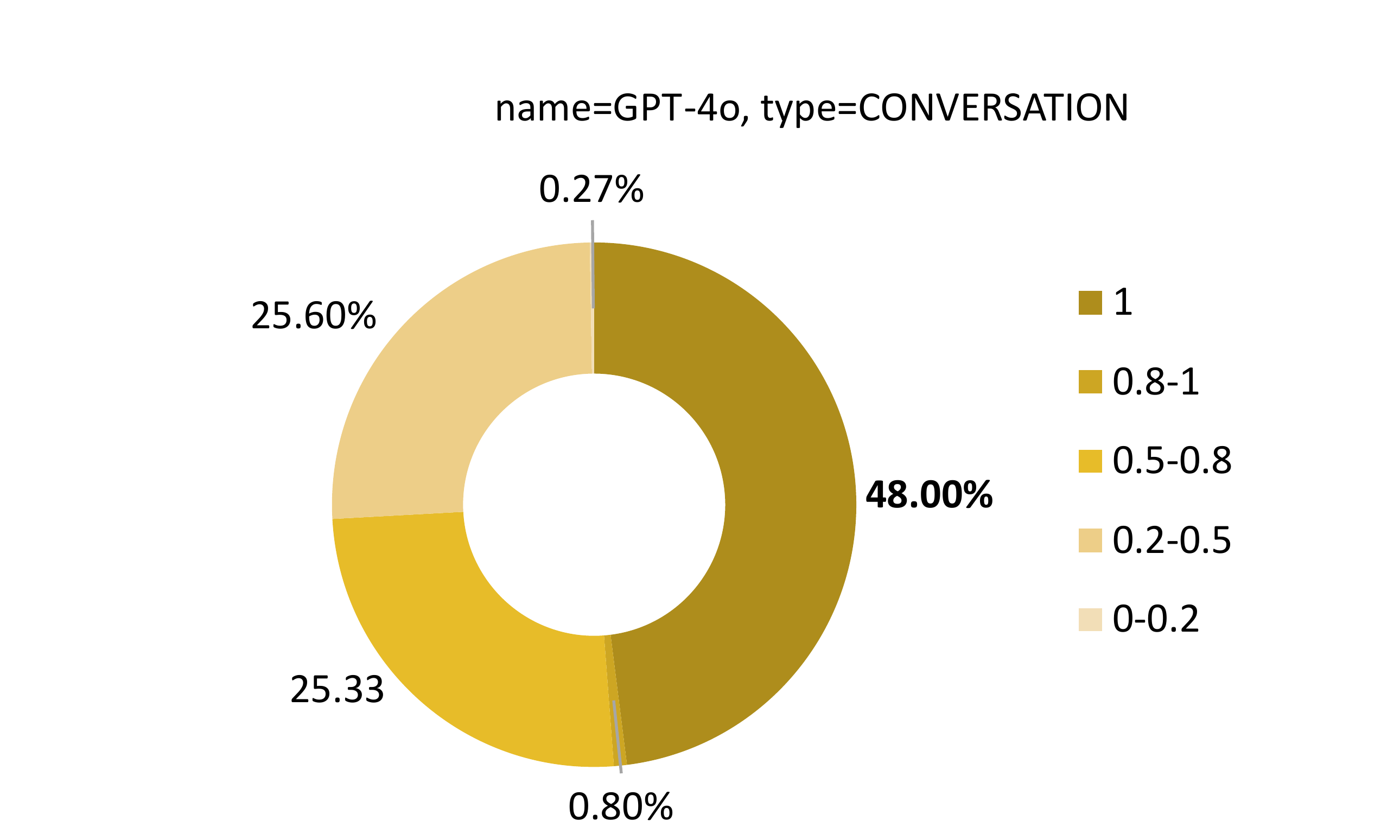}
                                                \includegraphics[width=0.5\columnwidth]{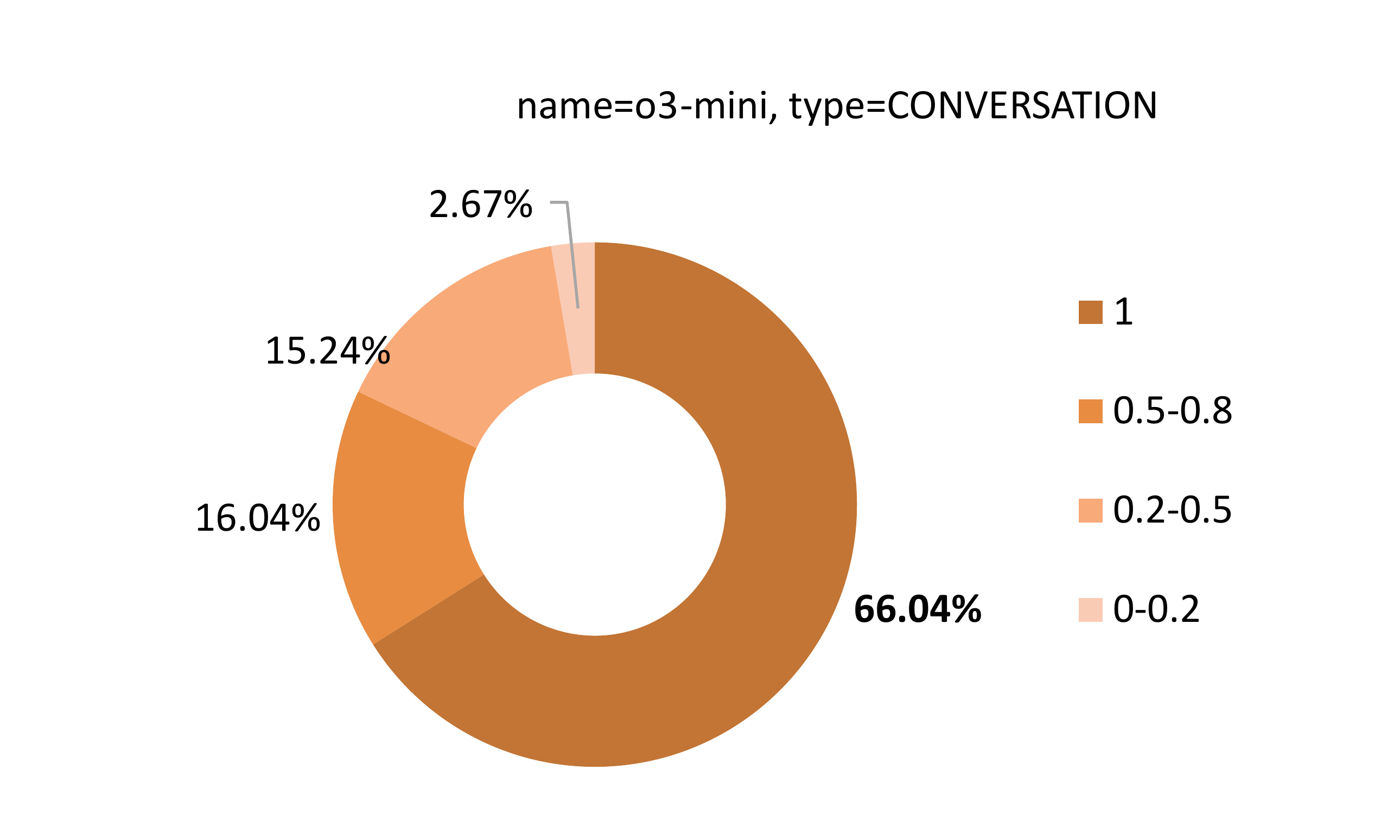}
                                                                        \includegraphics[width=0.5\columnwidth]{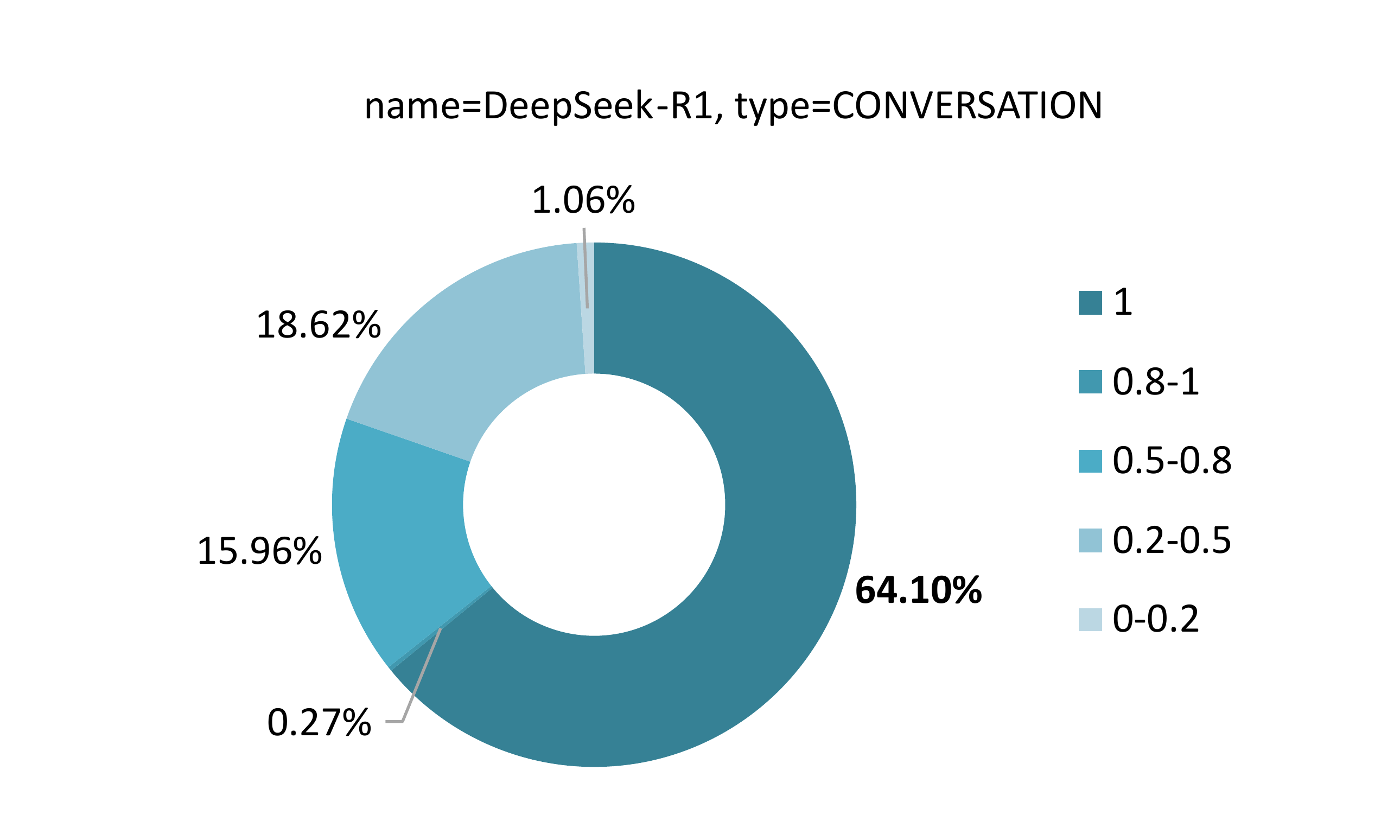}
  \caption{The Anchor Scores distributions of metaphor-literal (ML) word imagination task with sentences in CONVERSATION on every model (the largest portion is in bold and the second largest is underlined).}
  \label{fig:result_imagination11}
\end{figure*}

In the Metaphorical Imagination experiment, we analyze the distribution of \textit{Anchor Scores}, which quantify the degree of lexical overlap between contextualized and decontextualized generation sets for the same target word. Figures~\ref{fig:result_imagination2}--\ref{fig:result_imagination11} report Anchor Score distributions across different tasks (Metaphor-to-Literal and Literal-to-Metaphor) and discourse genres. Higher Anchor Scores indicate the presence of stable lexical associations that persist across contextual conditions, providing empirical evidence for lexical invariance in LLM behavior. Such associations may support certain types of metaphor processing, although their interaction with context-sensitive inference remains an open question. Future work may further investigate how these lexical associations can be more explicitly characterized.

\begin{table*}[!t]
  \centering  
  \begin{tabular}{>{\centering\arraybackslash}p{3cm}p{6.5cm}p{5.5cm}}
      \hline
    \textbf{Terminology}&  \textbf{Definition}&\textbf{Notes/Examples}\\
    \hline
    Selection Preference Violation (SPV)&  The disparity between the context of a word within a sentence and its frequently used contexts is an indicator of this word’s metaphorical usage \cite{tian-etal-2024-theory}.&I \textbf{drank} a bottle of  water. 

Cars \textbf{drink} gasoline.

(Drinking usually connected with human, not car.)\\\hline
 Metaphor Identification Procedure (MIP) & A metaphor is identified if the contextual meaning of the word differs from its basic meaning \cite{tian-etal-2024-theory}.&They \textbf{fall} in love.

Metaphorical: obsessed in feelings

Literal: dropping down\\\hline
 Conceptual Metaphor Theory (CMT) & Metaphors are mappings between source domain (describes tangible objects or concepts) and target domain (represents abstract ideas).&
Metaphor: The arm race.

Source domain: COMPETITION

Target domain:
ARMS BUILDUP
\\\hline
 Semantic Attribute& The salient property or relational feature that is selectively mapped from the source domain to the target domain in a metaphor.& Metaphor: The computer is a turtle.
Semantic attribute: The speed (Slowness)\\\hline
 Reference Plane \(\gamma_i\)& Constructed by the embeddings of three
sentences \(R_1\), \(R_1'\) and \(S_i\), implying target semantic attribute. Representing the ideal attribute the metaphor intended to convey.&\(R_1\): The computer runs fast.

\(R_1'\): The computer runs slow.

\(S_1\): The computer processes fast.\\\hline
 Interpretation Plane \(\beta_i\)& The plane defined by LLM-generated interpretation \(M_i\) and the two human-annotated interpretations \(R_i\) and \(R_i'\) to evalutate the diviation of interpertations.&\(R_1\): The computer runs fast.

\(R_1'\): The computer runs slow.

\(M_1\): The computer runs at a high speed.\\\hline
 Trigger Word Error \cite{wachowiak-gromann-2023-gpt}& The model predict  wrong source domains that were not metaphorically related, because models only infer from the words that are commonly co-occurred instead of  considering context.&Metaphors with the word \textit{arm} may falsely activate war-related interpretations due to frequent lexical co-occurrence, even when the context does not support such mappings.\\\hline
    Lexical Invariance& The tendency of a model to produce the same or highly similar lexical realizations for a given word or concept regardless of whether it is presented in isolation or embedded within a sentential context&Even when the surrounding context does not support such a mapping, models tend to consistently associate metaphors containing the word \textit{arm} with \textit{war}.\\\hline
 Syntactic Influence& The influence of metaphorical syntactic structures in metaphor analysis.&The accuracy of metaphor detection varies depending on different syntactic irregularity settings.\\\hline
 Random Shuffle
& Sentence words
are randomly reordered, disrupting both semantic coherence and syntactic structure, creating unrelated words without meaningful patterns.&council case \textbf{appealed} stated by The.
\\\hline
 Part-of-speech (POS) Shuffle
& Preserving the overall meaning of metaphors
and altering the specific metaphorical words with synonyms
of the same meaning but different POS.&The council \textbf{complainant} (n.) by case stated.
\\\hline
 Metaphorical Word Reposition& The metaphorical word is rearranged to the beginning, a random intermediate location, or the end of the sentence.&\textbf{appealed} The council by case stated.

The council by case \textbf{appealed} stated.

The council by case stated \textbf{appealed}.\\\hline
  \end{tabular}
  \caption{Terminology and definitions in this study.}
  \label{tab:definition_part1} 
\end{table*}




\end{document}